\documentclass[pdflatex,sn-mathphys-num]{sn-jnl}

\usepackage{graphicx}%
\usepackage{multirow}%
\usepackage{amsmath,amssymb,amsfonts}
\usepackage[title]{appendix}%
\usepackage{xcolor}%
\usepackage{textcomp}%
\usepackage{manyfoot}%
\usepackage{booktabs}%
\usepackage[linesnumbered,ruled,vlined]{algorithm2e}
\usepackage{algpseudocode}%
\usepackage{listings}%

\theoremstyle{thmstyleone}%
\theoremstyle{thmstyletwo}%

\theoremstyle{thmstylethree}%
\usepackage{array}
\usepackage{url}

\usepackage{placeins}

\begin{document}

\title[Online Structure Learning and Planning for Autonomous Robot Navigation using Active Inference]{Online Structure Learning and Planning for Autonomous Robot Navigation using Active Inference}

\author*[1]{\fnm{Daria} \sur{de Tinguy}}\email{daria.detinguy@ugent.be}

\author[2]{\fnm{Tim} \sur{verbelen}}\email{Tim.verbelen@verses.ai}
\author[3]{\fnm{Emilio} \sur{Gamba}}

\author[1]{\fnm{Bart} \sur{Dhoedt}}\email{Bart.dhoedt@ugent.be}

\affil*[1]{\orgname{Ghent University}, \orgaddress{\country{Belgium}}}

\affil[2]{\orgname{Verses}, \orgaddress{\country{canada}}}

\affil[3]{\orgname{FlandersMake}, \orgaddress{\country{Belgium}}}


\abstract{
Autonomous navigation in unfamiliar environments requires robots to simultaneously explore, localise, and plan under uncertainty, without relying on predefined maps or extensive training. We present Active Inference MAPping and Planning (AIMAPP), a framework unifying mapping, localisation, and decision-making within a single generative model, drawing on cognitive-mapping concepts from animal navigation (topological organisation, discrete spatial representations and predictive belief updating) as design inspiration.
The agent builds and updates a sparse topological map online, learns state transitions dynamically, and plans actions by minimising Expected Free Energy. This allows it to balance goal-directed and exploratory behaviours.
We implemented AIMAPP as a ROS-compatible system that is sensor and robot-agnostic and integrates with diverse hardware configurations. It operates in a fully self-supervised manner, is resilient to sensor failure, continues operating under odometric drift, and supports both exploration and goal-directed navigation without any pre-training. We evaluate the system in large-scale real and simulated environments against state-of-the-art planning baselines, demonstrating its adaptability to ambiguous observations, environmental changes, and sensor noise. The model offers a modular, self-supervised solution to scalable navigation in unstructured settings. AIMAPP is available at \url{https://github.com/decide-ugent/aimapp}.
}


\maketitle

\section{Introduction}




\begin{figure}[!ht]
    \centering
    \includegraphics[width=1.0\linewidth]{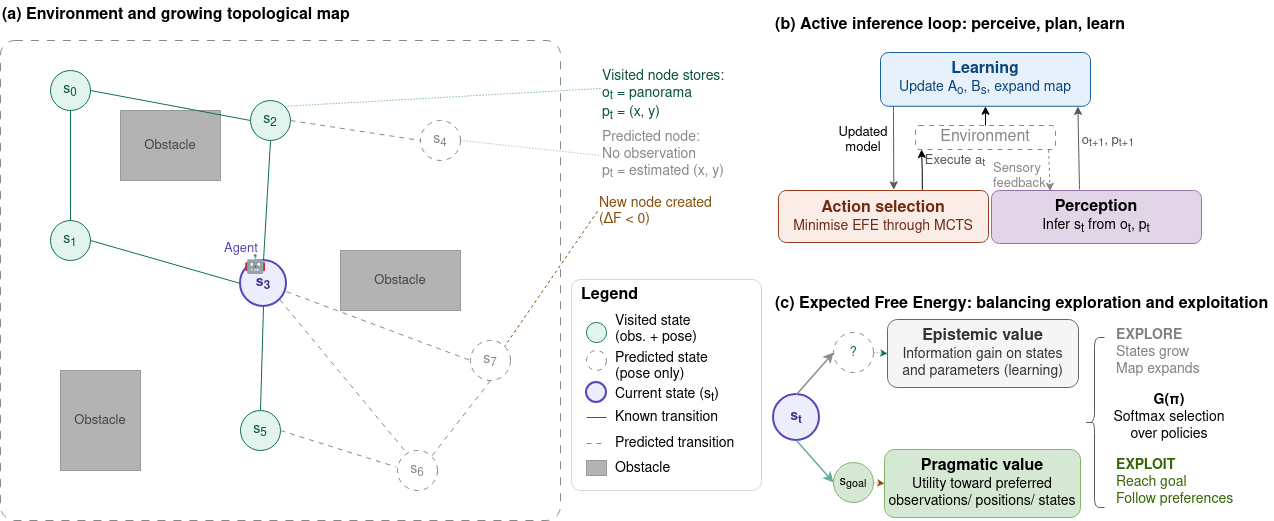}
    \caption{Overview of the AIMAPP framework. (a)~The agent incrementally builds a topological graph; visited nodes (solid) store a panoramic observation and metric position, predicted nodes (dashed) hold only an estimated position. (b–c)~Navigation follows a perception–learning–action cycle in which actions are selected by minimising Expected Free Energy, decomposed into epistemic and pragmatic value.
}
    \label{fig:whatwedo}
\end{figure}
Autonomous mobile robots increasingly operate in environments for which no prior map is available (warehouses being reconfigured, homes the robot has never entered, outdoor spaces with shifting layouts). In such settings, the robot must localise, build a representation of its surroundings, and plan its motion concurrently and online, often under sensor noise, perceptual aliasing and partial observability~\cite{path_planning_survey,survey_RL_robots,pietro}. Each of these subproblems has been studied extensively in isolation, but combining them into a single robust system that does not require pre-training, hand-engineered priors, or globally consistent metric reconstruction remains an open challenge.

Classical approaches address subsets of this problem with characteristic trade-offs. Metric SLAM systems such as ORB-SLAM3~\cite{orbslam3} and FAST-LIO2~\cite{FAST-LIO2} provide accurate pose estimation but are prone to drift in large or unstructured environments and offer no exploration policy of their own. Path planners such as ETPNav~\cite{ETPNav} achieve high path quality at substantial computational cost. Frontier-based exploration~\cite{frontiers} is lightweight but greedy, often producing redundant backtracking. Learned approaches such as Neural-SLAM~\cite{NNslam} improve flexibility through end-to-end training but require extensive offline data collection. Cross-embodiment models such as ViKiNG~\cite{viking} and NoMaD~\cite{Nomad} achieve impressive generalisation to unseen environments but depend on large-scale pre-training or teleoperated demonstrations in the target environment to construct their topological maps before autonomous deployment. These data requirements limit adaptability to genuinely novel settings where no prior experience is available, and no human intervention is desired.
 
The Active Inference Framework (AIF)~\cite{AIF_book} offers a different starting point. AIF is a computational theory of perception and action under uncertainty in which an agent maintains probabilistic beliefs about latent states of the world, predicts the consequences of candidate actions, and selects actions that minimise the expected divergence between predicted and preferred outcomes. AIF is a natural fit for online navigation because it derives both perception and action from the same probabilistic objective: exploration (resolving uncertainty about the world) and goal-reaching (steering observations toward preferences) emerge as two complementary contributions of the same Expected Free Energy (EFE) term, rather than as separate behavioural modes that must be hand-arbitrated.

Building on this idea, we propose AIMAPP (Active Inference MAPping and Planning, Figure~\ref{fig:whatwedo}), a navigation framework that maintains a sparse topological map linking sensory observations to local metric coordinates, and uses it to plan trajectories balancing information gain with goal pursuit. AIMAPP does not rely on pre-training, predefined map structures, or globally consistent metric representations~\cite{world_model_and_inference}. Instead, it adapts its beliefs continuously from sensorimotor feedback.
To keep planning tractable, the agent evaluates only context-relevant policies through Monte Carlo Tree Search (MCTS) over EFE~\cite{ours_model}. Localisation is handled probabilistically rather than relying solely on raw odometry, providing resilience to drift in recognisable environments (the mechanism is analysed in Section~\ref{sec:localisation}, Figure~\ref{fig:loc_ex}). Goals are autonomously assigned and reassessed, and the agent recovers from most motion failures without requiring manual intervention.

AIMAPP exhibits the following key properties:
\begin{itemize}
    \item \textbf{Online learning and self-supervision}: the agent builds and refines its world model from sensorimotor experience, without pre-training or human supervision. This probabilistic formulation provides resilience to sensor drift and tolerates moderate environmental change.
    \item \textbf{Modality and platform flexibility}:  any sensor producing comparable observations (RGB, depth, LiDAR) can be integrated, and the model deploys on any ROS-compatible robot.
    \item \textbf{Unified explore/exploit}: exploration and goal-reaching emerge from the same Expected Free Energy objective without task-specific tuning.
\end{itemize}

We evaluate AIMAPP in simulated and real environments of varying scale and complexity. Across exploration trials, AIMAPP achieves coverage efficiency comparable to or exceeding state-of-the-art planning-based methods (FAEL~\cite{FAEL}, GBPlanner~\cite{gbplanner}, Frontiers~\cite{frontiers}), while achieving coverage efficiency within 74\% of human-teleoperated performance on average across all simulated and real-world environments (including Appendix~\ref{app:explo_paths}). The system demonstrates robustness to moved obstacles in simulation, sensor drift and sensor uncertainty in the real world, and continues operating in conditions (sandy, carpeted or mixed flooring) where odometry-based baselines such as Frontiers with Nav2~\cite{nav2} fail outright. Exploration and goal-reaching are achieved within a single architecture, without task-specific re-tuning, across environments ranging from simulated houses and warehouses to real-world garages up to 325~m$^2$.

Together, these contributions demonstrate that Active Inference, originally developed in computational neuroscience, can be adapted into a modular and robust navigation strategy for real robots, in which navigation is treated as hypothesis testing: predictions about future states are generated, evaluated through action, and revised based on surprise. 






\section{Related Work}


Autonomous navigation in unknown environments requires solving several interconnected problems: localisation, mapping, path planning and obstacle avoidance. We review representative approaches below, organised by the trade-offs most relevant to the present work.

Classical navigation methods typically rely on metric mapping via Simultaneous Localisation and Mapping (SLAM). Notable examples include ORB-SLAM3~\cite{orbslam3} for visual features, PLP-SLAM~\cite{PLP-SLAM} for line-and-plane features, and FAST-LIO2~\cite{FAST-LIO2} and depth-based SLAM~\cite{depth_based_LiDAR_slam} for LiDAR-based pose estimation. These approaches excel in structured, uniformly lit conditions, but are prone to drift over time, even with loop closure, particularly in large or dynamic environments. Furthermore, metric methods often scale poorly in memory and do not integrate planning, as path planning and obstacle avoidance require additional modules~\cite{review_path_plan}. 
Topological mapping offers a more scalable alternative to metric maps, representing the environment as a graph of connected states rather than dense coordinates~\cite{gaussian_process_nav,map_induction}. This improves scalability and can better handle perceptual aliasing. 
Topological representations have been adopted in several learning-based navigation systems: ViKiNG~\cite{viking} builds a topological graph from offline traversal data for cross-environment transfer, while NoMaD~\cite{Nomad} constructs a graph from teleoperated demonstrations in the target environment. In bio-inspired SLAM, RatSLAM~\cite{ratslam} builds an experience graph using pose cells and visual templates, achieving robust loop closure without metric consistency, but it provides no autonomous decision-making or goal-directed behaviour. Moreover, most topological methods rely on heuristics or prior data for node creation and are sensitive to environmental change or perceptual noise. AIMAPP inherits the scalability of topological maps while determining node creation through a free-energy criterion rather than a hand-designed heuristic or pre-collected dataset.

Reinforcement learning enables agents to acquire navigation policies through interaction with the environment, improving decision-making under sensor noise and partial observability~\cite{survey_RL_robots}. Neural-SLAM~\cite{NNslam} integrates cognitive mapping with policy learning for visual exploration. Self-supervised methods such as BYOL~\cite{BYOL} and RECON~\cite{RECON} learn visual representations without explicit supervision but require months of data collection. World-model methods learn environment dynamics for planning: \cite{pietro} uses latent Bayesian surprise to drive exploration, while~\cite{world_model_explo} predicts action outcomes to guide policy selection. Active NTFields~\cite{Liu_2025} takes a physics-informed approach, learning arrival-time fields online from depth observations for fast motion planning without pre-training. Bio-inspired neural networks have also been applied to reactive, collision-free navigation~\cite{bio-inspired-robots}, though most assume static environments and lack probabilistic reasoning.
 
These methods are sometimes described as "zero-shot" in the sense that they transfer to new environments without fine-tuning. However, this usage differs from online structure learning, where an agent discovers the environment's structure from the first interaction without any prior data collection. ViKiNG requires extensive offline training on diverse environments; NoMaD requires preliminary teleoperation (or fixed policy) in the target environment; our model requires neither. Most learning-based methods are also task-specific, trained for either exploration~\cite{BYOL,RECON,NNslam} or goal-reaching~\cite{viking,RL_path_plan,world_model_explo}. In contrast, AIMAPP requires no pre-training or teleoperated demonstrations, supports any input modality (visual, semantic, point cloud) and unifies exploration and goal-reaching within a single online-learning framework.
 
AIF treats navigation as inference rather than control: the agent acts to minimise EFE, integrating sensing, prediction, localisation and planning within a single generative process~\cite{AIF_book}. A comprehensive survey of AIF for robotics is provided by Lanillos et al.~\cite{AIF_survey_robotics}, covering estimation, control, planning and learning. Early AIF navigation work addressed small grid worlds~\cite{plan_nav_AIF_friston} or simple structured environments~\cite{plan_nav_AIF_friston,Human_rodent_spatial_rep,weird_HAIF}. Sophisticated inference~\cite{sophisticated_AIF} extended the planning horizon through recursive EFE evaluation, while Fountas et al.~\cite{MCTS_AIF} demonstrated that deep AIF combined with MCTS can scale to more complex tasks, a combination directly relevant to our MCTS-EFE integration.
 
More recently, \c{C}atal et al.~\cite{GSLAM} proposed GSLAM, a SLAM solution learning visual state representations with deep neural networks and constructing topological maps through an experience graph, with validation on a real warehouse robot. Their work showed that localisation and mapping can emerge naturally from free-energy minimisation. Other work has combined AIF with imitation learning for dynamic replanning in simple real-world scenarios~\cite{aif_nav_imitation}. However, both require pre-training over possible actions or observations and do not support exploration in open-ended environments.
 
AIMAPP differs from these prior AIF approaches in three respects: firstly, it operates in a fully online fashion without pre-training or offline data collection. Secondly, it unifies exploration and goal-reaching within a single generative model rather than addressing one task in isolation; and finally, it demonstrates competitive performance against state-of-the-art classical planning methods (FAEL, GBPlanner, Frontiers) in real environments up to 325~m$^2$, a scale not previously addressed by AIF navigation systems. Relative to our own prior work,~\cite{ours_model} introduced the discrete-state POMDP and fixed policies integration in small 2D MiniGrid environments without realistic observations, while \cite{ours_hierarchy} proposed the hierarchical generative model with learned visual representations. The present paper offers the new $(s_t, p_t)$ decomposition coupling topological and metric information, the online map-expansion mechanism governed by free-energy comparison, the MCTS-EFE integration, the ROS-compatible, sensor-agnostic system architecture, and the experimental evaluation at scale in seven real and simulated environments against non-learning baselines.

\section{Method}
\label{sec:method}

\begin{figure}[!ht]
    \centering
    \includegraphics[width=0.45\linewidth]{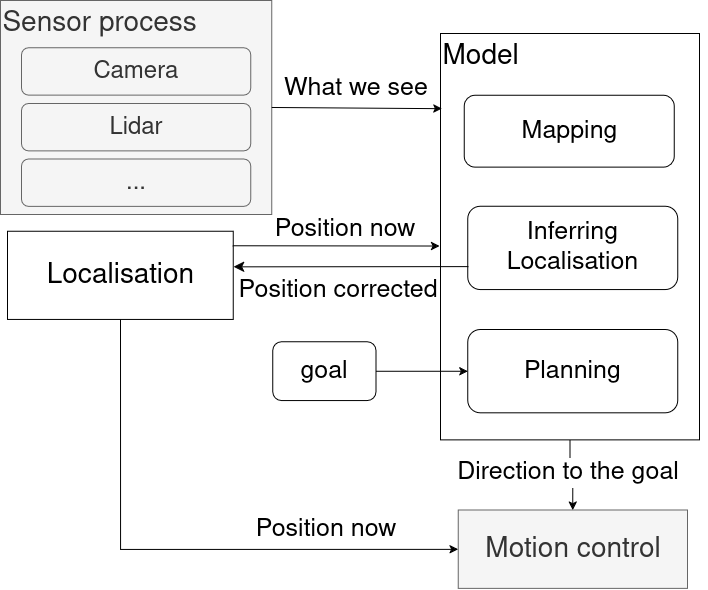}
    \caption{Overview of the system architecture. Any ROS-compatible solution can replace grey modules. The cognitive layer (localisation, mapping, and action selection) relies on the AIF framework, while perception and motion planning use traditional approaches. Believed odometry takes precedence over sensor odometry. Preferences (goals) are provided by the user when a target observation is to be reached.}
    \label{fig:archi}
\end{figure}

AIMAPP operates through a recurring sense-predict-learn-act loop formalised under the Active Inference Framework. At each step, the agent 1) infers its current state from sensory observations (perception), 2) simulates possible action sequences and scores them against its beliefs and goals (prediction and action selection), 3) updates its internal model based on the discrepancy between previously predicted and actual outcomes, as well as currently predicted outcomes (learning) and 4) executes the selected action. This loop is implemented as a partially observable Markov decision process (POMDP), where candidate policies $\pi$ are evaluated using Monte Carlo Tree Search (MCTS) and scored according to EFE. The system architecture, shown in Figure~\ref{fig:archi}, separates the AIF-based cognitive layer (localisation, mapping, and action selection) from interchangeable perception and motion planning modules, enabling integration with any ROS-compatible platform.

\subsection{State Representation and Map Structure}
\label{sec:mapping}


The agent maintains a cognitive map, an internal representation inspired by biological navigation systems~\cite{world_model_and_inference,cscg_structuring_knowledge,humans-mapping,humans-cognitive-map,cognitive_graph}. Cognitive maps support flexible navigation and spatial reasoning, capturing both the layout of the environment and the agent's experience within it. In our approach, the cognitive map is implemented as a topological graph containing metric information, where nodes represent distinct agent states ($s$), each corresponding to a spatial location ($p$) with an associated observation ($o$). Metric information is retained locally for each node, enabling approximate spatial reasoning without requiring globally consistent coordinates. The map grows adaptively as the agent explores, adding new states when new poses are expected in given directions, supporting scalability to large or dynamic environments.


The latent state is decomposed into $s_t$ (a discrete index representing a node in the topological graph) and $p_t$ (continuous 2D metric coordinates used for local spatial reasoning). The agent observes $o_t$, in the present implementation a $360^\circ$ panoramic RGB image used for place recognition and state inference.

The decomposition of the latent variable into topological state $s_t$ and metric position $p_t$ serves both computational and functional purposes. Functionally, this separation loosely mirrors the distinction observed in mammalian navigation systems between place-cell-based location identity and grid-cell-based metric path integration~\cite{grid_cell_nav,Human_rodent_spatial_rep}. Computationally, the factorisation enables independent update dynamics: position beliefs evolve continuously through odometry integration, while state beliefs update discretely upon observation and position matching. This separation is particularly advantageous in aliased environments, where identical observations may correspond to distinct locations. By maintaining a separate position estimate $p_t$, the agent can disambiguate perceptually identical states through accumulated metric displacement, even when observations alone are insufficient~\cite{cscg_structuring_knowledge}. In our setting, a unified representation of $p$ and $s$ into a single distribution would require either discretising metric space (sacrificing localisation precision) or treating place identity continuously (increasing computational complexity without functional benefit) and serve to lead the map expansion over the expected states.


\paragraph{Node creation and connectivity}
New nodes are created when a motion leads to a predicted position $p_t$ exceeding a radius of influence from existing states. This expansion step arises from a discrepancy between the current model parameters $A_p$ and expected model parameters $\hat{A_p}$ if the agent were to move to that new position, as defined by the expected information gain on parameters in the EFE (Eq.~\eqref{eq:efe_wt_c}) and the free energy comparison in Eq.~\eqref{eq:PvsP}.

\begin{figure}[!htb]
    \centering
    \includegraphics[width=0.5\linewidth]{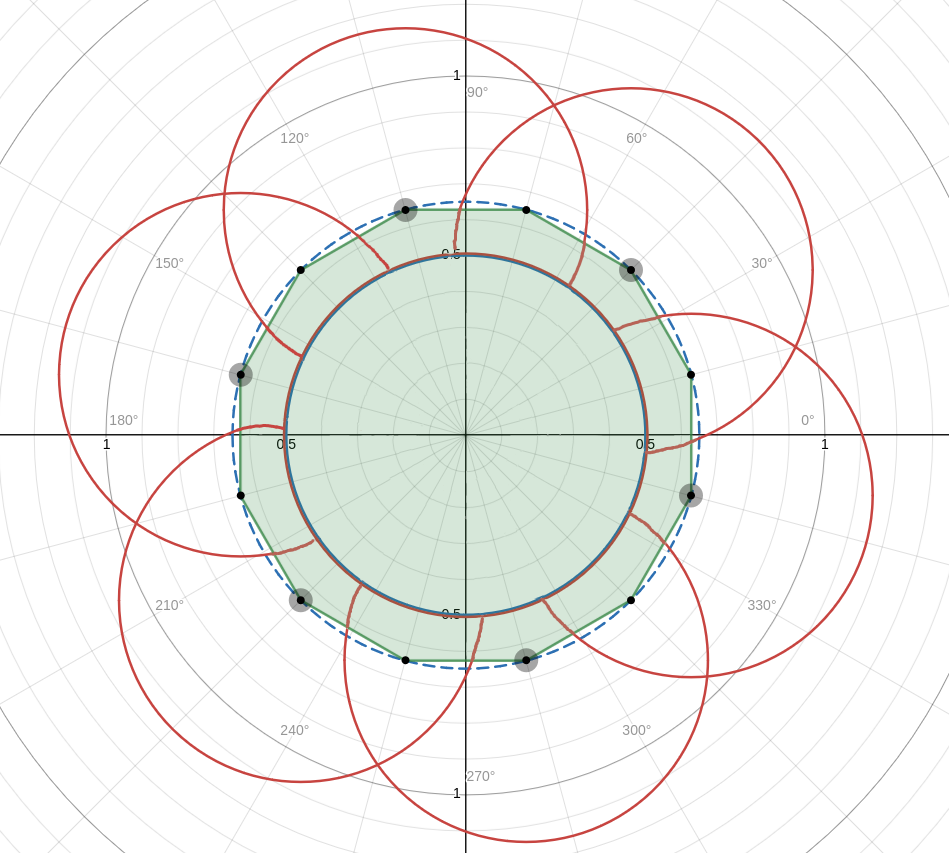}
    \caption{Influence of a node at position (0,0) on adjacent node creation for an influence radius of 0.5~m and 12 discrete headings spanning $360^\circ$. Red circles represent the influence radius of each newly created node. The blue dashed circle marks the robot's radius, including a padding term to account for its physical size (important near walls). The 12 black dots correspond to the midpoints of the 12 possible action directions. Dots with a black aura indicate valid positions where a new node can be created while respecting the minimum distance constraint (red radius) of adjacent nodes. Dots without an aura represent invalid positions (too close to an existing node) and could instead be created farther away (e.g., at 1~m). This arrangement allows the agent to maintain open junctions for future node expansion.}
    \label{fig:inf_radius}
\end{figure}

The discrete set of available actions (e.g., 12 headings plus a "stay" action, the value is user-defined) defines the potential connectivity. This process is illustrated in Figure~\ref{fig:inf_radius}: each state can spawn neighbours only when motion carries the agent outside the influence zone of previous nodes, ensuring sparse but informative coverage.

\paragraph{Observations and place recognition}

\begin{figure}[!htb]
    \centering
    \includegraphics[width=0.50\linewidth]{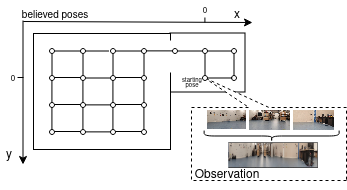}
    \caption{Schematic of a topological map in a simple two-room environment. Dots are states (nodes), each containing a $360^\circ$ panorama obtained through stitching images together and associated with a position. Links between nodes are plausible transitions.}
    \label{fig:map_obs_schema}
\end{figure}

Observations are stored as $360^\circ$ panoramas stitched from consecutive camera frames (Figure~\ref{fig:map_obs_schema}). Each node encodes a coarse spatial position and a panoramic visual observation. Incoming observations are compared to stored visual memories using the Structural Similarity Index (SSIM) to determine familiarity. If no match is found, this signals either a novel location or a significant environmental change.

\paragraph{Sensory and motion agnosticism}

AIMAPP is designed to be modular: it accepts recognised observations from any perception system and relies on a motion controller to report goal-directed stops. This allows interchangeable components within the architecture, as shown in Figure~\ref{fig:archi}. The current implementation uses panoramic images compared via SSIM, but the generative model operates on abstract observation variables and is agnostic to how those observations are derived from raw sensory input. Limitations of the current perception front-end and directions for more robust encodings are discussed in Sections~\ref{sec:limitations} and~\ref{sec:future_work}.

\subsection{Generative Model and Inference}
\label{sec:gen_model}


World models are internal representations of the environment generating predictions about possible future sensory information. Classically, worlds are modelled as a POMDP with, at any time $t$, the current observation $o_t$ and determined past motion $a_{t-1}$ from which the agent can infer state $s_t$. 


Given the decomposition of the latent state into $s_t$ and $p_t$ introduced in Section~\ref{sec:mapping}, the generative model takes the extended form of Equation~\eqref{eq1}. Tildes~(~$\tilde{}$~) denote sequences over time.

\begin{equation} 
\begin{aligned}
\mathcal{P}(\tilde{o}, \tilde{s}, \tilde{p} ,\tilde{a}) =& P(o_0| s_{0})P(s_0)P(p_0) \\ & \prod_{t=1}^\tau 
P(o_t| s_{t})P(s_t, p_t|s_{t-1},p_{t-1},a_{t-1})  
\end{aligned}
\label{eq1}
\end{equation} 

\begin{figure}[!htb]
    \centering
    \includegraphics[width=0.70\linewidth]{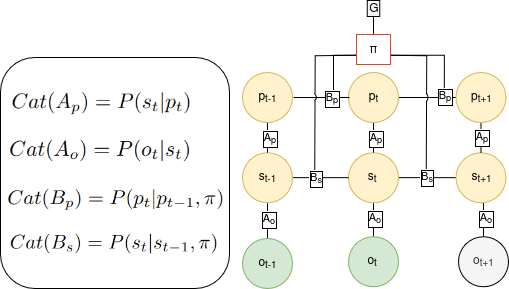}
    \caption{POMDP of our model. $Cat$ stands for Categorical. The model integrates states ($s_t$), positions ($p_t$), and observations ($o_t$) over time, guided by policies ($\pi$) and expected free energy ($G$). The categorical distributions define transition and observation likelihoods: $A_p$ (position likelihoods), $A_o$ (observation likelihoods), $B_p$ (position transitions), and $B_s$ (state transitions). This structure underpins the inference scheme in Equation~\eqref{eq2}, enabling the agent to infer hidden states and positions from sensory observations and prior beliefs.}
    \label{fig:pomdp}
\end{figure}

The generative model is parameterised by four distributions, illustrated in the POMDP graph of Figure~\ref{fig:pomdp}:
\begin{itemize}
    \item State transitions ($B_s = P(s_t|s_{t-1}, a_{t-1})$): the likelihood of moving between topological states given an action.
    \item Position transitions ($B_p = P(p_t|p_{t-1}, a_{t-1})$): the likelihood of moving between metric positions given an action.
    \item Observation likelihoods ($A_o = P(o_t|s_t)$): the probability of observing a particular panorama at a given state.
    \item Position likelihoods ($A_p = P(p_t|s_t)$): the probability of occupying a particular metric position at a given state.
\end{itemize}

In practice, calculating the true posterior $\mathcal{P}$ is typically intractable, because the model evidence $P(o)$ requires marginalising over all latent states~\cite{AIF_book}. Thus, we resort to variational inference, which introduces the approximate posterior $Q$. In our temporal model, it takes the form defined in Equation~\eqref{eq2}. The goal is to make $Q(s)$ as close as possible to $\mathcal{P}(s|o)$, which is achieved by minimising their Kullback-Leibler (KL) divergence.

\begin{equation} 
Q(\tilde{s},\tilde{p}| \tilde{o}, \tilde{a}) = Q(s_0, p_0| o_0) \prod_{t=1}^\tau
Q(s_t, p_t| s_{t-1},p_{t-1}, a_{t-1}, o_t )  
\label{eq2}
\end{equation} 


How well this approximation fits the evidence is measured by the Variational Free Energy (VFE) denoted $F$, defined in Equation~\eqref{eq:F}. Negative $F$ is known as the Evidence Lower Bound (ELBO) in machine learning.

\begin{equation}
    F_Q = \underbrace{D_{KL}[Q(\tilde{s},\tilde{p}|\tilde{a},\tilde{o}))||\mathcal{P}(\tilde{s},\tilde{p}|\tilde{a},\tilde{o})]}_\text{posterior approximation} -\underbrace{\log[ P(\tilde{o})]}_\text{log evidence}
    \label{eq:F}
\end{equation}

Equation~\eqref{eq:F} decomposes VFE into two terms: the KL divergence between $Q$ and the true posterior (which measures approximation quality) and the negative log-evidence (which measures how well the generative model explains observations). Although this decomposition references the intractable true posterior, VFE itself can be computed without it: an equivalent expression depends only on the approximate posterior $Q$ and the joint generative model $P(\tilde{o}, \tilde{s}, \tilde{p}, \tilde{a})$, both of which are available by construction. Minimising $F$ therefore simultaneously tightens $Q$ toward the true posterior and increases the model evidence, all without ever evaluating the true posterior directly~\cite{AIF_book}.


\subsection{Localisation}
\label{sec:localisation}

Localisation in AIMAPP applies the VFE minimisation introduced in Equation~\ref{eq:F} to the topological map. At each step, the agent infers which state $s_t$ best explains the current observation $o_t$ and motion prediction $P(p_t|p_{t-1},a_{t-1})$ by evaluating the posterior distribution over states, combining predicted motion with sensory evidence. This makes localisation belief-driven: the agent does not depend solely on raw odometry or sensor data, but continuously evaluates which internal state best accounts for its sensory history. When uncertainty over $s_t$ is high, localisation becomes an active process: the agent triggers goal-oriented navigation toward previously observed locations to re-localise.

A key implication of this design is that internal localisation is robust to drift. Since the agent prioritises consistency between predicted and observed outcomes, the exact physical position is less critical than whether its internal model correctly explains the sensory data. In familiar areas, mismatches caused by odometry drift are corrected when the agent revisits a state with a recognisable stored observation; between such events, the belief layer tracks sensor odometry closely. In unexplored areas, however, drift cannot be directly corrected, and the agent must rely on its generative model to maintain consistency until additional information becomes available. If severe drift accumulates before the agent revisits a known region, the resulting map may contain spatially inconsistent nodes, leading to incorrect re-localisation or redundant state creation upon return.

If the agent is confident in its current state $s_t$ (considering VFE against a threshold defined in Appendix~\ref{app:model_param}), the likelihood matrix $A_o$ is updated to associate the current observation with the believed state, improving robustness against perceptual aliasing and minor visual variations and reducing future surprise. However, if VFE confidence is low, the agent prioritises re-localisation, searching for familiar observations before updating the model.

\begin{figure}[!htb]
    \centering
    \includegraphics[width=0.5\linewidth]{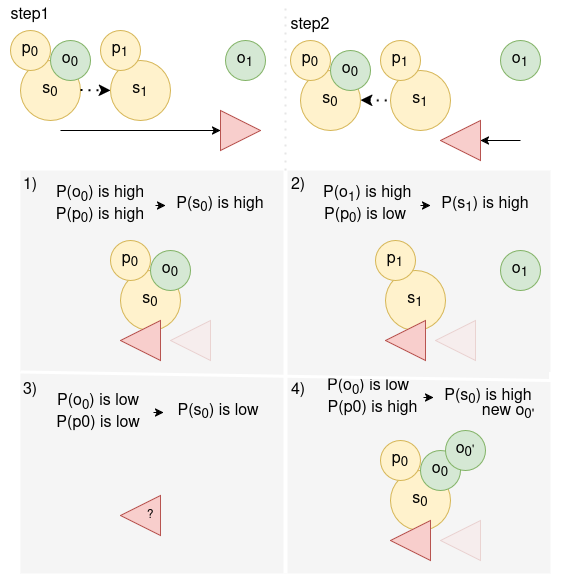}
    \caption{Impact of drift on the agent's localisation. The top row illustrates the scenario: the red triangle is the agent; the solid line shows the true trajectory; the dashed line shows the trajectory perceived through odometry; the yellow circles are inferred states; the green circles are observations. The agent intends to move 1~m right from $s_0$ but actually moves 2~m. Its model creates state $s_1$ at the perceived position, while the true observation $o_1$ belongs to the real position. On the next step, the agent moves 1~m left (without drift) and expects to return to $s_0$. Four localisation outcomes are then possible: 1)~High confidence in both: recognises $o_0$, infers $s_0$, drift physically uncorrected. 2)~Low motion confidence, high observation confidence: recognises $o_1$, infers $s_1$. 3)~Low confidence in both: enters exploratory phase. 4)~High position confidence, low observation confidence: adds $o_0'$ to $s_0$.}
    \label{fig:loc_ex}
\end{figure}

To illustrate this, consider the scenario in Figure~\ref{fig:loc_ex}. The agent starts at state $s_0$ with observation $o_0$ at position $p_0$. It intends to move forward 1~m, but due to drift, it actually moves 2~m. Its model, however, creates a new state $s_1$ at $p_1=p_0+1$~m, consistent with the intended motion rather than the true displacement. After moving back 1~m (without drift), the agent expects to be at $s_0$. At this point, it faces an inference problem: which state best explains the current observation?

Depending on its confidence in motion and observation, four outcomes are possible:
\begin{itemize}
    \item Figure~\ref{fig:loc_ex}~1) High confidence in motion and observation: the agent recognises $o_0$ even though it is not exactly at the correct position, infers it is at $s_0$, and updates its belief accordingly. Drift remains uncorrected physically, but the model is internally consistent.
    \item Figure~\ref{fig:loc_ex}~2) Low confidence in motion, high confidence in observation $o_1$: the agent recognises $o_1$, infers it is at $s_1$, and updates its belief over motion to match perception. In this simple example, the drift would be corrected.
    \item Figure~\ref{fig:loc_ex}~3) Low confidence in both: the agent has high uncertainty about its location. When $P(s_t)$ falls below a user-defined threshold (Appendix~\ref{app:model_param}), it enters an exploratory phase, seeking consecutive familiar observations to re-localise.
    \item Figure~\ref{fig:loc_ex}~4) High confidence in position, low confidence in observation: the agent assumes it is at $s_0$ but fails to recognise the input. It therefore adds a new observation ($o_0'$) to $s_0$, refining the observation model to account for variability (e.g., changes in lighting). The previous observation is not replaced; both are linked to the same state.
\end{itemize}

\subsection{Planning and Decision-Making}
\label{sec:planning}

According to the Free Energy Principle, agents must minimise free energy to form a model that best explains their environment~\cite{AIF_book}. For perception, this is implemented by minimising VFE during inference, as described in Section~\ref{sec:gen_model}. For action and planning, the same principle extends into the future: agents should act to minimise the Expected Free Energy $G$ under their candidate policies.

Formally, a policy $\pi$ is a sequence of future actions. The EFE $G(\pi)$ quantifies how well following $\pi$ is expected to reduce surprise~\cite{plan_nav_AIF_friston}. It decomposes into two complementary contributions:
\begin{itemize}
    \item \textbf{Exploration (epistemic value)}: expected information gain reducing uncertainty about the environment.
    \item \textbf{Exploitation (pragmatic value)}: expected alignment with preferred observations, such as reaching a specific target.
\end{itemize}

This decomposition illustrates that exploration and exploitation are not separate objectives bolted onto the model, but emerge naturally from minimising EFE. This parallels a rat in a maze~\cite{mice_in_labyrith}, which alternates between venturing into unknown corridors (uncertainty reduction) and returning to rewarding locations (preference fulfilment).


Candidate policies are generated from the current state through MCTS~\cite{MCTS, MCTS_AIF}; the algorithm definition and pseudo-code are provided in Appendix~\ref{app:MCTS}. This approach incrementally builds a search tree by simulating possible action sequences and focusing computation on the most promising branches, considering EFE over policies. This yields three advantages over predefined or exhaustive policy sets: (1)~scalability, as it avoids combinatorial explosion in large action spaces; (2)~adaptability, as it dynamically prioritises policies relevant to the agent's current context; and (3)~efficiency, as it limits computation to policies that meaningfully differ in their predicted outcomes. 


To determine the most relevant policy at any time $t$, all policies are assessed through a softmax transformation (Eq.~\eqref{eq:P(pi)}):
\begin{equation}
    P(\pi) = \sigma (-\gamma G(\pi) - H)
    \label{eq:P(pi)}
\end{equation}
where $\sigma$ is the softmax function and $\gamma$ is a precision parameter (inverse temperature) controlling action stochasticity. Higher values of $\gamma$ bias the agent toward policies with minimal expected free energy, while lower values promote greater randomness in action selection.


A general issue with prediction over future steps through probability distribution, without relying on hierarchical structures, is that the further into the future the prediction lies, the more diluted the probabilities become. To address this, we introduce an inductive term $H$~\cite{inductive_AIF} in the policy evaluation. This term propagates pragmatic value from preferred states beyond the horizon range of the policies back to the current state, adding weight to actions leading toward those states. For a detailed derivation of this term, refer to Appendix~\ref{app:efe_terms}.

While the inductive prior $H$ propagates utility from distant states back to the current decision point, it does not fully resolve the finite horizon limitation inherent to AIF and MCTS. Limiting search depth means that rollouts do not reach terminal states, and standard guarantees on sample mean convergence do not strictly apply. In practice, this may reduce action selection accuracy for objectives separated by many state transitions from the current position. Hierarchical planning~\cite{ours_hierarchy} offers a principled extension for such scenarios; we leave this integration to future work.


The total EFE of a particular policy is the sum over its constituent time steps (Eq.~\eqref{eq:G(pi)}):
\begin{equation}
    G(\pi) = \sum_\tau G(\pi,\tau) 
    \label{eq:G(pi)}
\end{equation}
where $\tau$ indexes consecutive steps of the future policy.

The per-step EFE $G(\pi,\tau)$ balances epistemic values (information gain over future states and parameters) with pragmatic values (preference satisfaction), and additionally accounts for collision avoidance. In our formulation, the collision perception $c$ is decoupled from the observation $o$ for clarity. Although collision information is fundamentally derived from the observation, separating it allows the model to treat obstacle avoidance as an independent pragmatic constraint. In practice, $c_\tau$ is obtained by filtering 2D LiDAR range readings into a binary signal per action direction (obstacle present or absent). The model treats this as a generic depth feature; any sensor providing directional obstacle information (e.g., a depth camera or stereo pair) could serve as the source without altering the decision-making process. The probability $P(c_\tau)$ is evaluated against the preference $C_c$ of not encountering an obstacle, which has a major impact on policy viability. The full decomposition is given in Equation~\eqref{eq:efe_wt_c}:

\begin{equation}
\begin{aligned}
G(\pi,\tau) = \mathbf{E}_{Q_{\pi}} [\log Q(s_{\tau},p_{\tau}, A_p | \pi) - \log Q(s_{\tau},p_{\tau}, A_p | c_{\tau}, \pi) \\- \log P(c_{\tau}|C_c) - \log P(o_{\tau}|C_o) \\
- \log P(p_{\tau}|C_p)  - \log P(s_{\tau}|C_s)] \\
=\underbrace{\mathbf{E}_{Q_{\pi}} [ \log Q(s_{\tau},p_{\tau} | c_{\tau}, o_{\tau},\pi) - \log Q(s_{\tau},p_{\tau} | \pi)]}_\text{expected information gain on states (inference)} \\
-\underbrace{\mathbf{E}_{Q_{\pi}} [\log Q(A_p |s_{\tau}, p_{\tau}, c_{\tau}, o_\tau, \pi) - \log Q(A_p |s_{\tau}, p_{\tau}, \pi)]}_\text{expected information gain on parameters (learning)} \\
-\underbrace{\mathbf{E}_{Q_{\pi}} [ \log P(c_{\tau}|C_c) ]}_\text{expected collision}
- \underbrace{\mathbf{E}_{Q_\pi}[\log(P(o_{\tau}| C_o))]}_\text{utility term on observation}
\\
- \underbrace{\mathbf{E}_{Q_\pi}[\log(P(p_{\tau}| C_p))]}_\text{utility term on position} - \underbrace{\mathbf{E}_{Q_\pi}[\log(P(s_{\tau}| C_s))]}_\text{utility term on state}
\end{aligned}
\label{eq:efe_wt_c}
\end{equation}

The per-step EFE $G(\pi,\tau)$ decomposes into six terms summarised in Table~\ref{tab:efe_terms} and detailed in Equation~\eqref{eq:efe_wt_c}.

\begin{table}[htb!]
\centering
\caption{Short description of the per-step EFE terms.}
\begin{tabular}{lll}
\textbf{Term} & \textbf{Type} & \textbf{Role} \\ \hline
Info.\ gain on states & Epistemic & Reduces uncertainty over $s_\tau, p_\tau$ \\
Info.\ gain on parameters & Epistemic & Improves the position model $A_p$ \\
Expected collision & Pragmatic  & Penalises paths toward obstacles \\
Observation preference & Pragmatic & Attracts toward target observations \\
Position preference & Pragmatic & Attracts toward target positions \\
State preference & Pragmatic  & Attracts toward target states \\ \hline
\end{tabular}
\label{tab:efe_terms}
\end{table}

In Equation~\eqref{eq:efe_wt_c}, the expectation $\mathbb{E}_{Q_\pi}[\cdot]$ is taken over the predicted distribution of future states, positions, observations, and collisions under policy $\pi$. The position likelihood $A_p$ associates each state with its metric position; rather than maintaining a parametric distribution over $A_p$, the model stores positions directly and evaluates spatial consistency when predicting state transitions (see Section~\ref{sec:mapping} and Appendix~\ref{app:model_param}).

The expected information gain quantifies the anticipated shift in the agent's belief over the state from the prior (e.g., $Q(s_\tau|\pi)$) to the posterior (e.g., $Q(s_\tau| c_\tau, o_\tau,\pi)$) when pursuing a particular policy $\pi$. The utility terms assess the expected log probability of observing preferred outcomes under the chosen policy. This value measures the likelihood of the policy guiding the agent toward its preferences (e.g., avoiding collisions, reaching target observations, positions, or states). Rather than representing "reward" in the reinforcement learning sense, these terms measure coherence between predicted outcomes and the agent's prior preferences, encoded in the distributions $C_c$, $C_o$, $C_p$, and $C_s$.

\subsection{Learning and Map Expansion}
\label{sec:learning}

Having determined how the agent infers its state and selects actions, we now describe how it updates its internal parameters in light of inferred poses and current and expected observations. This learning step closes the perception-learning-action loop, enabling the agent to refine its world model so that it continues to explain sensory observations accurately.

The agent compares its current generative model, as defined by parameters governing pose likelihoods ($A_p$), observation likelihoods ($A_o$), and state transitions ($B_s$), against an expanded model incorporating newly predicted or observed states~\cite{ours_model}. 
Whether to expand the map is determined by comparing the free energy of the current model against that of an expanded model. This comparison is evaluated over the pose likelihood $A_p$ (Eq.~\eqref{eq:PvsP}~\cite{bayesian_model_reduc,surpervised_struct_learning}): if the expanded model achieves lower free energy ($\Delta F<0$), the map is expanded by adding the new state to $A_p$, and $B_p$, $B_s$ and $A_o$ are resized accordingly to match the new state dimensionality. The transition model $B_s$ and observation likelihood $A_o$ are additionally updated independently of map expansion: $B_s$ is refined via Dirichlet pseudo-counts whenever transitions are observed, and $A_o$ is updated when localisation confidence is high.

\begin{equation}
\Delta F = F[\hat{A_p}(\theta)] - F[A_p(\theta)]
\label{eq:PvsP}
\end{equation}

Here, $\hat{A_p}$ represents the updated pose likelihood model, and $\theta$ its parameters. If $\Delta F$ is negative (meaning the new model has lower free energy), the agent updates its internal structure $A_p$ to incorporate the newly predicted information.

When a new state is predicted, the pose model $A_p$ is expanded, triggering corresponding updates to the pose transition tensor $B_p$ and to all other state-related matrices. The updated observation model $A_o$ assigns uniform probabilities to unvisited states, reflecting initial uncertainty (high entropy, low $P(s_t)$). The transition model $B_s$ is also adapted to connect new states with existing ones. The Dirichlet pseudo-count update on state transition has learning rates depending on whether the trajectory was imagined or physically experienced, and whether it was deemed feasible or impossible. Concisely, an experienced transition reinforces the transition belief more than a predicted transition; inversely, a physically experienced obstruction reduces the transition belief more strongly than a predicted obstacle (details in Appendix~\ref{app:model_param} Table~\ref{tab:tran_lr}).

A similar expansion mechanism applies during inference when a novel sensory input is encountered. In this case, only $A_o$ is expanded, since the update pertains to the observation dimension rather than the state dimension.

By continuously applying this update process, the agent maintains an adaptable, self-consistent world model integrating both real and imagined experience. This ensures that exploration not only extends the map but also improves the accuracy of navigation when pursuing specific goals.

\subsection{Worked Example}
\label{sec:example}

To illustrate the perception-learning-action cycle concretely, consider the following scenario from exploration. At timestep $t$, the agent believes it occupies state $s_3$ with position $p_t = (2.1, 3.4)$~m in its local coordinate frame. The topological map currently contains one visited node and 5 nodes in total. The observation likelihood matrix $A_o$ is a $5 \times 5$ distribution associating the first node's stored panorama with its corresponding state. The rest of the matrix is filled with near-zero values (high uncertainty).

The agent evaluates candidate policies via MCTS (depth 10, 30 simulations) and selects action $a_t = \text{heading}_7$ (moving approximately $210^\circ$ relative to the robot's global orientation). After executing this motion, odometry reports a displacement of 0.6~m, yielding a predicted position $p_{t+1} = (1.6, 2.9)$~m.

Upon arrival, the agent captures a new $360^\circ$ panorama $o_{t+1}$ and computes SSIM scores against all stored observations. Suppose the highest SSIM score is 42\%, falling below the recognition threshold (set to 65\% in this work). Simultaneously, the predicted position $p_{t+1}$ lies within the 0.5~m influence radius of one unvisited node $s_5$. Together, the posterior belief concentrates on the new state: $Q(s_{t+1} = s_5) \approx 1$.
Considering the depth of the observation (obtained from LiDAR) and the influence radius of the current node, the model $A_p$ expands in all free directions beyond the existing nodes' influence radius from the current believed state $s_5$. New states $s$ are created to match the new dimensionality of $A_p$ in $B_p$. The transition matrix $B_s$ and likelihood $A_o$ expand in proportion to match the new map dimension. The transition matrix $B_s$ is updated via Dirichlet pseudo-counts to reflect the newly traversed edge $(s_3, a_t) \rightarrow s_5$ and the believed connections between nodes given by the current observation. 
\section{Results}

\label{sec:results}

We evaluate AIMAPP across simulated and real-world environments, assessing exploration efficiency, robustness to drift and environmental changes, and goal-reaching performance. Computational scalability is reported at the end of this section.

\subsection{Experimental Setup}
\label{sec:setup}

\paragraph{Environments}

Experiments were conducted in four simulated and three real-world environments. The simulated setups comprised three warehouses of increasing size: mini (36~m$^2$), small (80~m$^2$), and large (280~m$^2$), based on the Amazon Gazebo environment~\cite{warehouse}, featuring aisles, boxes, and industrial obstacles such as forklifts. A 175~m$^2$ house environment~\cite{house} without doors, containing kitchens, playrooms, and bedrooms, was also used. Both warehouses and the house included objects that are challenging for LiDAR-based detection, such as curved chairs and forklift legs. Real-world experiments were performed in three settings: (1)~a small, fully controlled 20.3~m$^2$ bedroom with drift-inducing flooring (carpet and wooden floor), (2)~a controlled 185~m$^2$ warehouse equipped with Qualisys motion-capture cameras for ground-truth odometry, and (3)~a large 325~m$^2$ sandy parking lot where moving cars occasionally altered the navigable space (navigation was paused while cars crossed). Environment layouts are provided in Appendix~\ref{app:envs}.

\paragraph{Baselines}

We compare AIMAPP against several exploration strategies that do not require pre-training:
\begin{itemize}
    \item \textbf{Frontier-based exploration}~\cite{frontiers}, using a 2D LiDAR and Nav2 SLAM~\cite{nav2_slam}.
    \item \textbf{GBPlanner}, an enhanced version of the 2021 DARPA SubT Challenge winner~\cite{Darpa_winners}, combining Voxblox~\cite{voxblox} with a topological map for 3D exploration planning. It operates with three cameras and two 3D LiDARs.
    \item \textbf{FAEL}~\cite{FAEL}, based on frontier logic, using 3D mapping via UFOMap~\cite{ufomap} and topological navigation. It relies on a 3D LiDAR.
    \item \textbf{Manual exploration}, in which a human operator teleoperated the robot while Nav2 SLAM cartographed the surroundings. The human has prior knowledge about the environment.
\end{itemize}
For fairness, all models operated with a maximum sensing range of 12 m regardless of sensor dimensionality. Because AIMAPP is a zero-shot learning agent, in the sense that it does not train over any prior data, we did not compare against learning-based methods~\cite{RECON,BYOL,viking,Nomad} requiring pre-training, which would have had prior knowledge of the environments or required a pre-mapping. Details about robots, sensors and each baseline configuration are provided in, respectively, Appendix~\ref{app:robots} and~\ref{app:other_models}.

\paragraph{Metrics}

We report the following metrics throughout this section. \textit{Coverage Efficiency} (CE) is defined as the area covered (in m$^2$) divided by the distance travelled (in m); time is excluded, as it depends on motion-planning parameters (wheel speed) rather than the decision-making process. \textit{Normalised Area Under the Coverage Curve} (nAUC) captures the progression of coverage relative to distance, with higher values indicating faster convergence to full coverage. \textit{Root Mean Square Error} (RMSE) over the $(x,y)$ axes quantifies positional drift with respect to ground-truth odometry. Additionally, we report the trajectory-length-normalised drift, computed as the absolute trajectory error per metre travelled (ATE/m), to enable comparison across runs of different lengths. For goal-reaching, we report the ratio of the agent's travelled path length to the optimal path computed by A* over the agent's topological graph (ideal path given agent knowledge), which assumes perfect knowledge of transitions (which implies that the A* can imagine passing through an impossible transition if it was deemed traversable or, instead, miss a shortcut) and an A* over an occupancy map of the environments (most ideal path). All metrics are averaged over a minimum of five runs per condition minimum, with standard deviations reported.

\paragraph{Protocol}

Across all scenarios, the agent begins each trial from a different initial position. Exploration results comprise five successful trials per environment per model. Failures and human interventions are documented in Appendix~\ref{app:human_int}. Drift measurements are conducted over five runs per model in the 185~m$^2$ warehouse using Qualisys cameras, exclusively for benchmarking (no navigation model has access to the ground-truth positioning). Goal-reaching is evaluated over 40 runs across all environments (simulated and real) with goals at various distances from the starting position; there is 17 goal-oriented real-world trials across three environments. In those goal-reaching experiments, the agent is assumed to know its starting position, thereby bypassing the initial re-localisation steps. This isolates the goal-reaching behaviour from the re-localisation behaviour, since re-localisation requires several preliminary steps in a recognisable region before goal-directed planning can begin. Re-localisation under the same generative model has been evaluated separately in a simplified 2D MiniGrid setting without realistic observations~\cite{ours_model}. Since our model does not construct a metric map, we associated LiDAR range measurements with internal state representations to evaluate spatial coverage between models.

\subsection{Exploration}
\label{sec:exploration}

\paragraph{Simulation Results}

\begin{figure}[!ht]
\centering
\begin{minipage}{0.35\columnwidth}
\centering
\includegraphics[width=\linewidth]{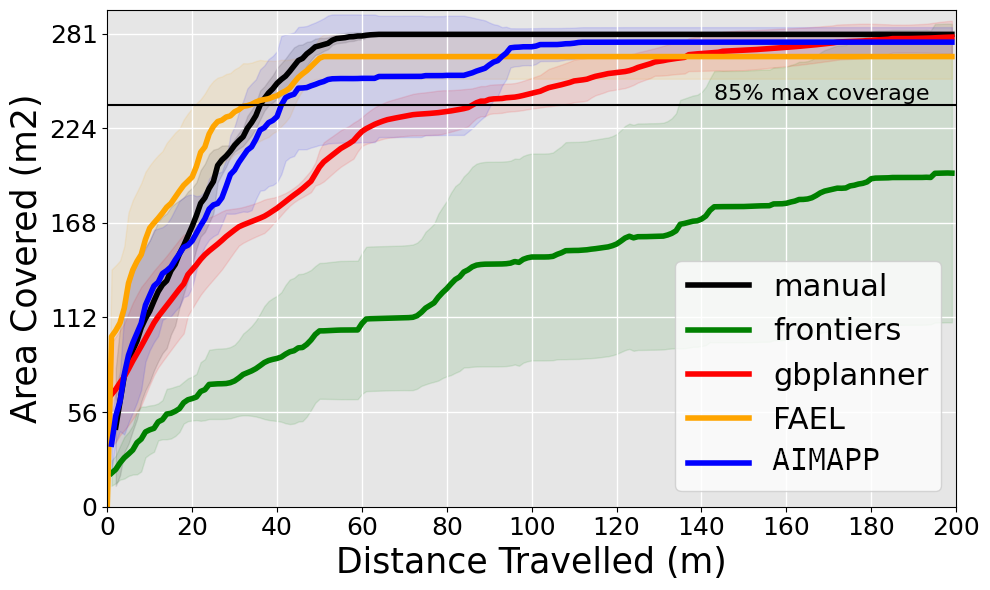}\\
(a) Large warehouse
\end{minipage}%
\begin{minipage}{0.35\columnwidth}
\centering
\includegraphics[width=\linewidth]{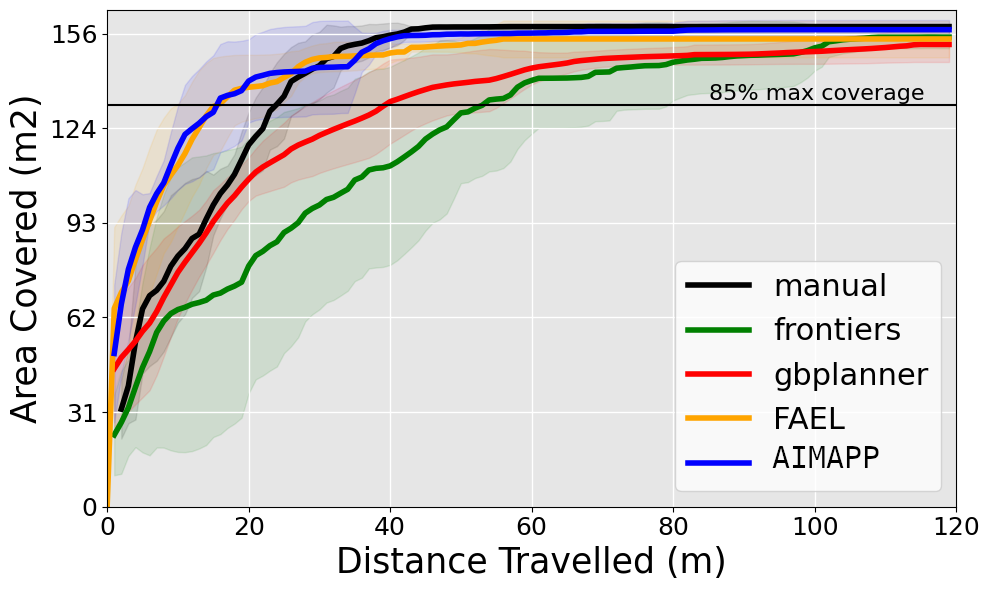}\\
(b) Home
\end{minipage}%
\begin{minipage}{0.35\columnwidth}
\centering
\includegraphics[width=\linewidth]{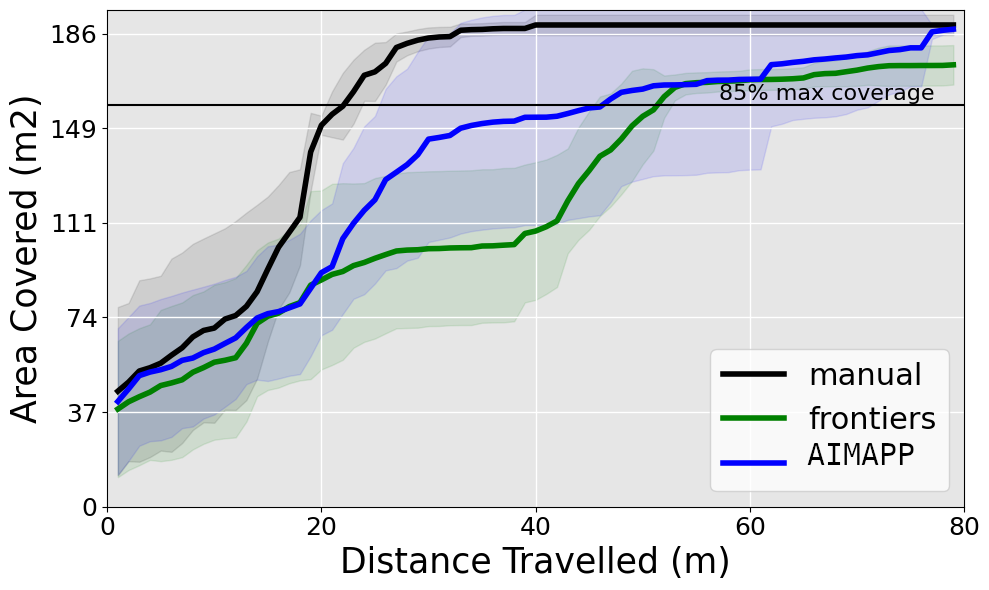}\\
(c) Real warehouse
\end{minipage}%
\caption{Coverage efficiency considering the agent's travelling distance: (a) large simulated warehouse, (b) simulated home, and (c) real warehouse (Frontiers and AIMAPP only).}
\label{img:coverage_2envs}
\end{figure}

The average coverage efficiency of each model in the two largest simulated environments, measured as explored area relative to distance travelled and averaged over five successful trials, is reported in Figure~\ref{img:coverage_2envs} and Table~\ref{tab:metric_explo}. Manual explorations are treated as a near-optimal reference, given that the human operator has a general understanding of the layout while navigating.

Across all simulated environments (including the additional environments in Appendix~\ref{app:explo_paths}, Table~\ref{tab:CE_app}), AIMAPP achieves a coverage efficiency of 90.3\% relative to manual exploration. In the large warehouse, AIMAPP attains a CE comparable to FAEL and substantially above GBPlanner and Frontiers. In the house environment, AIMAPP and FAEL both exceed manual performance, while GBPlanner and Frontiers lag. The manual navigation is outperformed due to the humans conscientiously visiting all areas while the models accept ignoring some small unknown parcels and focus on bigger unknown areas, leaving those small unvisited parcels for later. The nAUC values follow a similar pattern: AIMAPP and FAEL achieve the highest normalised coverage progression, while GBPlanner and Frontiers are consistently lower.

A coverage plateau at approximately 90\% is observed in AIMAPP. This arises because the model prioritises updating nearby unvisited states already represented in its internal graph rather than extending exploration beyond the current detection range, resulting in node refinements without proportional increases in spatial coverage. In smaller environments ($\leq$80~m$^2$), all methods converge to similar performance levels (see Appendix~\ref{app:explo_paths}).

Frontiers performed worst overall, as it repeatedly attempts to reach unreachable frontier cells, wasting significant travel distance. The GBPlanner method prioritises safe exploration paths, often performing redundant back-and-forth movements before expanding to new zones.

\begin{table}[ht!]
\caption{Exploration efficiency metrics across environments. 
CE: Coverage Efficiency (m$^2$/m), nAUC: Normalised Area Under Coverage Curve. Values are mean~$\pm$~std over 5 runs.}
\begin{tabular}{llcc}
\hline
\textbf{Env.}          & \textbf{Model} & CE            & nAUC          \\ \hline
\multirow{5}{*}{\begin{tabular}[c]{@{}l@{}}Simulated\\ Large Warehouse\end{tabular}}              & Manual         & 4.25$\pm$0.43 & 0.55$\pm$0.02 \\
                       & AIMAPP           & 4.02$\pm$1.24 & 0.62$\pm$0.06 \\
                       & Frontiers      & 0.60$\pm$0.35 & 0.53$\pm$0.15 \\
                       & FAEL           & 4.65$\pm$0.45 & 0.77$\pm$0.07 \\
                       & GBPlanner      & 1.13$\pm$0.17 & 0.56$\pm$0.04 \\ \hline
\multirow{5}{*}{\begin{tabular}[c]{@{}l@{}}Simulated House \end{tabular}} & Manual         & 4.02$\pm$0.74 & 0.70$\pm$0.08 \\
                       & AIMAPP           & 4.89$\pm$1.10 & 0.87$\pm$0.10 \\
                       & Frontiers      & 2.18$\pm$0.15 & 0.77$\pm$0.17 \\
                       & FAEL           & 5.23$\pm$1.90 & 0.79$\pm$0.06 \\
                       & GBPlanner      & 3.70$\pm$0.15 & 0.50$\pm$0.04 \\ \hline
\multirow{5}{*}{\begin{tabular}[c]{@{}l@{}}Real Warehouse\end{tabular}}              & Manual         & 4.89$\pm$0.53 & 0.65$\pm$0.09 \\
                       & AIMAPP           & 2.68$\pm$0.40 & 0.70$\pm$0.10 \\
                       & Frontiers      & 2.11$\pm$0.52 & 0.62$\pm$0.15
\end{tabular}
\label{tab:metric_explo}
\end{table}

\paragraph{Real-world Results}

In real-world trials, only AIMAPP and Frontiers could be deployed, as FAEL and GBPlanner require sensors unavailable on the physical platform. Both were tested from five different starting positions in the warehouse environment, which contained two long aisles and a large open area. Results are reported in Figure~\ref{img:coverage_2envs}(c) and Table~\ref{tab:metric_explo}. AIMAPP was also tested in two additional environments, a small bedroom (~20.3m$^2$) and a dynamic parking lot (325~m$^2$), in which Frontiers consistently failed before completing the trial owing to excessive odometric drift on wooden and sandy flooring; failure modes are documented in Appendix~\ref{app:human_int}.

AIMAPP consistently achieves faster coverage than Frontiers. The difference stems from strategy: AIMAPP moves efficiently between unexplored regions, while Frontiers repeatedly revisits the same aisles due to attraction toward unreachable frontier cells without planning through alternative paths. Performance varies with initial placement, reflecting the influence of warehouse geometry.

Frontiers performed comparatively better in the real warehouse than in the simulated house of similar dimensions, likely because long aisles are more forgiving for its greedy strategy. Conversely, AIMAPP performed slightly less efficiently in the real warehouse than in simulation, as LiDAR detection errors occasionally caused it to attempt reaching unreachable goals (behind an aisle), introducing inefficiencies in motion planning. Exploration results for the parking lot and bedroom are provided in Appendix~\ref{app:explo_paths}. In the parking, navigation was paused whenever a car moved through the space; we did not evaluate the obstacle avoidance performance of the motion planner, as it is not part of the proposed model. 

\paragraph{Success Rates}

Table~\ref{tab:success_rate} reports the success rate for obtaining five successful exploration trials per environment. GBPlanner achieved the highest reliability (87\%), followed by AIMAPP (79\%), Frontiers (52\%), and FAEL (48\%). Failures in AIMAPP occurred primarily when localisation drift in large unexplored areas produced a non-recoverable map overlap, or when the robot flipped after colliding with an obstacle undetected by the LiDAR. GBPlanner failed only when undetected obstacles caused the robot to flip. Frontiers frequently stalled by persisting in attempts to reach unreachable goals, while FAEL exhibited failures due to mismatches between topological node creation and obstacle detection in its 3D mapping. FAEL and GBPlanner were not evaluated in real-world settings due to sensor requirements incompatible with the available hardware. Detailed intervention logs and per-environment breakdowns are provided in Appendix~\ref{app:human_int}.

\begin{table}[h!tb]
\vspace{-5mm}
\centering
\caption{Percentage of exploration success rate for each model over each environment.}
\begin{tabular}{l|cccc}
\begin{tabular}[c]{@{}l@{}}External \\ intervention\end{tabular} & AIMAPP  & FAEL     & Gbplanner  & Frontiers \\ \hline
Home            & 0.71 & 0.45 & 0.83 & 0.63        \\ \hline
Big warehouse   & 1 & 0.5 & 1 & 0.83       \\ \hline
Small warehouse & 1 & 0.63 & 0.87 & 0.83        \\ \hline
Mini warehouse  & 0.83 & 0.83 & 1 & 0.83       \\ \hline
Real home       & 0.6 & x & x & 0.09       \\ \hline
Real parking    & 0.75 & x & x & 0.00       \\ \hline
Real warehouse  & 0.63 & x & x & 0.45    
\end{tabular}
\label{tab:success_rate}
\vspace{-5mm}
\end{table}

\subsection{Robustness}
\label{sec:robustness}

\paragraph{Obstacle Adaptation}

\begin{figure}[!ht]
\centering    
\includegraphics[width=0.51\columnwidth]{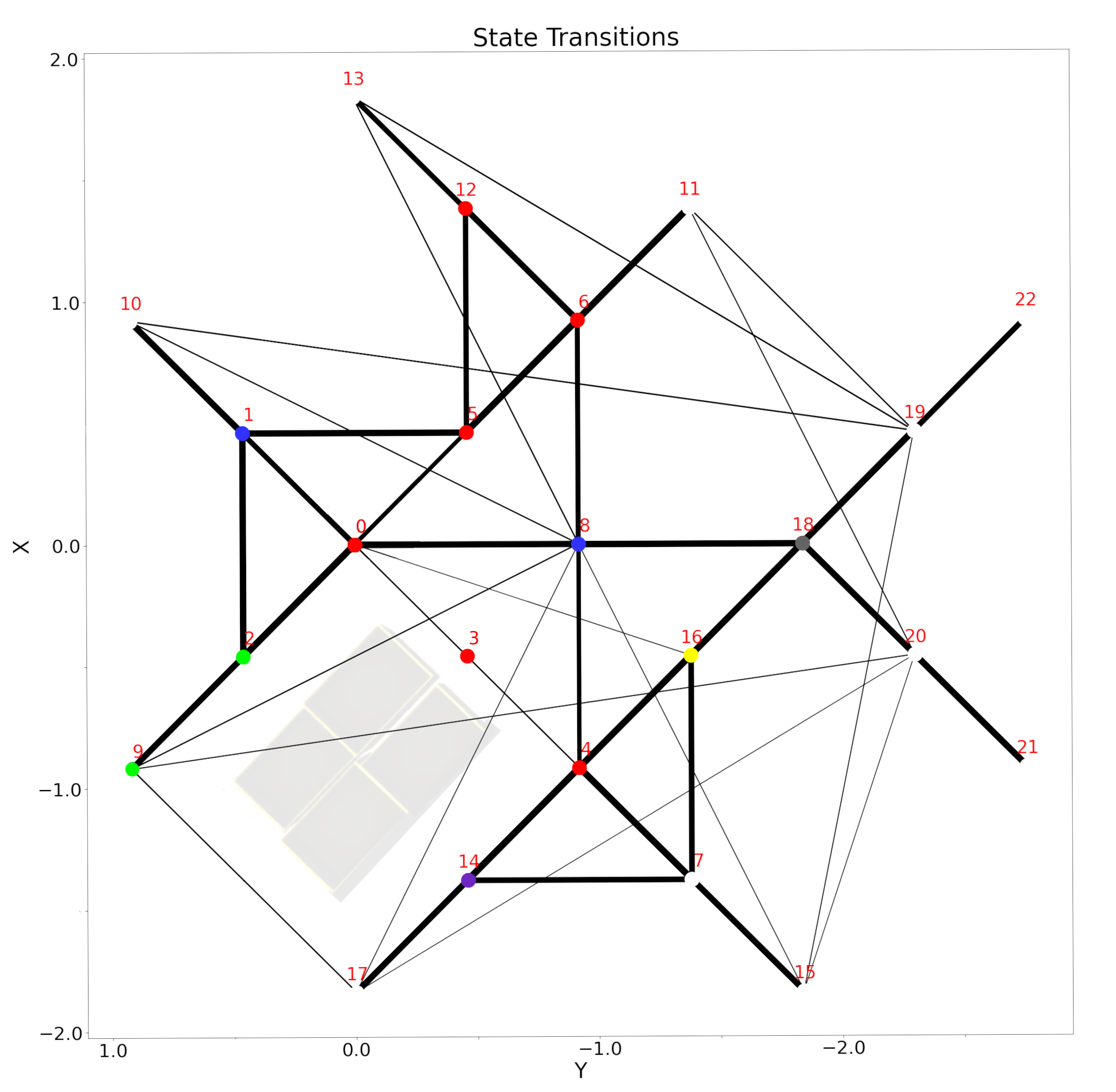}%
\label{img:ob_before}
\includegraphics[width=0.45\columnwidth]{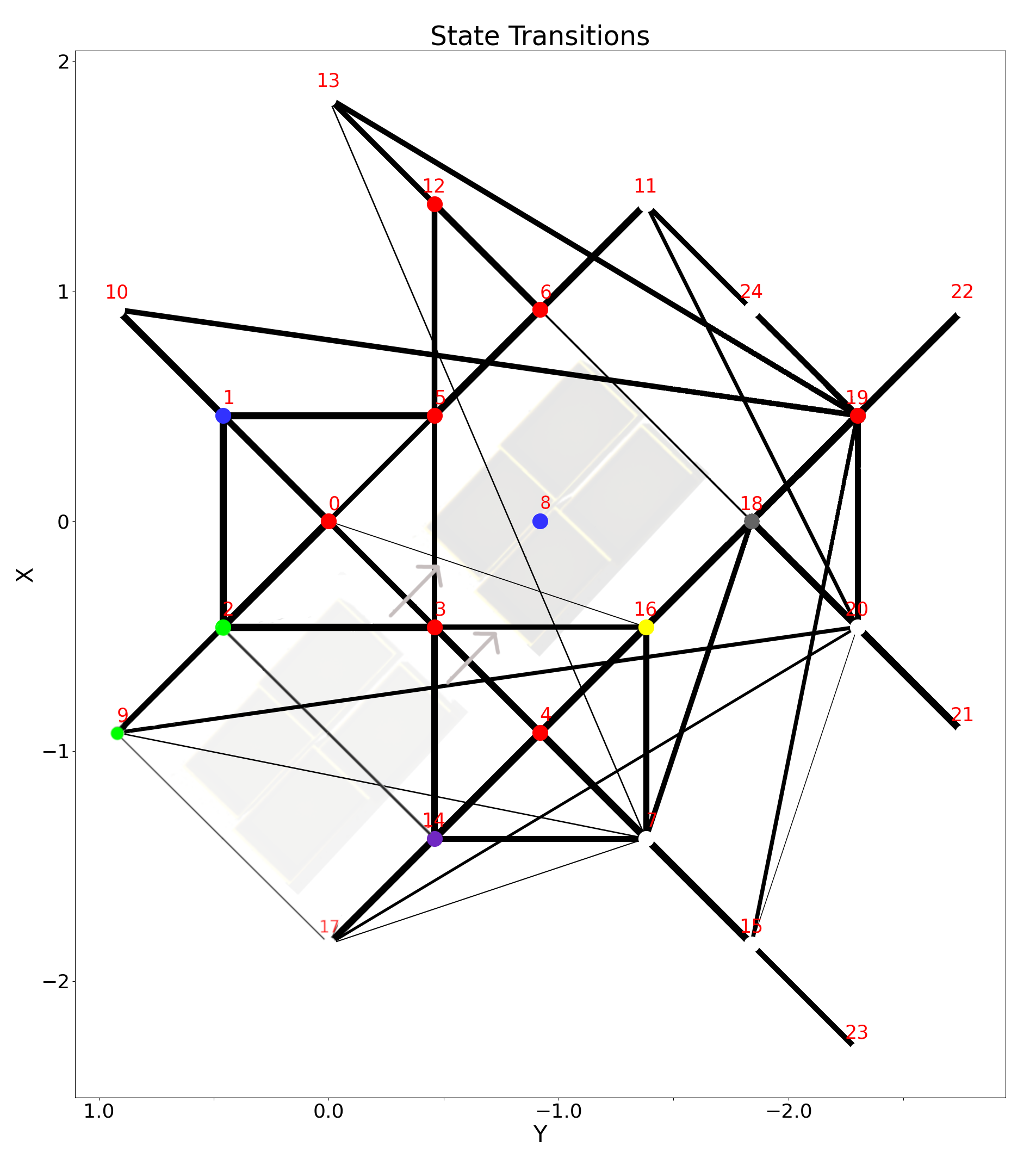}%
\label{img:ob_after}
\caption{Adaptation to a displaced obstacle: a box was moved from (a) $(-1,0)$ to (b) $(0,-1)$ during exploration of a 25~m$^2$ environment. Over twenty steps around the box, the map weakens impossible transitions and reinforces previously improbable links. New nodes are created if pertinent.}
\label{img:ob_adaptation}
\end{figure}

AIMAPP dynamically adapts to changes in the environment during navigation by continuously updating its internal topological map. This process is illustrated in Figure~\ref{img:ob_adaptation} in a small-scale (25~m$^2$) environment where the position of a box is displaced from $(-1,0)$ to $(0,-1)$ during exploration. As the agent moves and attempts transitions, it incrementally weakens the likelihood of inaccessible paths and reinforces the plausibility of reachable ones. Failed attempts to move toward an obstructed location trigger significant updates to the agent's belief structure, while successful transitions reinforce existing connections.

In Figure~\ref{img:ob_after}, state 8 (position $(0,-1)$) becomes inaccessible after being blocked by the displaced box. All transitions leading to it are suppressed as the agent gathers evidence by navigating around the obstacle over approximately twenty steps. Numerical details regarding the belief update mechanism and associated learning rates are provided in Appendix~\ref{app:model_param}.

\paragraph{Drift Measurement}

\begin{figure}[!htb]
\centering
\includegraphics[width=0.45\linewidth]{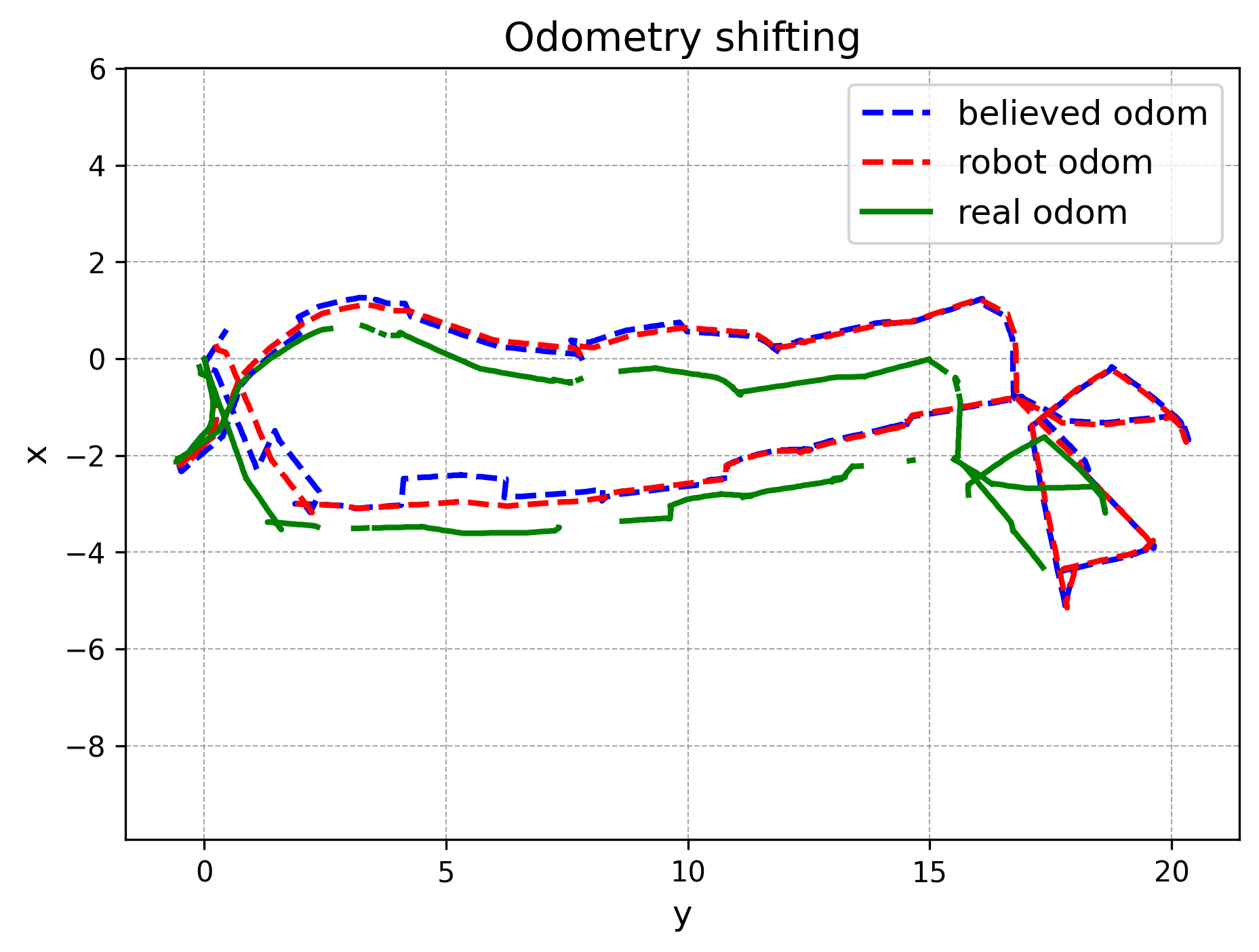}
\caption{Drift comparison in the $(x,y)$ plane between AIMAPP's internal belief (model odometry), the robot's onboard sensor odometry, and the ground-truth trajectory measured with Qualisys. Missing segments correspond to gaps in ground-truth coverage.}
\label{fig:drift}
\vspace{-5mm}
\end{figure}

To quantify drift under controlled conditions, we conduct experiments in the 185~m$^2$ warehouse equipped with Qualisys motion-capture cameras providing ground-truth odometry. These cameras are exclusively used for benchmarking and are not available to any navigation model. The trajectories estimated by AIMAPP (believed odom), the robot's onboard sensor odometry (robot odom), and the ground truth (real odom) are qualitatively compared in Figure~\ref{fig:drift}. Missing segments correspond to gaps in Qualisys coverage.



Table~\ref{tab:RMSE} reports both the absolute RMSE on the $(x,y)$
axes and the trajectory-length-normalised drift (ATE/m), averaged
across five successful runs. In absolute terms, AIMAPP and
Frontiers achieves comparable mean RMSE, but AIMAPP exhibits
substantially lower variance, indicating consistently reliable trajectories where Frontiers occasionally diverges. The manual baseline was obtained over shorter trajectories at operator-chosen speeds, providing less opportunity for absolute drift to accumulate.

\begin{table}[htb!]
\centering
\caption{RMSE over $x$ and $y$, and ATE/m, averaged over 5 runs per model.}
\begin{tabular}{l|cc|c|c}
 & \multicolumn{2}{c|}{AIMAPP} & Frontiers & Manual \\
 & model & sensor & sensor & sensor \\ \hline
RMSE (x,y) & $1.83 \pm0.77$ & $1.65 \pm0.79$ & $1.68\pm 2.00$ & $1.40 \pm0.25$ \\
ATE/m & \multicolumn{1}{l}{$0.20 \pm 0.10$} & \multicolumn{1}{l|}{$0.21 \pm 0.10$} & \multicolumn{1}{l|}{$0.58 \pm 0.17$} & \multicolumn{1}{l}{$0.06 \pm 0.02$}
\end{tabular}
\label{tab:RMSE}
\end{table}

Once the trajectory length is factored out, AIMAPP achieves substantially
lower per-metre drift than Frontiers (ATE/m), approximately three times lower. We attribute Frontiers' higher per-metre drift to its greedy exploration strategy: repeated attempts to reach unreachable frontier cells and redundant backtracking accumulate odometric error without proportional spatial progress, inflating the per-metre error. 

The near-equivalence of the AIMAPP model and sensor rows confirms that the belief layer tracks sensor odometry closely on the metric trajectory. It does not continuously correct drift, but corrects it opportunistically when stored panoramic observations are recognised. In the case of exploration, this mechanism does not occur when navigating in previously unexplored locations. The manual baseline achieves the lowest per-metre drift, reflecting smooth, efficient trajectories with minimal redundant motion.

AIMAPP's drift advantage, therefore, combines two properties: lower per-metre error than the autonomous baseline (Frontiers), and lower variance in absolute RMSE, with no catastrophic failures. On surfaces where Nav2's occupancy mapping becomes unusable (sandy or carpeted
flooring), AIMAPP continues operating while Frontiers fails outright.

\paragraph{Sensor Failure and Severe Drift}

A more pronounced distinction emerges in environments prone to severe drift. In the parking lot (uneven, sandy terrain) and the bedroom (mixed wood and carpet flooring), wheel slippage caused large odometry errors that could not be quantified. Under these conditions, AIMAPP's belief-driven map gradually lost metric alignment with the ground truth, yet the agent remained operational: it could continue exploring and reliably reach goal observations, provided that the goal was not estimated to lie beyond a wall. The system also tolerates temporary sensor failures; when odometry, camera, or LiDAR streams were restarted after a failure, the agent resumed operation without requiring reinitialisation.

By contrast, Frontier-based exploration, which relies on Nav2, fails under these conditions: once drift is accumulated, its occupancy maps become inconsistent and unusable, and restarting the odometry is not an option for recovery. Consequently, no quantitative comparison with Frontiers was possible in the bedroom or parking lot scenarios.

\subsection{Goal-Reaching}
\label{sec:goal_reaching}

\paragraph{Goal-Directed Navigation Strategy}

In our framework, goal-directed navigation emerges from the same Active Inference principles that drive exploration. While exploration favours previously unobserved states for their potential information gain, goal pursuit biases the agent toward states likely to generate a desired observation provided as input by the user. In these experiments, visual observations seen during exploration were given as the preferred observation $C_o$, with a weight of 10 on the pragmatic value to encourage the agent toward the goal over further exploration. All tests were conducted in partially explored environments where the goal had been observed during the prior exploration phase, to isolate goal-reaching behaviour from exploration.

This preference only influences behaviour if the agent can imagine a feasible trajectory toward the relevant states during planning. Far-removed high-utility states that do not appear in the planning horizon of the MCTS would not guide navigation. To address this, we employ an inductive prior $H$ (formally introduced in Equations~\eqref{eq:P(pi)} and~\eqref{eq:H(n)}), which propagates utility along sequences of plausible actions leading to the goal. This mechanism allows the agent to evaluate paths rather than isolated states, effectively assigning intermediate states a utility proportional to their potential to reach the target observation, even outside its planning horizon.

\begin{figure*}[!htb]
\centering
\begin{minipage}{0.30\linewidth}
\centering
\includegraphics[width=\linewidth]{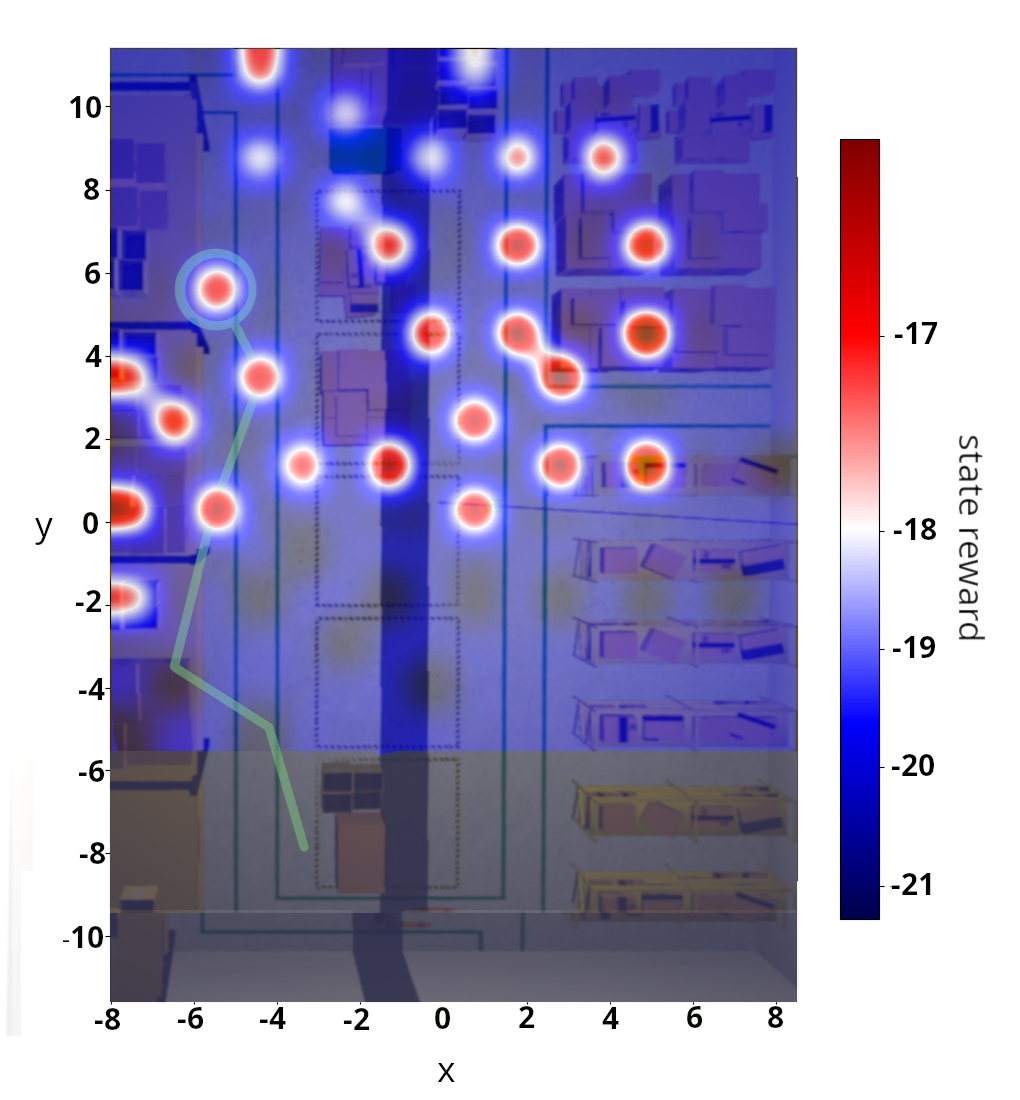}\\
(a) Step 0
\end{minipage}%
\hfill
\begin{minipage}{0.32\linewidth}
\centering
\includegraphics[width=\linewidth]{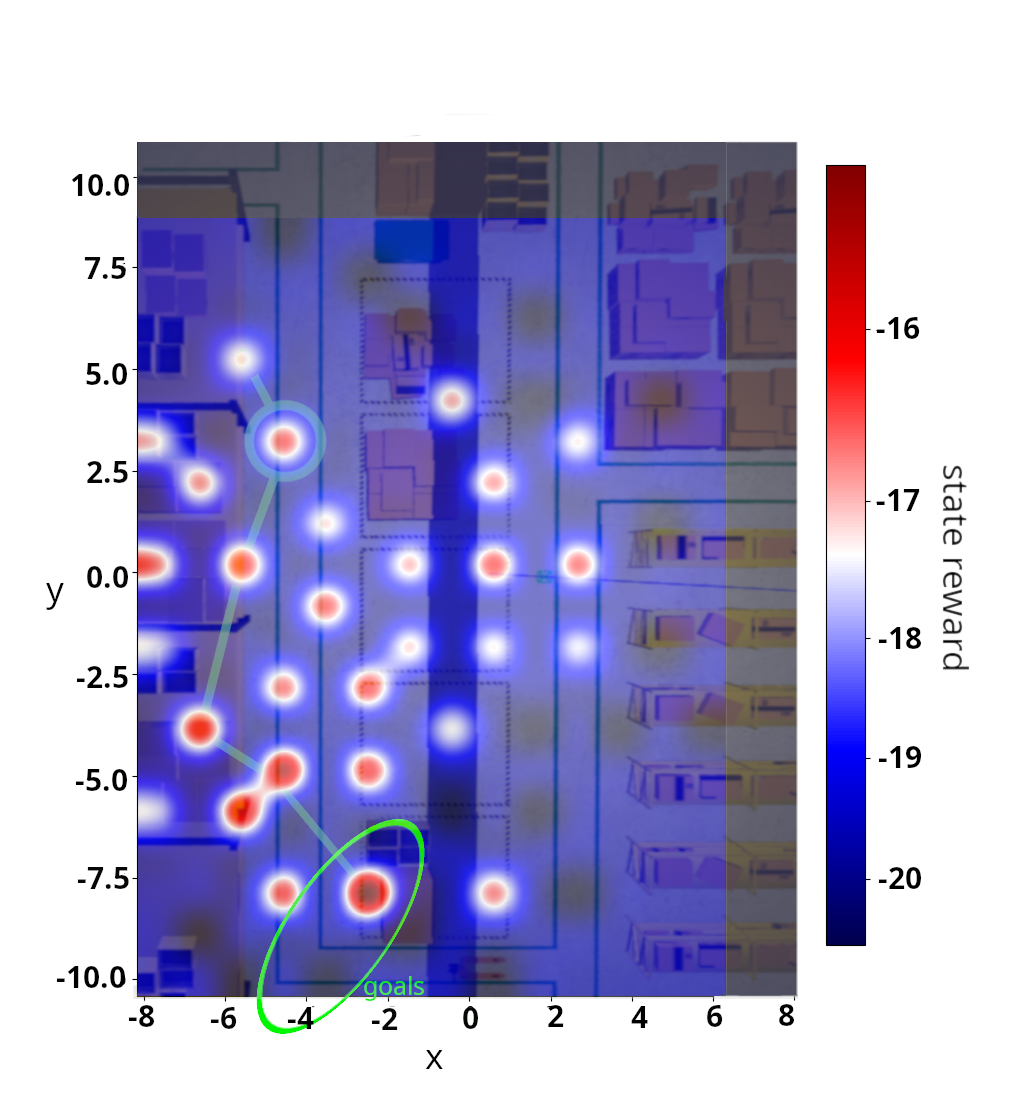}\\
(b) Step 1
\end{minipage}%
\hfill
\begin{minipage}{0.30\linewidth}
\centering
\includegraphics[width=\linewidth]{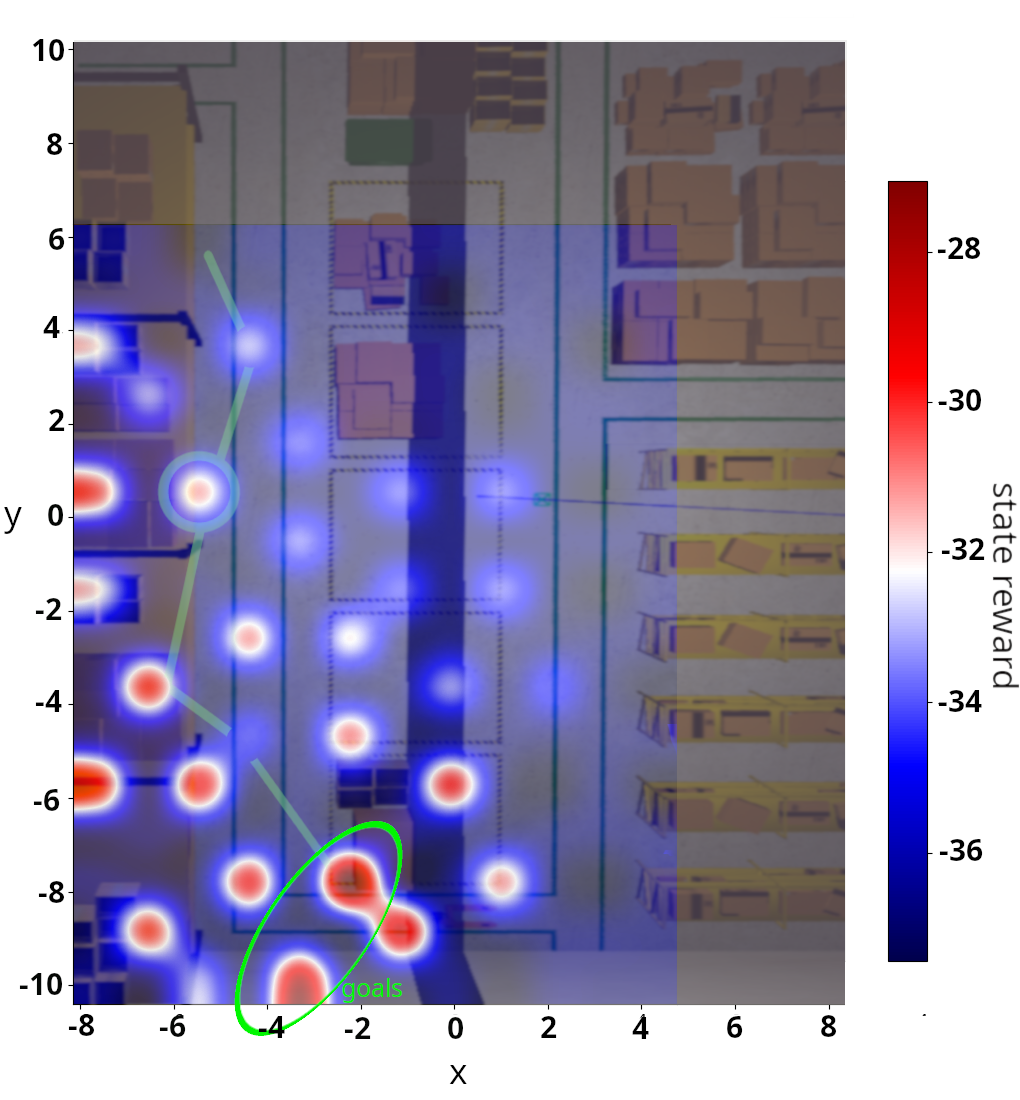}\\
(c) Step 2
\end{minipage}%
\\[2ex]
\begin{minipage}{0.32\linewidth}
\centering
\includegraphics[width=\linewidth]{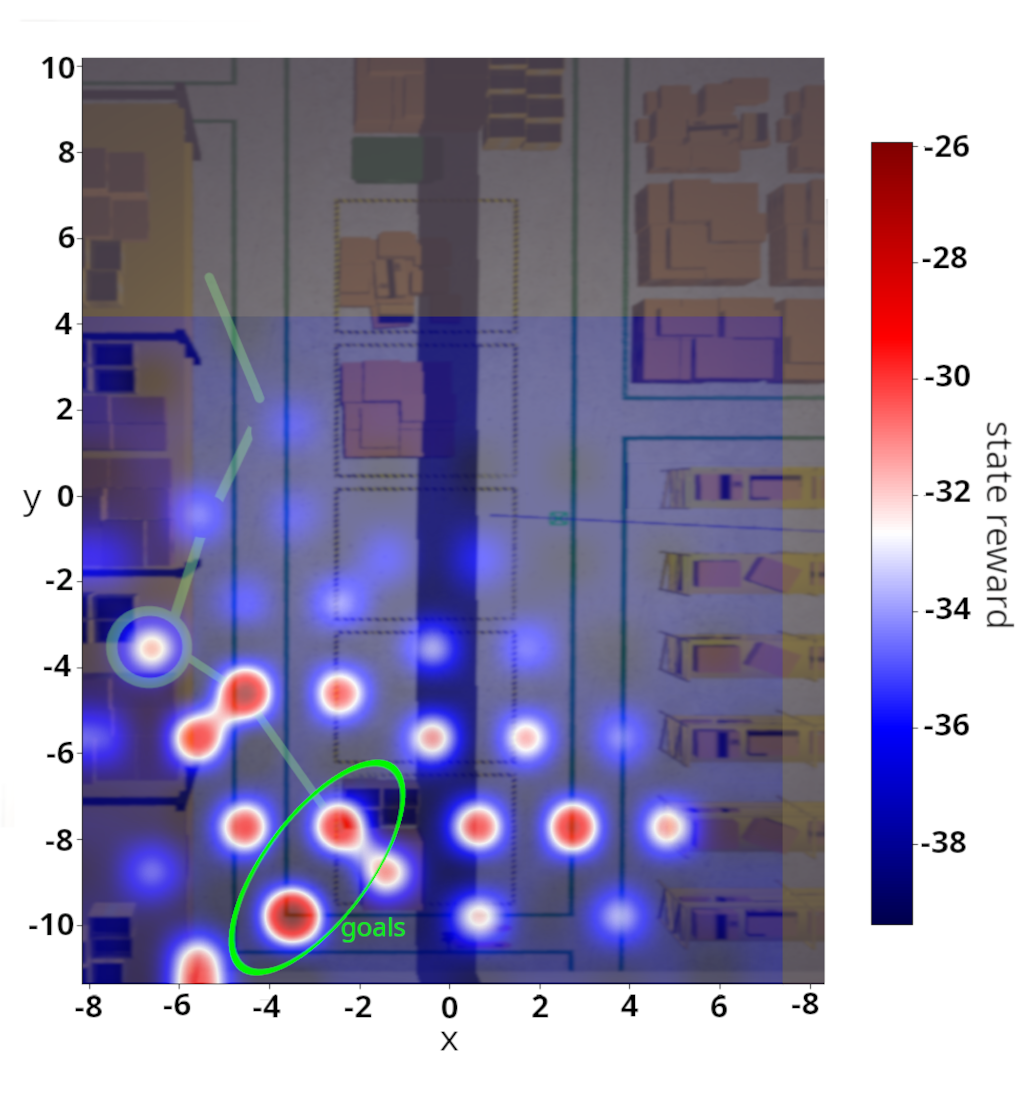}\\
(d) Step 3
\end{minipage}%
\hfill
\begin{minipage}{0.32\linewidth}
\centering
\includegraphics[width=\linewidth]{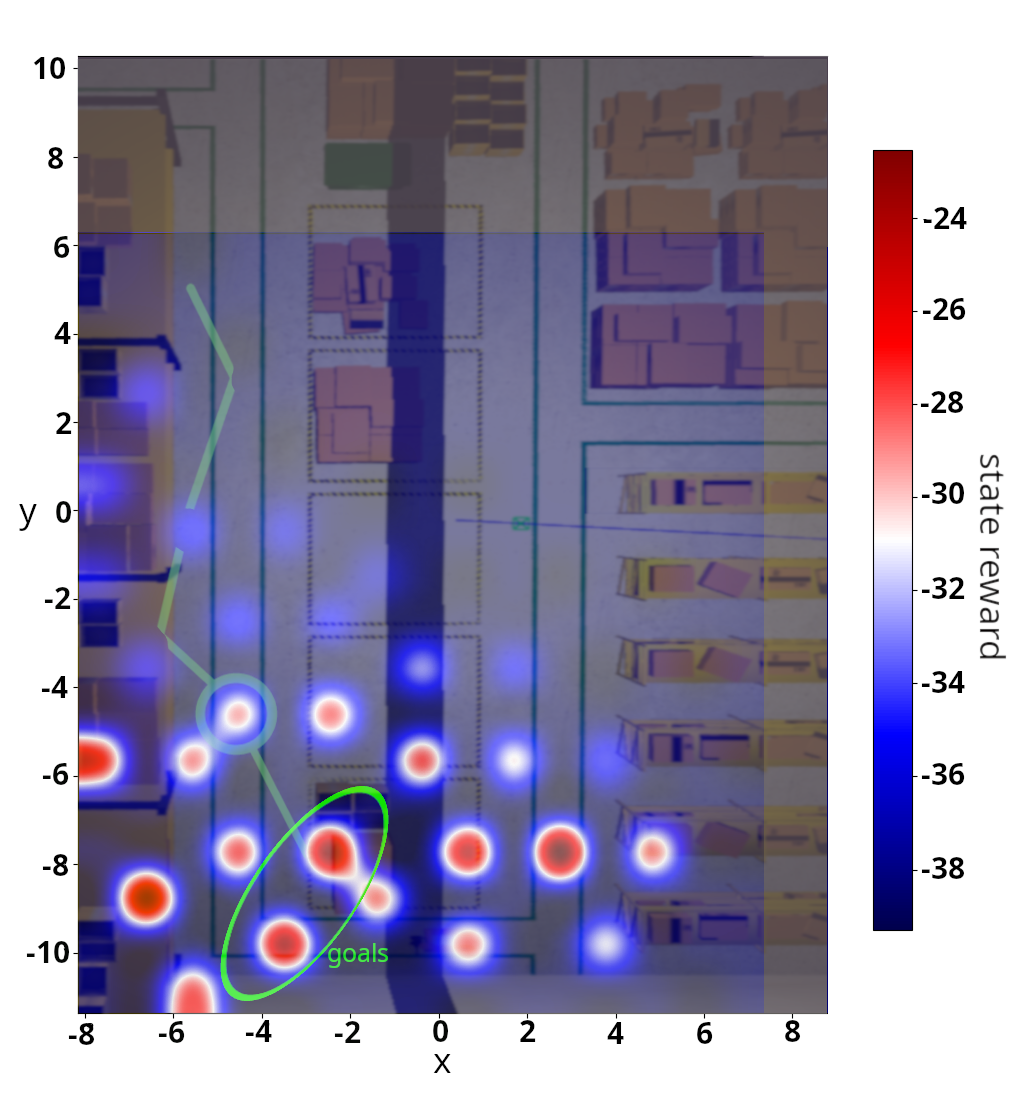}\\
(e) Step 4
\end{minipage}%
\\[2ex]
\caption{Utility heatmaps over the agent's topological map during MCTS planning toward a goal observation held by two states (circled neon green). Higher values indicate more attractive locations. The agent reached the goal in five steps from the starting pose (circled light green in step~0). The full trajectory is shown in dark green, with the current position circled. In step~0, the agent is attracted in the correct direction before it can directly imagine the goal state.}
\label{img:heatmap_goals}
\end{figure*}

This process is illustrated in Figure~\ref{img:heatmap_goals} in the simulated large warehouse. Heatmaps show the evolving utility of paths over five planning steps, with deeper red showing higher attractiveness. Initially (step~0), the agent (light green circle) is still biased in the correct general direction despite the goal states (circled in neon green) not yet being directly imagined from the current position. Two adjacent states correspond to the desired observation given to the agent; thus, there are two circled nodes. The executed path in light green is shown over all steps to help visualisation over time. As the agent transitions through steps 1 to 4, intermediate states leading toward the goal increase in utility, while states in the opposite direction are progressively suppressed. At each step, the agent simultaneously updates its positional belief through re-localisation and evaluates future trajectories, maintaining flexibility in response to newly encountered information.

\paragraph{Quantitative Goal-Reaching Performance}

To quantitatively assess goal-directed navigation, we compared the executed path length against two baselines: 1) the ideal path through AIMAPP's topological map as constructed during prior exploration, and 2) an A* path computed over the occupancy grid of each environment. We report path lengths in metres and their three-way decomposition as Path Length Ratios (PLR) in Table~\ref{tab:goal_reaching_decomposition}. The executed path against A* on a complete occupancy map, and against AIMAPP's ideal transition planning (as we would expect the model to behave). Moreover, AIMAPP's ideal transition plan is compared to A* to analyse how good AIMAPP can be, at its best. In all goal-reaching runs, the agent was assumed to know its starting position to bypass re-localisation, isolating goal-reaching from re-localisation behaviour.

\begin{table}[ht!]
\centering
\caption{Goal-reaching path efficiency across 40 runs. PLR: Path Length Ratio (executed / reference).}
\begin{tabular}{lcc}
\hline
\textbf{Comparison} & \textbf{PLR (mean $\pm$ std)} & \textbf{Median} \\ \hline
Executed vs. A* (true graph)        & $1.260 \pm 0.202$ & $1.240$ \\
Executed vs. AIMAPP plan            & $1.163 \pm 0.160$ & $1.186$ \\
AIMAPP plan vs. A* (agent graph)    & $1.085 \pm 0.111$ & $1.049$ \\ \hline
\end{tabular}
\label{tab:goal_reaching_decomposition}
\end{table}

\begin{figure}[!htb]
    \centering
    \begin{minipage}{0.4\linewidth}
        \centering
        \includegraphics[width=\linewidth]{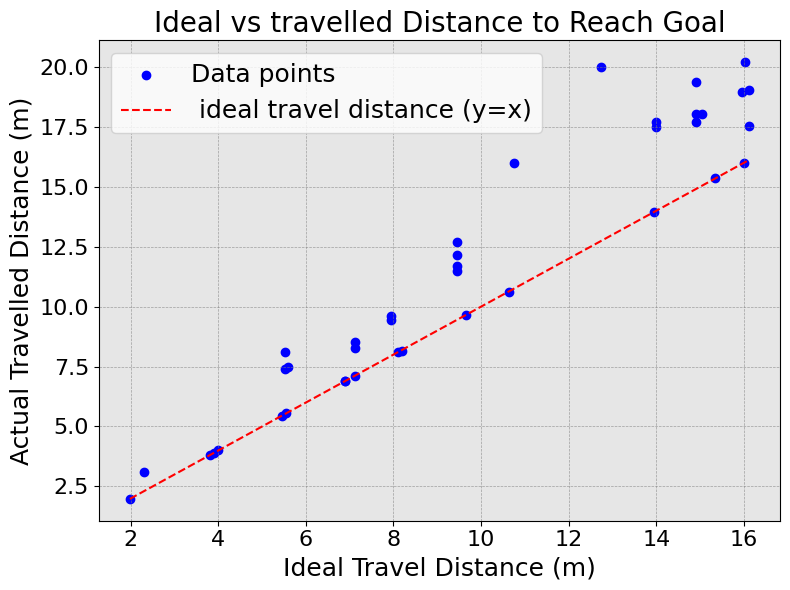}\\
        (a) Executed vs.\ A* distance per run
    \end{minipage}\hfill
    \begin{minipage}{0.58\linewidth}
        \centering
        \includegraphics[width=\linewidth]{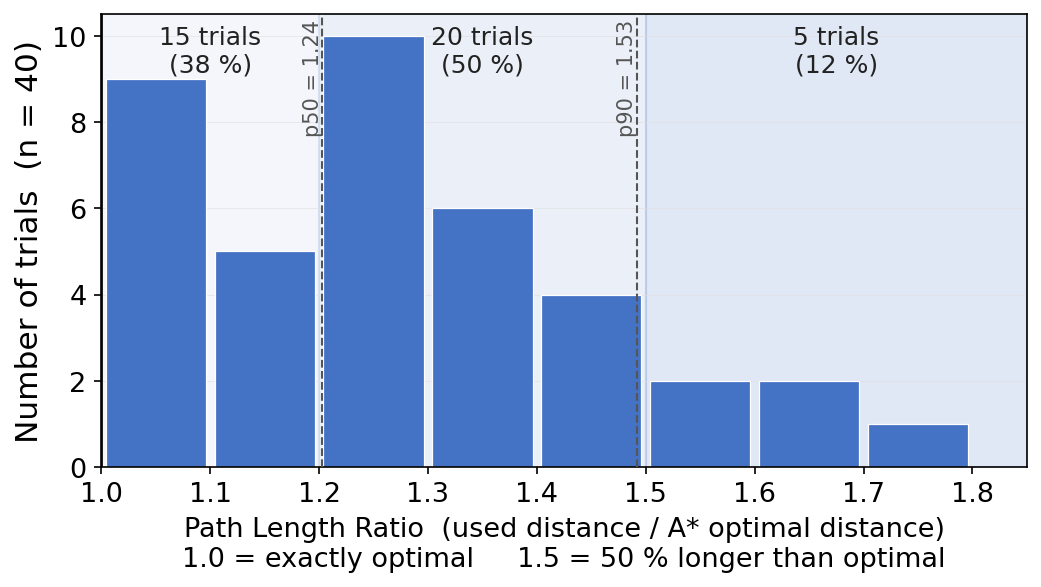}\\
        (b) PLR distribution across runs
    \end{minipage}
    \caption{Goal-reaching path efficiency over 40 runs across all environments. (a)~Executed distance against ideal AIMAPP distance per run. (b)~Distribution of the Path Length Ratio (executed~/~A*); 38\% of trials lie within 20\% of A*, 50\% between 20\% and 50\% over A*, and 12\% over 50\%, with median PLR~$=1.24$.}
    \label{fig:goals_average_results}
\end{figure}

We conducted 40 goal-reaching runs across all environments with goals at varying distances from the starting position. The end-to-end PLR decomposes approximately multiplicatively: $1.260 \approx 1.085 \times 1.163$. Most of the inefficiency arises when the agent encounters transitions previously believed feasible but actually in an obstacle and must replan. Figure~\ref{fig:goals_average_results}(a) shows the executed distance against the ideal travelled distance (in AIMAPP topological graph) for each run, and Figure~\ref{fig:goals_average_results}(b) shows the distribution of PLR across runs. The agent reached the goal within 20\% of the A* optimum in 38\% of trials, within 50\% in 88\% of trials, and exceeded 50\% in 12\% of trials, with a median PLR of $1.24$. 

\begin{figure*}[!ht]
\centering
\begin{minipage}{0.30\linewidth}
\centering
\includegraphics[width=\linewidth]{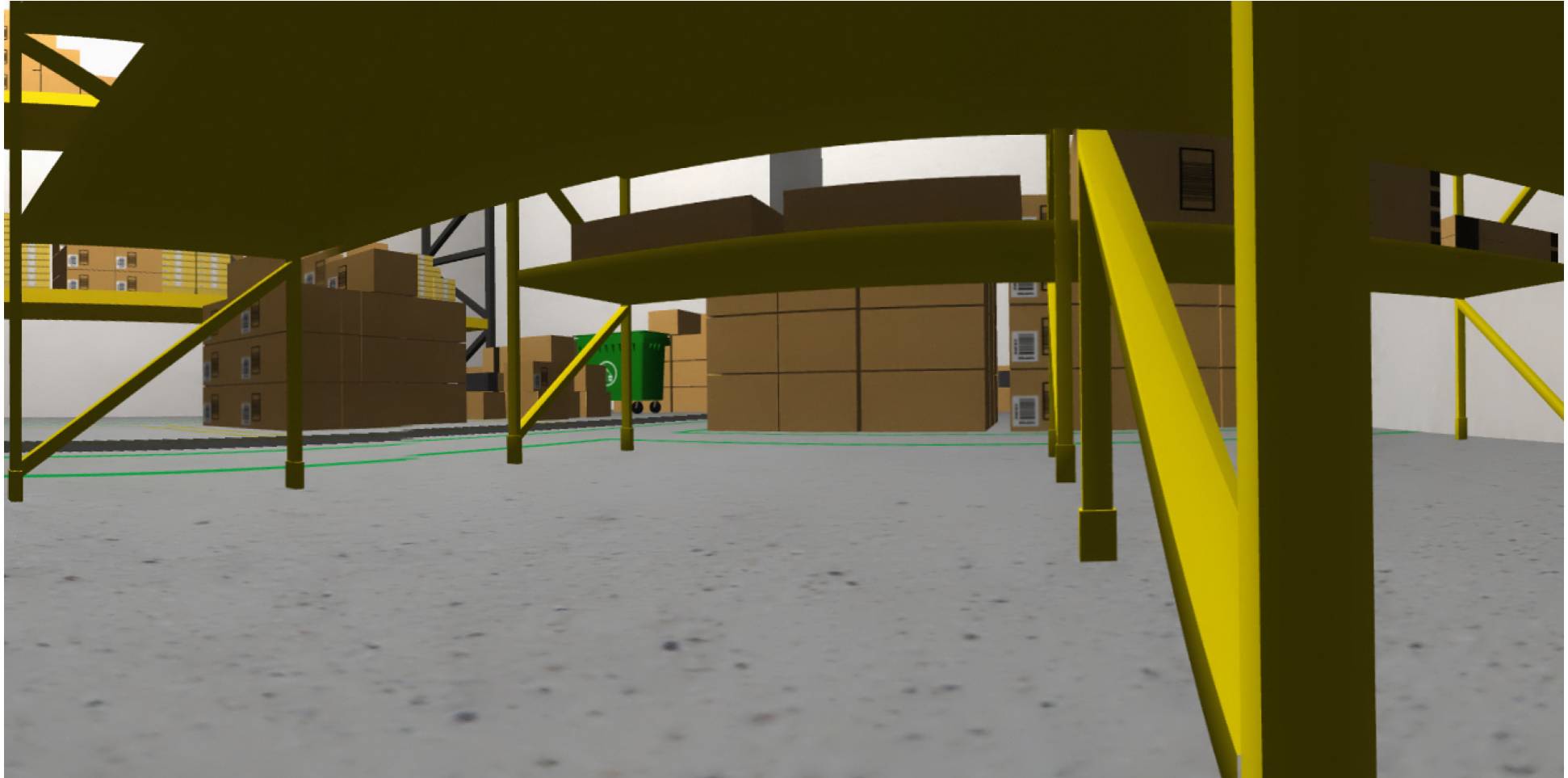}\\
(a) Warehouse goal image
\label{img:goal_image_warehouse}
\end{minipage}\hfil
\begin{minipage}{0.30\linewidth}
\centering
\includegraphics[width=\linewidth]{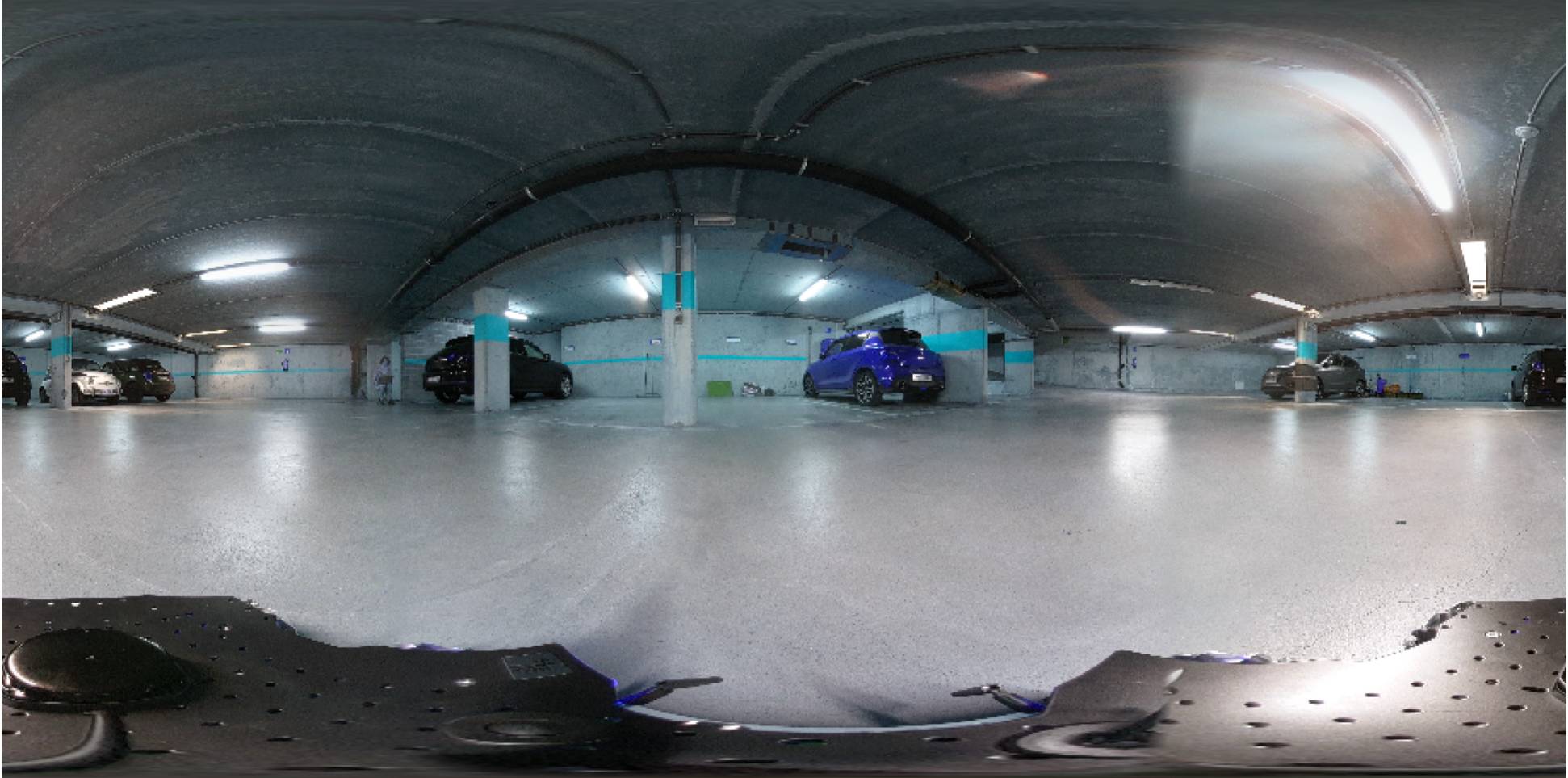}\\
(b) Parking lot goal image
\label{img:goal_image_garage}
\end{minipage}\\[1ex]
\begin{minipage}{0.40\linewidth}
\centering
\includegraphics[width=\linewidth]{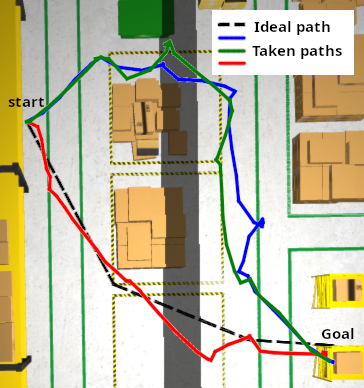}\\
(c) Warehouse paths to goal
\label{img:path_to_goal_warehouse}
\end{minipage}\hfil
\begin{minipage}{0.30\linewidth}
\centering
\includegraphics[width=1.2\linewidth]{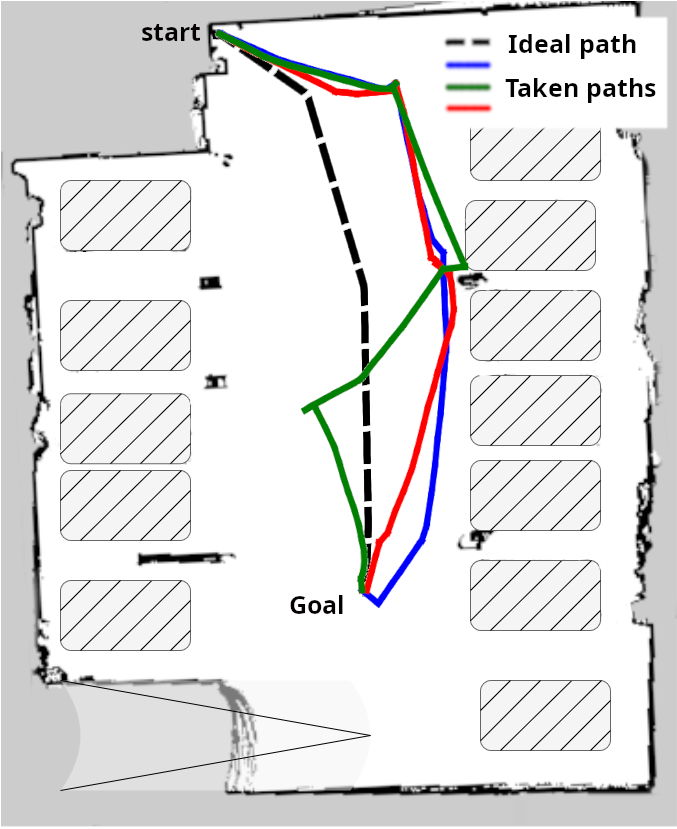}\\
(d) Parking paths to goal
\label{img:path_to_goal_garage}
\end{minipage}
\caption{Goal images and example paths taken by the agent (red, green, blue) from the start to the goal, alongside the ideal trajectory (dashed black line), in the simulated large warehouse (a, c) and the real parking lot (b, d).}
\label{img:paths_to_goals}
\end{figure*}

Performance is not uniform across goals. One goal in the simulated large warehouse consistently produced PLR~$\approx 1.50$ across five runs with low variance ($\pm 0.27$), accounting for three of the five worst-performing trials in the dataset. The A* distance for this goal is short (4.53~m) while the agent's executed distance is $7.4-8.1~m$, a near-doubling that we attribute to a conservative belief in $B_p$: the learned transition model assigns low feasibility to a direct path that A* on the true graph traverses freely. The remaining failures are distributed across distinct goals, suggesting that systematic detour behaviour is goal-specific rather than environment-specific. An example of such a case can be seen in Figure~\ref{img:paths_to_goals}.

Figure~\ref{img:paths_to_goals}(c) shows three executed paths from a fixed starting location to the same goal in the simulated warehouse, alongside the ideal trajectory (dashed black). Across trials, the agent reached the target observation while prioritising transitions with high expected utility; slightly longer paths are preferred when they offer higher confidence in expected outcomes, consistent with the Active Inference principle that action selection balances efficiency with belief consistency.

In real-world trials, the agent was evaluated in three environments of varying complexity: the controlled bedroom (2 runs), the structured warehouse (3 runs) and the semi-structured parking lot (12 runs). Figure~\ref{img:paths_to_goals}(d) illustrates trajectories taken in the parking lot to reach the corresponding goal image (b). Despite sensor drift, environmental change between exploration and goal-reaching phases (cars moving through the parking space) and varying lighting conditions (light on or off), the agent successfully reached its goals in 15 of 17 real-world runs. When initial localisation errors or unexpected obstacles occurred, belief updates and replanning enabled the agent to correct its course, as illustrated by minor detours in the green trajectory, where the agent got temporarily lost after drifting over an undetected obstacle (a car with a body higher than its lidar range).

Across the seventeen real-world goal-reaching experiments, only two failures occurred: one in the bedroom, caused by severe drift producing an incorrect re-localisation with the goal located beyond a wall, and one in the parking lot, where the robot's holonomic wheels became physically stuck in sandy terrain.

\subsection{Computational Scalability}
\label{sec:cpu}

To evaluate the scalability of AIMAPP, we measured its computational footprint in terms of memory usage and runtime performance on a Jetson Orin Nano platform (15V power, Jetpack~6.1, Ubuntu~22.04, ROS2~Humble).

\paragraph{Model Size and Memory Usage}

Model size was assessed by serialising parameters into a \texttt{.pkl} file across 31 independent runs. The number of stored states had no measurable effect on model size, confirming that the state-space representation remains lightweight as the map grows. The dominant memory factor is the storage of RGB panoramic observations. Model size (in MB) scales linearly with the number of stored observations:
\[
\text{Model size (MB)} \approx 1.32 \times (\text{number of observations}) + 0.17.
\]
The largest model contained 36 unique observations and required only 44.9~MB.

\paragraph{Runtime and Resource Usage}

\begin{figure}[ht!]
\centering    
\includegraphics[width=0.35\columnwidth]{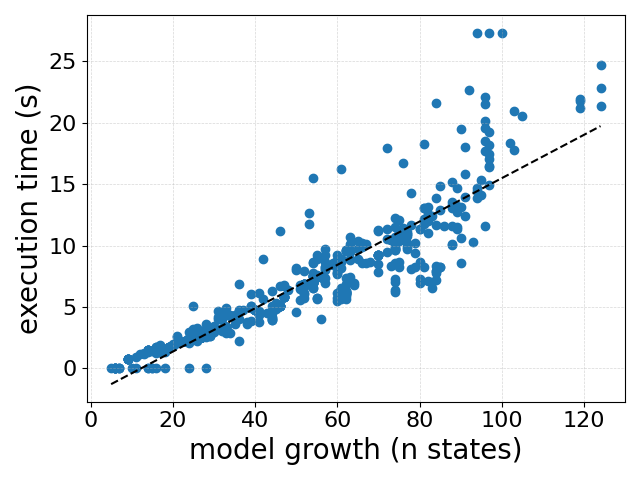}
\label{img:execution_time}
\includegraphics[width=0.35\columnwidth]{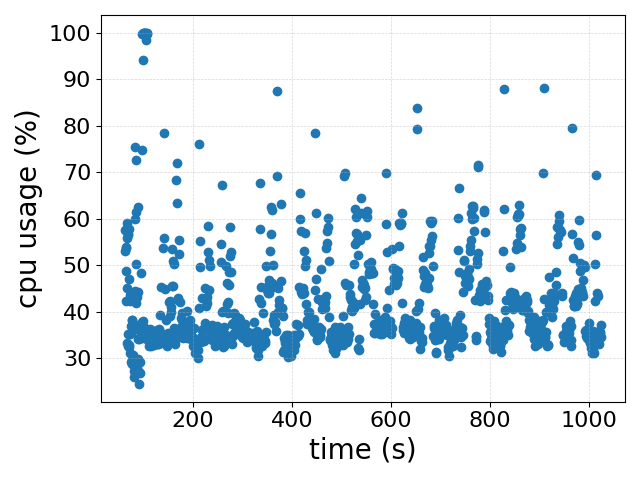}
\label{img:cpu_per}
\caption{Impact of AIMAPP with MCTS policy evaluation and Nav2 on the Jetson Orin Nano, measured over 31 experiments.}
\label{img:computational_costs}
\end{figure}

Runtime performance was profiled over 31 navigation runs on the Jetson Orin Nano. During each run, AIMAPP operated concurrently with the Nav2 stack, camera drivers, plotting, and logging processes. The average system resource consumption was $47.5\% \pm 16.8\%$ CPU and $36.8\%\pm2.2\%$ RAM. Nav2 accounted for approximately 30\% of CPU usage and remained stable across all trials. At no point did computation exceed platform capacity or cause performance degradation. As shown in Figure~\ref{img:cpu_per}, CPU load varied dynamically with task demands (periodic data and plots saving requiring more resources) but remained within approximately 60\% of available capacity. For comparison, the Frontiers baseline exhibited $42.7\pm26.2\%$ CPU and $26.3\pm1.2\%$ RAM usage over 14 runs.

\paragraph{Execution Time and Model Dimensionality}

Model execution time scaled linearly with model dimensionality (the number of states considered at each planning step):
\[
\text{Execution time (s)} \approx 0.18 \times \text{Model dimension} - 2.18.
\]
Figure~\ref{img:execution_time} illustrates this relationship. The use of MCTS for policy evaluation introduces computational overhead that scales with planning depth and simulation count. In our implementation, we used depth~10 and 30~simulations, chosen empirically as sufficient for the tested environments. Preliminary experiments in small environments suggest these could be reduced to depth~5 and 20~simulations with minimal performance degradation, improving real-time performance on resource-constrained platforms.

\section{Discussion}
\label{sec:discussion}

The presented results demonstrate that AIMAPP achieves competitive exploration and goal-directed navigation performance, comparable to state-of-the-art planning methods that do not require pre-training, in both simulated and real-world environments up to 325~m$^2$. We now discuss the limitations of the current system, situate it within a broader context relative to existing approaches, and identify directions for future work.

\subsection{Limitations}
\label{sec:limitations}

\paragraph{Planning and State Representation}

The current implementation relies on discretised actions ($x$ headings) and a fixed node influence radius, which may limit performance in highly constrained environments such as narrow passages where finer spatial resolution is required. Autonomous adaptive discretisation, reducing the influence radius or increasing angular resolution in cluttered regions, could address this limitation. 

The discrete-state representation does not account for the uneven distribution of information across space. A large open hall may be over-represented in the topological graph despite containing little novel information, while a narrow alley may provide greater utility but be collapsed into a few nodes. This inefficiency becomes more pronounced in larger environments.

The coverage plateau observed at approximately 90\% (Section~\ref{sec:exploration}) reflects a related limitation: the agent prioritises refining nearby unvisited states already in its graph rather than extending exploration beyond the current detection range. While this conservative strategy contributes to robustness, it reduces the rate at which new spatial coverage is acquired in the later stages of exploration.

\paragraph{Localisation and Drift}

Perceptual aliasing in symmetric environments is mitigated by the position variable $p_t$, which accumulates metric displacement over time. While identical visual observations would yield equal SSIM scores, the accompanying position estimates enable disambiguation. However, this mechanism depends on continuous tracking; if the robot were teleported (kidnapped robot problem), it would be unable to distinguish between symmetric locations and would likely converge to an incorrect belief with probability proportional to the number of indistinguishable locations. In our experiments, repeated visual patterns in corridor environments were successfully handled through this position-based disambiguation.

We investigated marginal message passing considering observations and actions up to four steps in a previous work over a simple 2D minigrid environment without realistic features~\cite{ours_model} as a form of belief smoothing for loop closure. However, this retrospective inference increased uncertainty over the current state rather than improving localisation, yielding performance below the single-step model we are currently using.  

Under extreme drift conditions, the belief-driven map gradually loses metric alignment with the ground truth. While the agent remained operational in these conditions, the two goal-reaching failures in real-world trials were both attributable to severe drift: one produced an incorrect re-localisation with the goal located beyond a wall, and the other resulted from the robot becoming physically stuck. Long-term metric consistency remains an open challenge for industrial deployment. Integrating a geometric back-end behind the topological layer, so that metric drift is corrected between re-localisation events rather than only at them, is a natural direction for closing this gap.


\paragraph{Perception}
The perception front-end is not part of the proposed Active Inference model; it is an interchangeable module whose current implementation uses panoramic images compared via SSIM as a proof-of-concept. This choice introduces several weaknesses that are specific to the perception pipeline rather than to the decision-making framework.

First, the SSIM-based pipeline is brittle to environmental change, i.e., small modifications in the scene (e.g., a displaced object or altered lighting) can prevent the agent from recognising the goal upon reaching the correct location. This brittleness was directly responsible for reduced goal recognition reliability in real-world trials, where environmental conditions varied between exploration and goal-reaching phases. In the limiting case, a goal location defined by a panoramic image may become entirely invalid if the scene has changed sufficiently that no stored observation matches the current input, even at the correct position. Replacing SSIM with semantic object detection or learned visual representations would address this without modifying the generative model.

Second, the observation likelihood model $A_o$ associates multiple observations with each state, providing a mechanism for capturing temporal variability. However, as the map expands, observation likelihood distributions for earlier states naturally dilute, potentially requiring the agent to revisit previously mapped locations to refresh perceptual beliefs. This dilution reflects an inherent trade-off in dynamically updated cognitive maps compared to static representations, and is independent of the particular observation modality.

Third, the current evaluation does not include continuously dynamic environments such as spaces with pedestrian traffic. While the framework accommodates moderate environmental changes (as demonstrated with relocated obstacles in simulation), performance in settings with frequent structural change remains untested. Handling such scenarios requires robust dynamic object filtering at the perception level and reactive obstacle avoidance at the motion-planning level, neither of which falls within the scope of this work.
Finally, the current implementation assumes approximate heading consistency between visits to the same location. The 360$^\circ$ panoramic coverage provides sufficient overlap to tolerate moderate heading variation, but a better perception module invariant to this kind of consideration would be beneficial.

\subsection{Broader Context}
\label{sec:broader_context}

\paragraph{Comparison with Baselines}

The baselines evaluated in this work reflect distinct design philosophies. AIMAPP and FAEL both emphasise exploration efficiency, achieving the highest coverage-to-distance ratios. However, FAEL requires a 3D LiDAR and achieves a success rate below 50\%, frequently stalling due to mismatches between topological node creation and obstacle detection in its 3D mapping. GBPlanner is tailored for robust subterranean exploration where narrow passages dominate; it achieves the highest success rate (87\%) but at the cost of conservative, distance-inefficient path planning with redundant back-and-forth movements in large environments. Frontiers, as a lightweight heuristic, performed worst in cluttered environments due to its greedy, reactive strategy.

A distinguishing property of AIMAPP is that it unifies exploration and goal-directed navigation within a single generative model, without requiring retraining, reinitialisation, or pipeline restructuring. Unlike FAEL, GBPlanner, and Frontiers, which are explicitly optimised for exploration alone, AIMAPP switches between discovering new areas and reaching specific observations by adjusting hyperparameters controlling the balance between pragmatic value and information gain. The framework naturally supports both objectives simultaneously.

\paragraph{Comparison with Online Learning Approaches}

Comparison with other online learning approaches represents a valuable direction.
Active NTFields~\cite{Liu_2025} is the most relevant comparator for the online learning property of AIMAPP, as it also maps an environment from depth observations without pre-training and enables fast motion planning once the map is constructed. However, the two systems differ in problem scope, sensor requirements and computational demands, which together preclude a direct comparison under our current experimental setup. Active NTFields separates mapping from planning: it first explores the environment using a standard frontier-coverage strategy to build a complete neural arrival-time field. It does not support goal-directed navigation during partial exploration. AIMAPP, by contrast, interleaves exploration and goal-reaching within a single decision-making loop via EFE minimisation, operating on incomplete maps throughout. Comparing exploration coverage alone would not capture this difference. On the sensor side, Active NTFields requires dense 3D depth images with accurate pose registration and GPU-accelerated neural network training (reported on a 3090 RTX with 128 GB RAM and 2.12 seconds per frame). AIMAPP operates with a 2D LiDAR and a forward-facing RGB camera on a Jetson Orin Nano. Evaluating their relative exploration efficiency would provide further insight into the trade-offs between physics-informed neural representations and topological generative models, which we leave to future work.

\paragraph{Relation to Cognitive Mapping}

The stepwise propagation of utility observed in the goal-reaching experiments (Figure~\ref{img:heatmap_goals}) resembles place cell firing in the hippocampus, where spatial locations associated with expected sensory outcomes are activated before the agent physically reaches them~\cite{hippo_nav}. Forward sweeps of place cell activation have been observed as organisms consider alternative routes, with the most robust sweeps predicting the direction of subsequent locomotion~\cite{Johnson2007,Kay2020}. This is consistent with the view of place cells as encoding a "predictive map" based on successor representations of likely state transitions~\cite{stachenfeld2017}, from which anticipated future
locations are derived before the agent arrives at them. We do not claim that AIMAPP implements these human mechanisms, only that the inductive prior produces a comparable forward-projection pattern from a probabilistic inference.

By integrating predicted observations with trajectory planning, AIMAPP dynamically selects actions maximising the likelihood of reaching desired sensory states while accounting for uncertainty. This parallel suggests that our AIF application and biological navigation may share structural features through the inductive prior mechanism and hippocampal predictive coding. The current perception pipeline, using panoramic images and SSIM-based comparison, could be replaced by learned visual representations (e.g., from a pre-trained encoder or a visual place recognition network) without modifying the underlying Active Inference framework. Such an extension would more closely parallel the hierarchical processing from visual cortex to hippocampus observed in biological systems, where abstract, viewpoint-invariant representations are formed before reaching place-coding regions~\cite{vision_based_place_reco}. The modularity of AIMAPP explicitly supports this separation: the generative model operates on abstract state and observation variables, remaining agnostic to how those observations are derived from raw sensory input.

\paragraph{Sensor Agnosticism}

While this work uses LiDAR to detect obstacles through the observation vector $o$, the model treats depth as a generic perceptual feature. Future implementations could replace LiDAR with depth-from-vision or any other sensing modality without altering the underlying decision-making process. This sensor-agnostic design, combined with the ROS-based integration, ensures that AIMAPP can be deployed across diverse robotic platforms with minimal adaptation.

\subsection{Future Work}
\label{sec:future_work}

Several directions could address the limitations identified above and extend the capabilities of the framework.

\paragraph{Perception Improvements}

The SSIM-based observation comparison should be replaced by a more robust recognition module. Semantic object detection~\cite{object_nav_semantic,object_nav_zero_shot} would improve goal recognition, particularly for task-critical observations where small environmental changes currently prevent the agent from confirming arrival at the correct location. More broadly, the panoramic image pipeline could be substituted with learned visual representations from a pre-trained encoder or a visual place recognition network~\cite{vision_based_place_reco}, which would provide viewpoint-invariant features better suited to real-world variability. Integrating dynamic object filtering would further improve performance in environments with moving entities. These replacements require no modification to the generative model itself, as the Active Inference framework operates on abstract observation variables.

\paragraph{Planning Improvements}

A long-standing limitation of the current framework is the finite planning horizon imposed by MCTS, which restricts the depth over which the agent can evaluate future trajectories. Probability dilution over long planning horizons remains an open challenge in AIF. Scalable methods for balancing information gain and goal achievement in large state spaces with long planning horizons remain an open challenge in AIF~\cite{AIF_bandit_explo_LATEST,sophisticated_AIF}. Hierarchical reasoning~\cite{ours_hierarchy} offers the most principled mitigation: by separating long-term topological planning from short-term metric control, a hierarchical generative model would enable the agent to reason over coarser spatial abstractions at higher levels while preserving fine-grained resolution at lower levels~\cite{robot_nav_hierarchy_ozan,ours_hierarchy}. This would simultaneously extend the effective planning horizon and accelerate coverage in large open areas. Complementarily, grouping states by information gain~\cite{chunk_space,grid-cell_fragmentation_map_building,RHNVBP} would reduce redundancy in the topological graph by collapsing over-represented regions (e.g., open halls) into fewer abstract
nodes, improving both map compactness and path planning efficiency. Within such a hierarchy, message-passing schemes~\cite{wouter_message_passing} could propagate beliefs across temporal scales, enabling the agent to reason about distant goals that currently fall outside its prediction horizon without sacrificing the computational tractability of the single-level model.

At the level of the current flat model, more sophisticated MCTS rollout policies incorporating learned heuristics (e.g., A* cost estimates) could improve planning efficiency as the map grows; however, this direction remains untested in the present
implementation.

\paragraph{Environmental Robustness}

Adaptive discretisation, autonomously reducing the influence radius or increasing angular resolution in cluttered regions, would improve performance in narrow passages. Continuous action spaces~\cite{AIF_book} offer a theoretically principled extension but require alternative inference schemes, such as gradient-based optimisation, and are left for future investigation.

Alternative temporal smoothing architectures should be explored to revisit belief smoothing for loop closure. 

Finally, while the current system tolerates temporary sensor failures by pausing until data becomes available, developing graceful degradation strategies for prolonged sensor loss would be essential for deployment in safety-critical settings.

\section{Conclusion}
We have presented AIMAPP, an Active Inference-based navigation framework unifying mapping, localisation, exploration, and goal-directed navigation within a single generative model. The agent builds and updates a sparse
topological map online, learns state transitions through sensorimotor feedback, and plans actions by minimising Expected Free Energy, requiring no pre-training, metric maps, or task-specific tuning.


Across seven environments (four simulated, three real-world, up to $325m^2$), AIMAPP achieved 90\% of human-teleoperated coverage efficiency in simulated environments, and 79\% exploration success rate overall. In goal-reaching experiments, AIMAPP achieved a median path length 24\% above the A* optimum across 40 trials, with 15 of 17 real-world trials successful.
These results provide evidence that AIF can serve as a practical navigation strategy for mobile robots, bridging neuroscience-inspired models of spatial cognition and the demands of real-world autonomy. The modular, sensor-agnostic architecture of AIMAPP supports integration with existing ROS-based systems and
permits straightforward substitution of perception, planning, or control modules as they improve. Limitations in goal recognition, planning horizon, and long-term metric consistency are future research areas with plausible solutions, including hierarchical reasoning, semantic perception, and adaptive state discretisation.

\backmatter

\bmhead{Data}
Our model and videos are available at \url{https://github.com/decide-ugent/aimapp}

\bmhead{Acknowledgements}
Large language models were used to assist with grammar correction and linguistic articulation during the preparation of this manuscript.
This research received funding from the Flemish Government (AI Research Program) under the “Onder-zoeksprogramma Artificiële Intelligentie (AI) Vlaanderen” programme and the Inter-university Microelectronics Centre (IMEC).





\begin{appendices} 
\section{Model Definition}

\subsection{Model Parameters}
\label{app:model_param}

This section outlines the key parameters used to configure our active inference-based navigation model. These parameters define the agent’s internal representation, planning process, and learning dynamics. Hyperparameters set at initialisation are presented Table~\ref{tab:hyperparameters}. All values were chosen empirically without comparison testing. 


\begin{table*}[b]
\centering
\caption{Hyperparameters for Active Inference-Based Navigation Model}
\label{tab:hyperparameters}
\resizebox{\textwidth}{!}{
\begin{tabular}{lll}
\hline
\textbf{Parameter} & \textbf{Description} & \textbf{Value Used} \\ \hline
Number of likelihood matrices & \begin{tabular}[c]{@{}l@{}}Number of observation types (e.g., visual, pose). \\ Theoretically extendable to LiDAR, radar, etc.\end{tabular} & 2 (visual and pose) \\ \hline
Number of available actions & \begin{tabular}[c]{@{}l@{}}Discretized actions over 360$^\circ$ rotation/movement space. \\ Odd numbers include a "stay" action.\end{tabular} & 13 actions \\ \hline
Physical radius & Collision buffer for state creation near obstacles. & Robot-specific \\ \hline
Influence radius & Minimum distance between states to promote map compactness. & 1\,m (env. $\leq 40m^2$), 2\,m (all others) \\ \hline
Lookahead nodes & \begin{tabular}[c]{@{}l@{}}Maximum consecutive nodes hypothesized \\ within LiDAR range for anticipatory map expansion.\end{tabular} & 8 nodes \\ \hline
SSIM threshold & \begin{tabular}[c]{@{}l@{}}Below this threshold, two images are considered the same occurrence \\ and the latest one overwrites the previous image.\end{tabular} & 65\% \\ \hline
$A_o$ Learning rate & \begin{tabular}[c]{@{}l@{}}How much new observations impact \\ the observation Likelihood matrix update.\end{tabular} & 1 \\ \hline
MCTS depth & \begin{tabular}[c]{@{}l@{}}How far can the simulation horizon predict \\ (also called the lookahead horizon) - in step-\end{tabular} & 10 \\ \hline
MCTS simulation runs & \begin{tabular}[c]{@{}l@{}}How many time do we run the MCTS simulation \\ before determining which action is the most ideal\end{tabular} & 30 \\ \hline
\end{tabular}
}
\end{table*}


\subsubsection{Transition Update}

\begin{equation}
    B_\pi = B_\pi + Q(s_t|s_{t-1}, \pi) Q(s_{t-1}) * B_\pi  * \lambda
    \label{eq:B_up}
\end{equation}

To update its beliefs about the environment, the agent uses a Dirichlet-based pseudo-count mechanism (Equation~\ref{eq:B_up}) with situation-dependent learning rates ($\lambda$). These rates vary based on whether the agent:
\begin{itemize}
    \item successfully reaches a location,
    \item is physically blocked while trying to move,
    \item or anticipates reaching (or failing to reach) a location based on sensory evidence.
\end{itemize}

Table~\ref{tab:tran_lr} lists the specific learning rates for each situation. This continual adjustment allows the agent to refine transition likelihoods in its internal model rapidly.




 \begin{table}[h!t]
    \centering
    \caption{Transition learning rate ($\lambda$) depending on the situation}
    \begin{tabular}{l|c|c|c|c}
        \textbf{Transitions} & \textbf{Possible} & \textbf{Impossible} & \shortstack{\textbf{Predicted} \\ \textbf{Possible}} & \shortstack{\textbf{Predicted} \\ \textbf{Impossible}} \\
        \hline
        Forward  & 7  & -7 & 5 & -5 \\
        \hline
        Reverse  & 5  & -5 & 3 & -3 \\
    \end{tabular}
    \label{tab:tran_lr}
\end{table}

\subsubsection{Uncertainty About Current State}

To determine whether an agent is lost, we evaluate its certainty about the current state. Specifically, we compute a Z-score to measure how strongly the most likely state stands out compared to the others. If this dominance falls below a user-defined threshold (set to 4 in this work), the agent is considered uncertain about its location.

\subsection{Expected Free Energy Terms}
\label{app:efe_terms}

H is an inductive term applied to the state preference $Cs$ defined in Equation~\ref{eq:H(n)}, which is computed inductively by propagating backwards in time from the goal state toward the current state (in $n$ steps, that might be lower than the planning horizon). This inductive process accumulates structural preferences and becomes silent (i.e., has no influence) if no specific target state is preferred or if the preferred state lies beyond the predictive horizon.
Let $Cs_0$ be a weighted preference vector over existing states, as derived from the joint preference over observations ($Co$) and transitions ($Cp$). Then the backwards recursion is defined as~\cite{inductive_AIF}:
\begin{equation}
\begin{aligned}
    Cs_n =& B_s^T\odot Cs_{n-1} \ \ \ \ for\ n = {0,..,\tau}\\
    H =& ln(\epsilon) \cdot (B_s^T\odot Cs_\tau)\odot s(\tau)
\end{aligned}
\label{eq:H(n)}
\end{equation}

$n$ is a backwards value propagating from the goal (thus the future, if under our prediction horizon) up to our current state $t$, if $Cs$ lies over the horizon, we can predict, this term will also be silent.


\subsection{Monte Carlo Tree Search}
\label{app:MCTS}

We integrate Monte Carlo Tree Search (MCTS) with Active Inference to evaluate the expected free energy of potential paths, overcoming the limitations of fixed policy sets in standard Active Inference models~\cite{weird_HAIF, ours_model, plan_nav_AIF_friston}. Unlike traditional approaches, where policies are predefined and restrict adaptability, MCTS dynamically simulates future action sequences (rollouts) from the current belief state, guided by the principle of minimising expected free energy~\cite{MCTS_AIF}. This enables deeper lookahead without combinatorial explosion, balancing goal-directed behaviour and exploration while reducing computational load.
MCTS also allows the agent to revise trajectories based on new evidence, resulting in robust, context-sensitive navigation in uncertain environments. Our implementation (Algorithm~\ref{algo:MCTS}) uses two key parameters: policy length (number of future steps to predict, set to 1) and lookahead policy (depth of simulation horizon, set to 10 in Equation~\ref{eq:H(n)}). For each step, MCTS runs 30 simulations (number chosen empirically) to determine the optimal action.

\begin{algorithm*}[hb!]
\label{algo:MCTS}
\TitleOfAlgo{AIF based MCTS}
\SetKwInput{KwInput}{Input}                
\SetKwInput{KwOutput}{Output}              
  
  \KwInput{$possible\_motions$,$policy\_length$, $lookahead\_policy$, $num\_simulation$}
  \KwOutput{best policy}
  \SetKwFunction{FExpand}{Expansion}
  \SetKwFunction{FPlan}{MCTS planning}
  \SetKwFunction{FSelect}{Selection}
  \SetKwFunction{FSimulate}{Simulation}
  \SetKwFunction{FBackpropagate}{Backpropagate}
  \SetKwFunction{FSelectBestAction}{SelectBestAction}
  \SetKwFunction{FInfer}{Infer}
\SetKwProg{Fn}{Function}{:}{}
\Fn{\FPlan{$possible\_motions$,$policy\_length$, $lookahead\_policy$,$num\_simulation$}}{
    $qs \gets$ get believed localisation over states() \;
    $qo \gets$ get expected observations for qs($qs$) \;
    $root \gets$ create root state($qs$, $qo$, $possible\_motions$)\;
    
    \For{$i \leftarrow 1$ \KwTo $policy\_length$}
    {
        
        \For{$i \leftarrow 1$ \KwTo $num\_simulation$}
        {
        $leaf \gets$ \FSelect{root}\;
        $expanded\_leaf \gets$ \FExpand{$leaf$}\;
        $reward \gets$ \FSimulate{$expanded\_leaf$, $lookahead\_policy$}\;
        \FBackpropagate{$expanded\_leaf$, $reward$}\;
        }
        $action_i \gets$ \FSelectBestAction{root}\;
        $root \gets leaf$\;
        selected actions $\gets action_i $
    }
    \KwRet selected actions\;
  }
\SetKwProg{Fn}{Function}{:}{}
\Fn{\FSelect{$node$}}{
    \While{$node$ has children}{
        $node \gets$ child of $node$ with highest UCB1 score\;
    }
    \KwRet $node$ \;
  }

\Fn{\FExpand{$node$}}{
    \ForEach{$action$ in $node.possible\_actions$}{
        $next\_pose \gets$ Transition($node.pose$, $action$)\;
    
        $next\_pose, qs', qo', G \gets$ \FInfer{$node.qs$, $action$}\;
        Create new child node with state $(qs', next\_pose, qo', G)$ \;
    }
    \KwRet $node$ \;
}
\Fn{\FSimulate{$node$, $lookahead\_policy$}}{
     \ForEach{$action$ in $node.possible\_actions$}{
        $action \gets$ choice from node $actions$\;
        $next\_pose, qs', qo', G \gets$ \FInfer{$qs$, $action$, $lookahead\_policy$}\;
        \uIf{best measured $G$}{
             $best\_G \gets G$\;
        }
    }
        $total\_reward \gets total\_reward + best\_G$\;
    \KwRet $total\_reward $ \;
}

\Fn{\FBackpropagate{$node$, $reward$}}{
    \While{node exists}{
        $node \gets$ update node($node$, $reward$)\;
        $node \gets node.parent$\;
    }
}

\Fn{\FSelectBestAction{root}}{
    \ForEach{$(a, child)$ in $root.childs$}{
        Compute average reward of $child$\;
    }
    \KwRet action $action$ with highest AIF policy score \;
}
\SetKwProg{Fn}{Function}{:}{}
\Fn{\FInfer{$qs$, $action$}}{
    $next\_pose \gets$ Transition($pose$, $action$)\;
        $qs' \gets$ BeliefTransition(qs, $action$)\;
        $qo' \gets$ ExpectedObservation($qs'$)\;
        $G \gets$ ExpectedFreeEnergy($qs'$, $qo'$, $qs$, $action$)\;
    \KwRet $next\_pose$, $qs'$, $qo'$, $G$ \;
  }
\end{algorithm*}

\section{Robotic Platforms and Sensors}
\label{app:robots}
Our system is robot-agnostic; however, we have to adapt the sensor pipeline to the specific sensors used. 
All simulated experiments use a Turtlebot3 Waffle~\cite{turtlebots}, except FAEL, which used a Jackal~\cite{jackal}. Our agent used a Pi camera and a 360-degree 2D LiDAR with 12~m range, Gbplanner uses two 3D lidars and three Pi cameras and FAEL one 3D Lidar, all with a maximum range of 12m for fair comparisons. 
Real-world trials used a RosbotXL~\cite{rosbotXL} with an 18~m range 360-degree LiDAR (restricted to 12~m for comparable results to simulated environments) and a 360-degree camera (Ricoh Theta X). In our implementation, the obstacle distance $c$ was extracted from the LiDAR sensor rather than the visual observation $o$ due to the absence of a depth or stereo camera; this substitution has no impact on the model, as the generative model treats depth as a generic perceptual feature.




\section{Environments}
\label{app:envs}

\begin{figure*}[htb!]
\centering
\begin{minipage}{0.28\linewidth}
\centering
\includegraphics[width=\linewidth]{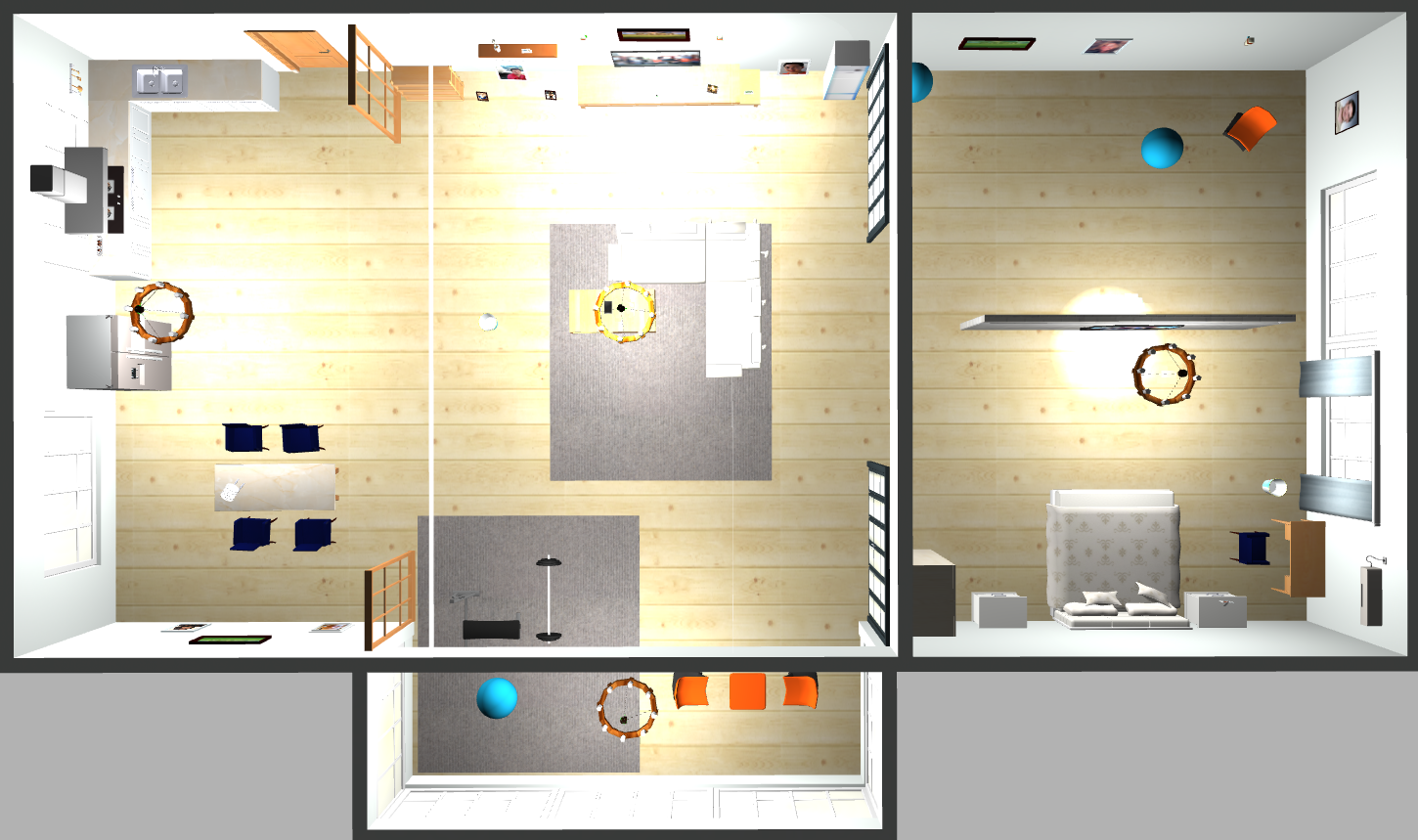}\\
(a) Home 175m$^2$
\end{minipage}%
\hfill
\begin{minipage}{0.28\linewidth}
\centering
\includegraphics[width=\linewidth]{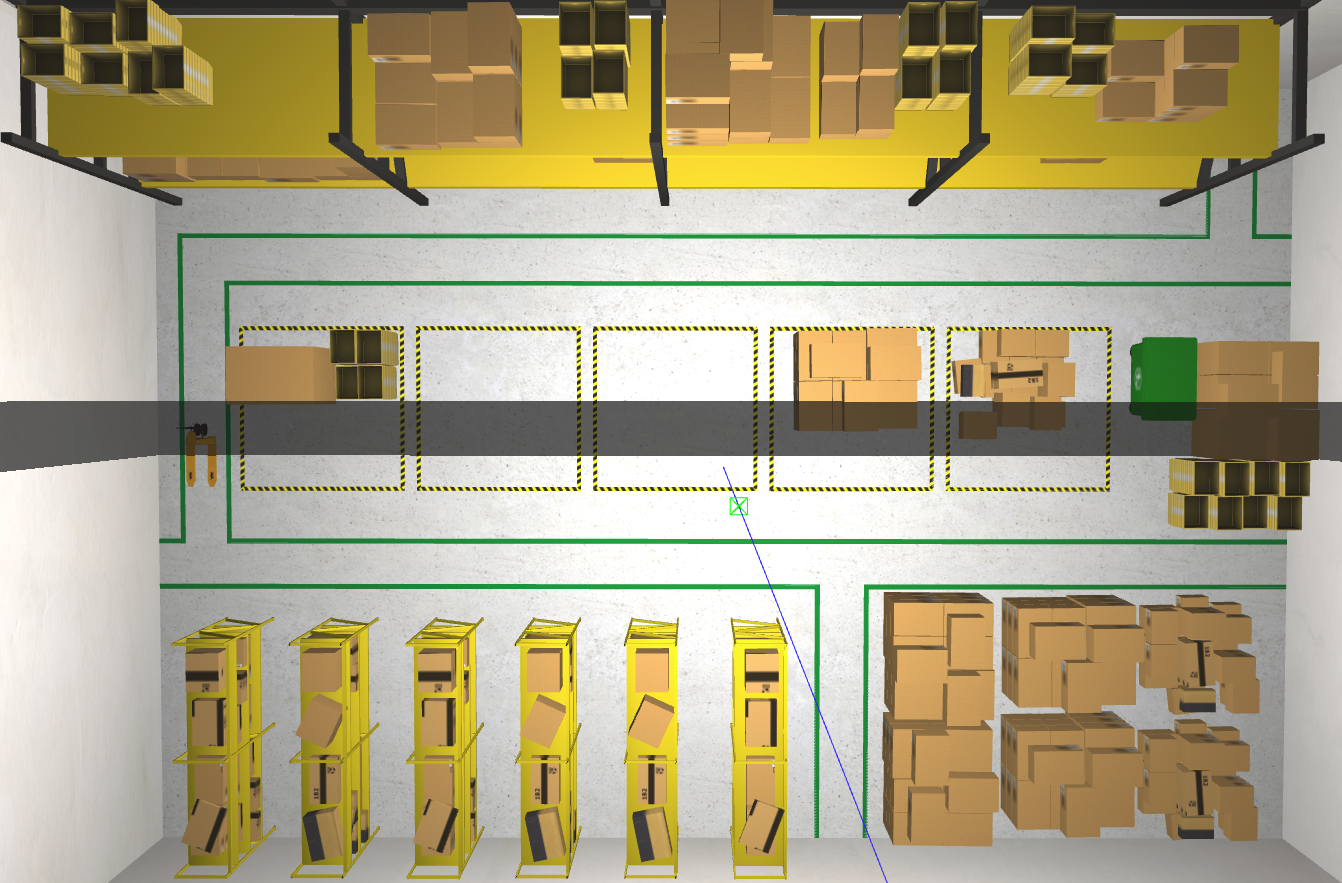}\\
(b) Big warehouse 280m$^2$
\end{minipage}%
\hfill
\begin{minipage}{0.28\linewidth}
\centering
\includegraphics[width=\linewidth]{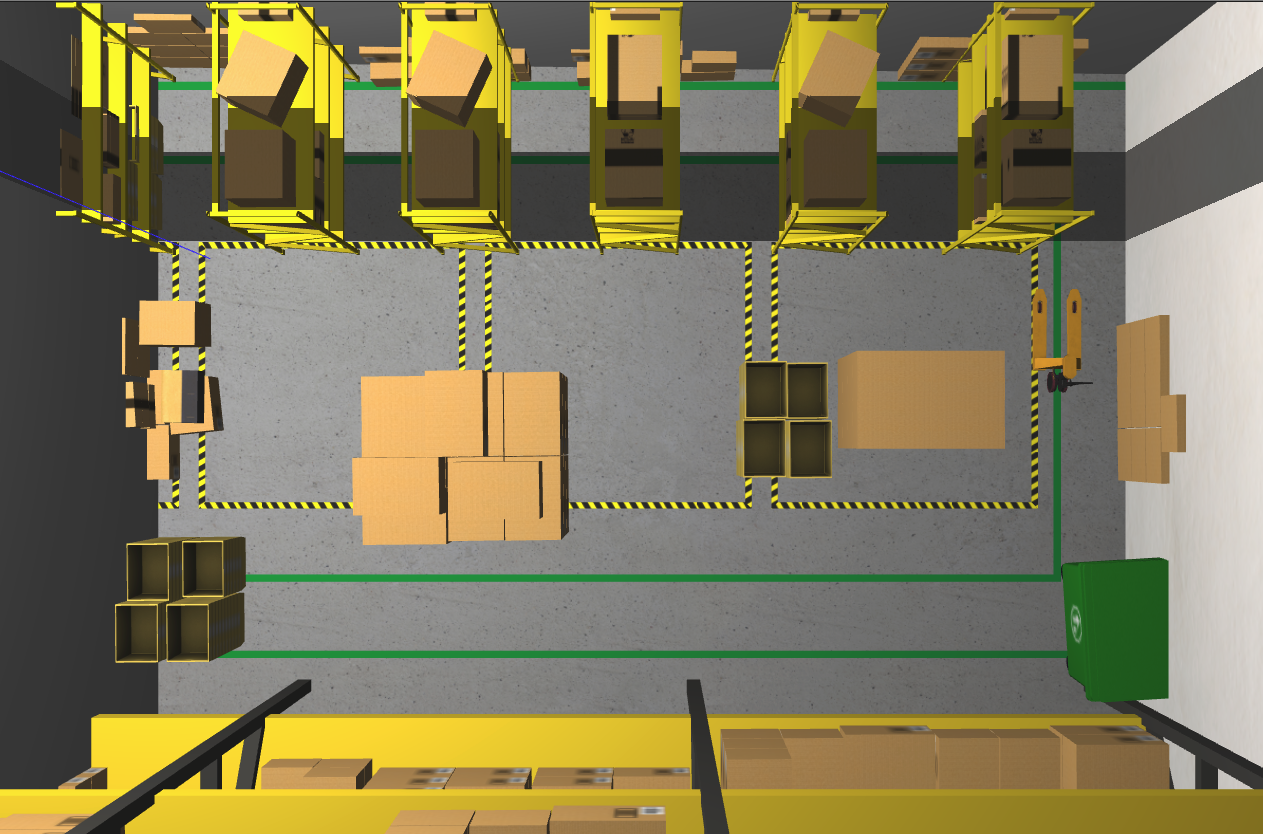}\\
(c) Small warehouse 80m$^2$
\end{minipage}%
\\[2ex]
\begin{minipage}{0.28\linewidth}
\centering
\includegraphics[width=\linewidth]{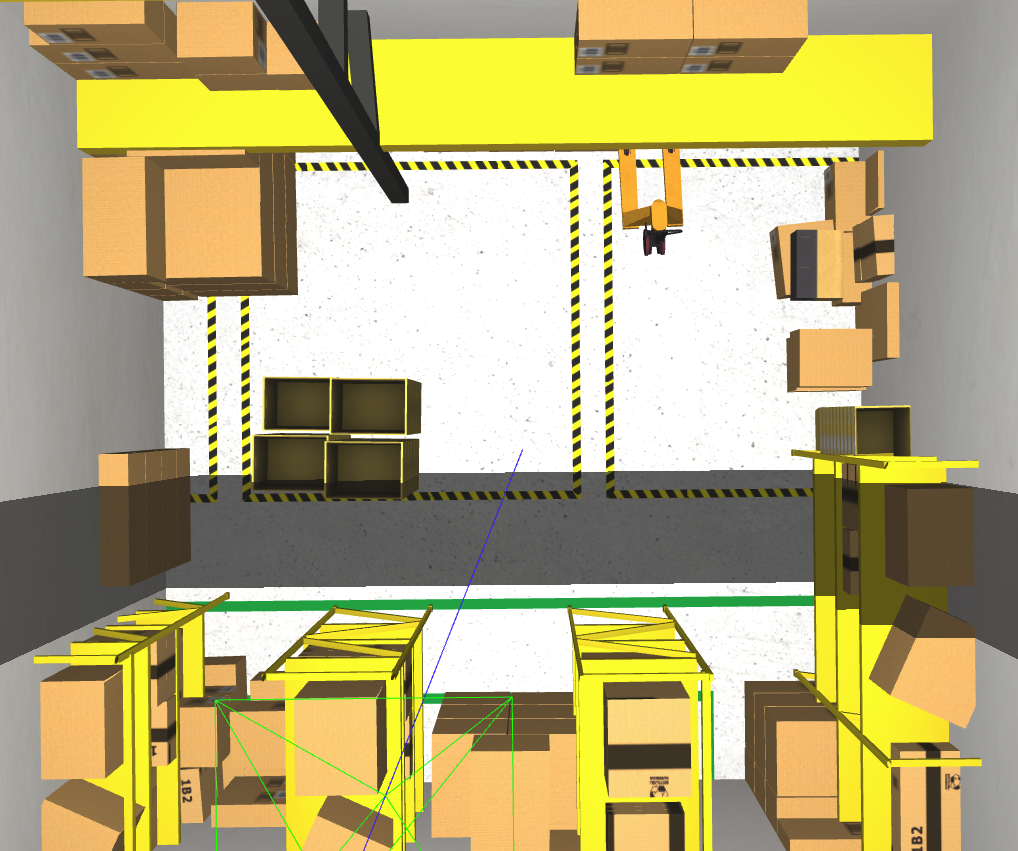}\\
(d) Mini warehouse 36m$^2$
\end{minipage}%
\hfill
\begin{minipage}{0.28\linewidth}
\centering
\includegraphics[width=\linewidth]{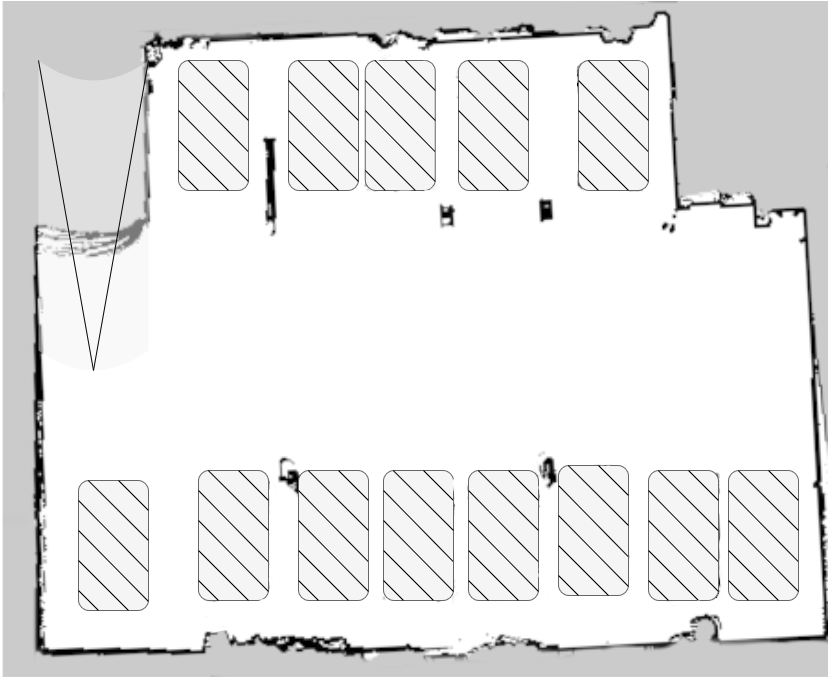}\\
(e) Parking lot 325m$^2$
\end{minipage}%
\hfill
\begin{minipage}{0.28\linewidth}
\centering
\includegraphics[width=\linewidth]{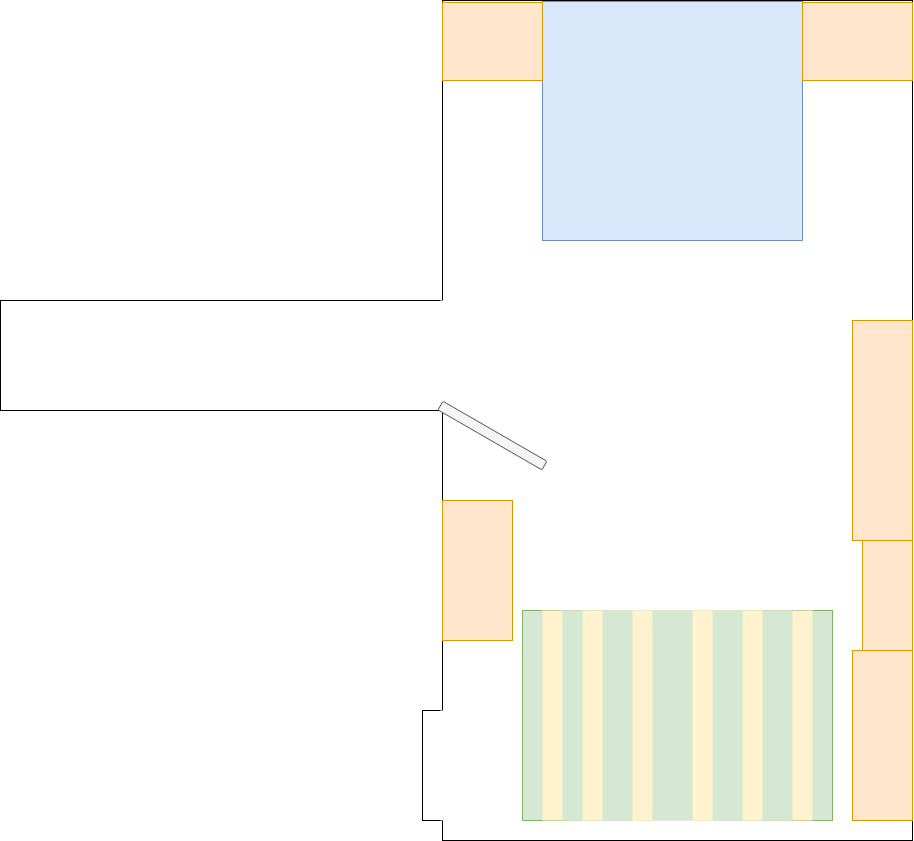}\\
(f) Real-world bedroom 20.27m$^2$
\end{minipage}%
\\[2ex]
\begin{minipage}{0.35\linewidth}
\centering
\includegraphics[width=\linewidth]{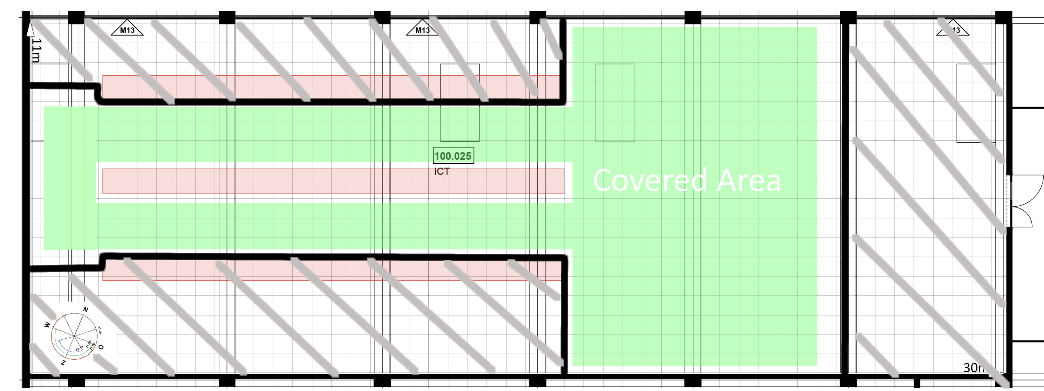}\\
(g) Real-world warehouse 186m$^2$
\end{minipage}%
\caption{Environments used in simulations and real-world experiments. Dashed grey areas are inaccessible; green areas correspond to the Qualysis odometry coverage.}
\label{fig:envs}
\end{figure*}

\begin{figure*}[htb!]
    \centering
   \includegraphics[width=0.50\linewidth]{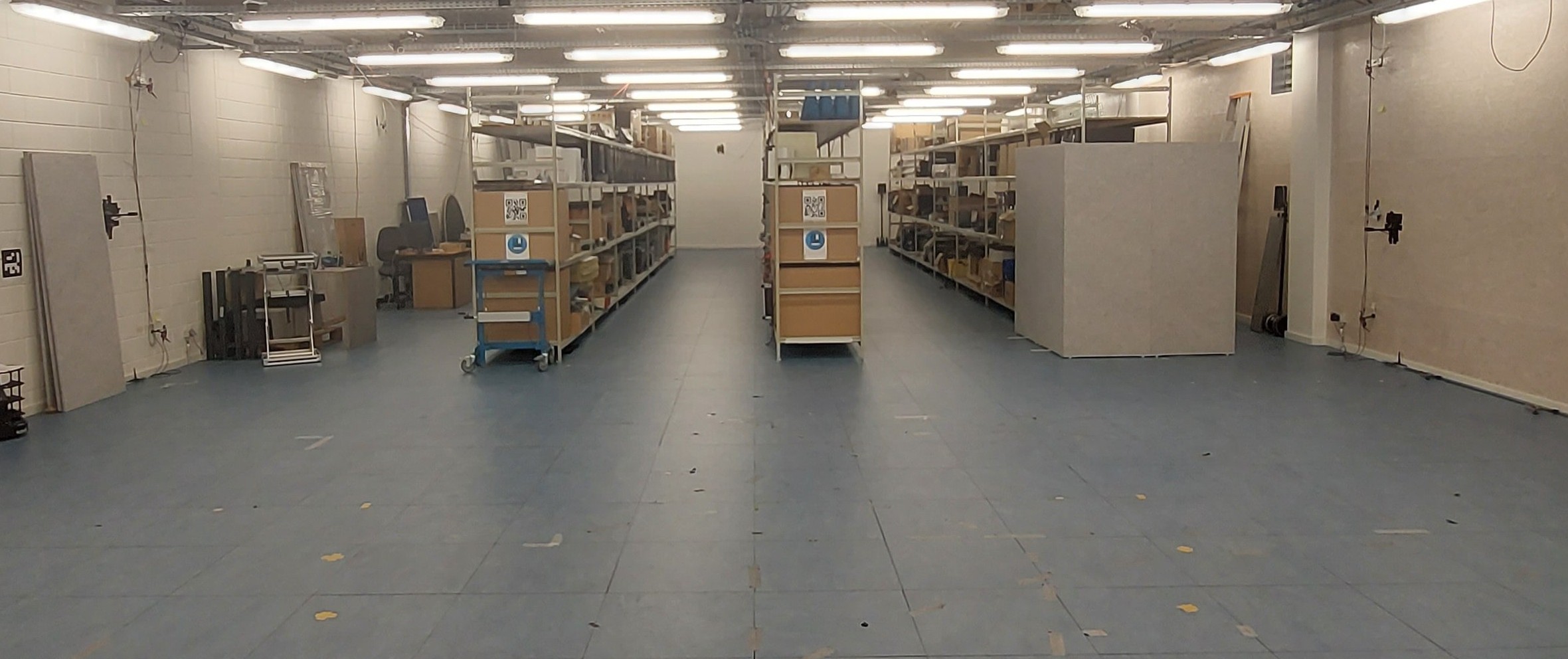}%
\hspace{0.01\linewidth}
\includegraphics[width=0.35\linewidth]{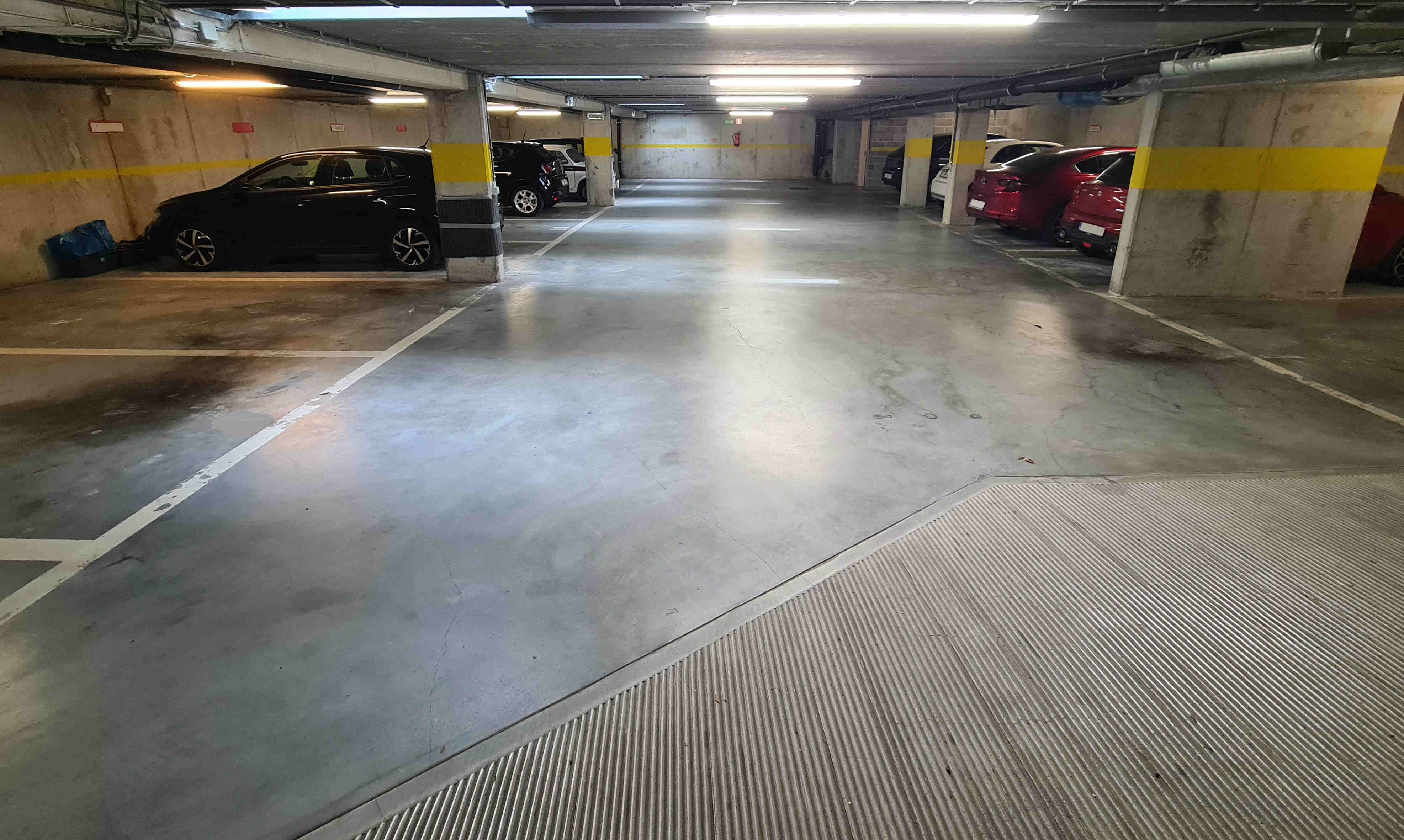}%
\hspace{0.01\linewidth}
    \caption{Picture of the structured warehouse (left) and semi-structured parking lot (right)}
    \label{fig:env_pictures}
\end{figure*}

Our experiments used a home~\cite{house} of 156m$^2$ and a warehouse~\cite{warehouse} of 3 different sizes, ranging from 36m$^2$, 80m$^2$ up to 280$^2$ and a real-world house of ~21m$^2$, a warehouse of 185m$^2$ and a parking lot of 325m$^2$. All environments are presented in Figure~\ref {fig:envs}. Pictures are presented in Figure~\ref{fig:env_pictures}.

\section{Details about experimental results}

\subsection{Adversarial Models}
\label{app:other_models}
GBPlanner, FAEL, and Frontiers were used with their given parameters, except for the map resolution, which was increased to 0.05m/cell. Other parameters, such as obstacle inflation, were modified to ensure that all agents could physically reach every location.

Despite our efforts, FAEL wouldn't fully turn around the central box in the mini warehouse, explaining why Figure~\ref{fig:experimental_results}(b) coverage is not 100\%.

\subsection{Exploration paths}
\label{app:explo_paths}

\begin{figure*}[htb!]
\centering
\begin{minipage}{0.4\linewidth}
\centering
\includegraphics[width=\linewidth]{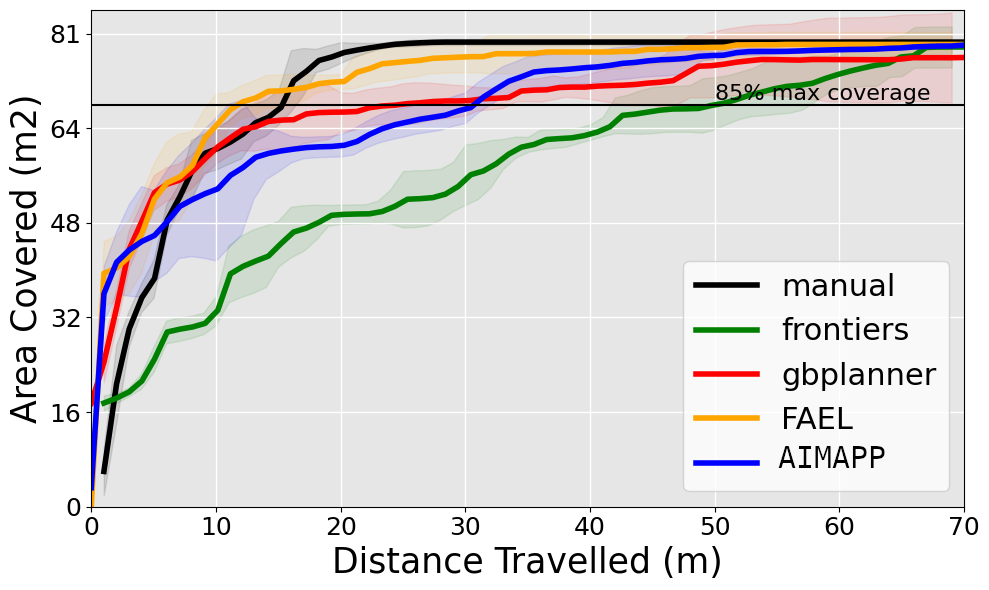}\\
(a) Small warehouse
\end{minipage}%
\hfill
\begin{minipage}{0.4\linewidth}
\centering
\includegraphics[width=\linewidth]{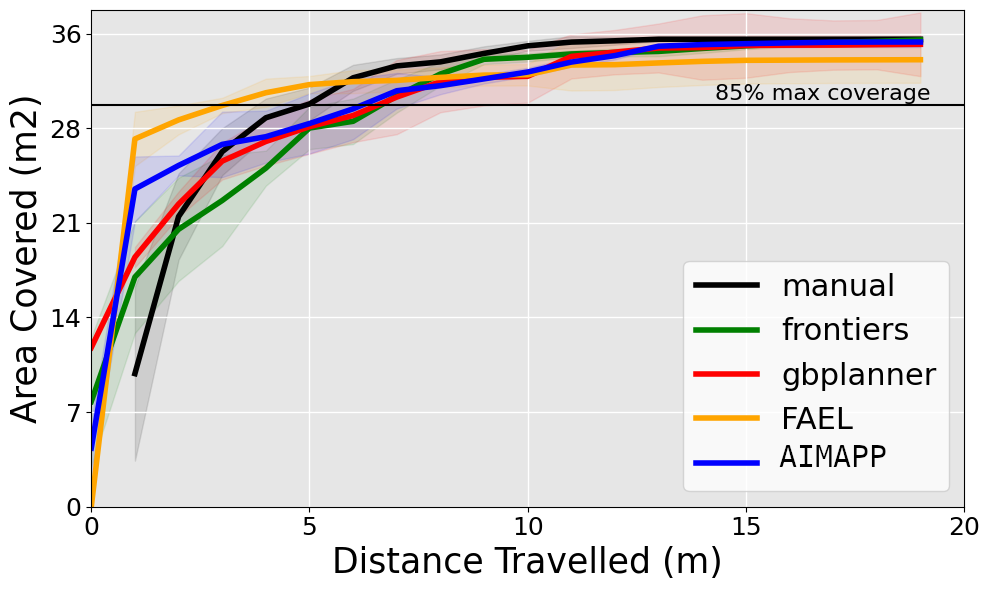}\\
(b) Mini warehouse
\end{minipage}%
\begin{minipage}{0.4\linewidth}
\centering
\includegraphics[width=\linewidth]{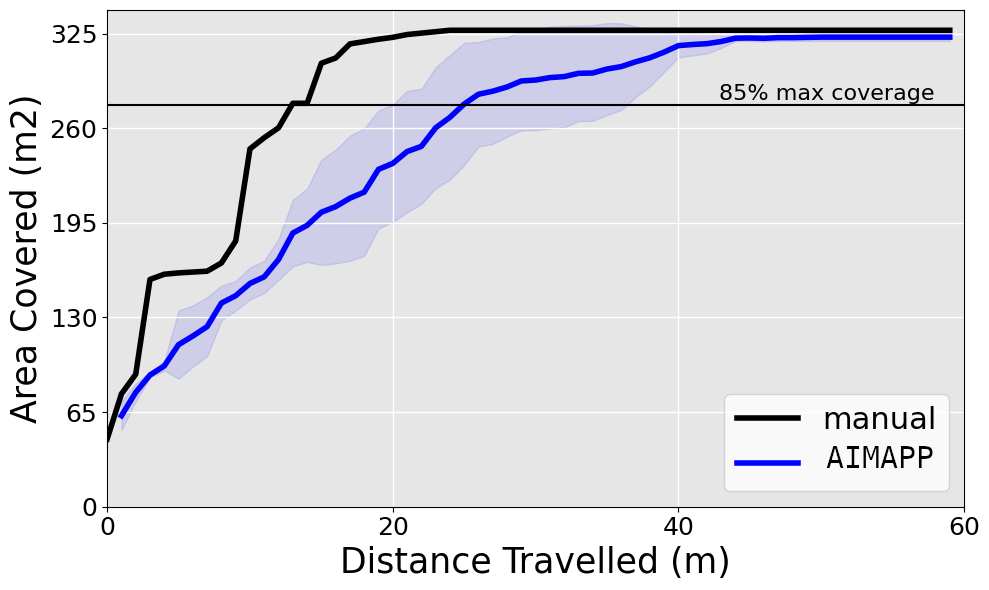}\\
(c) Parking lot coverage
\end{minipage}%
\\[3ex]
\begin{minipage}{0.23\linewidth}
\centering
\includegraphics[width=\linewidth]{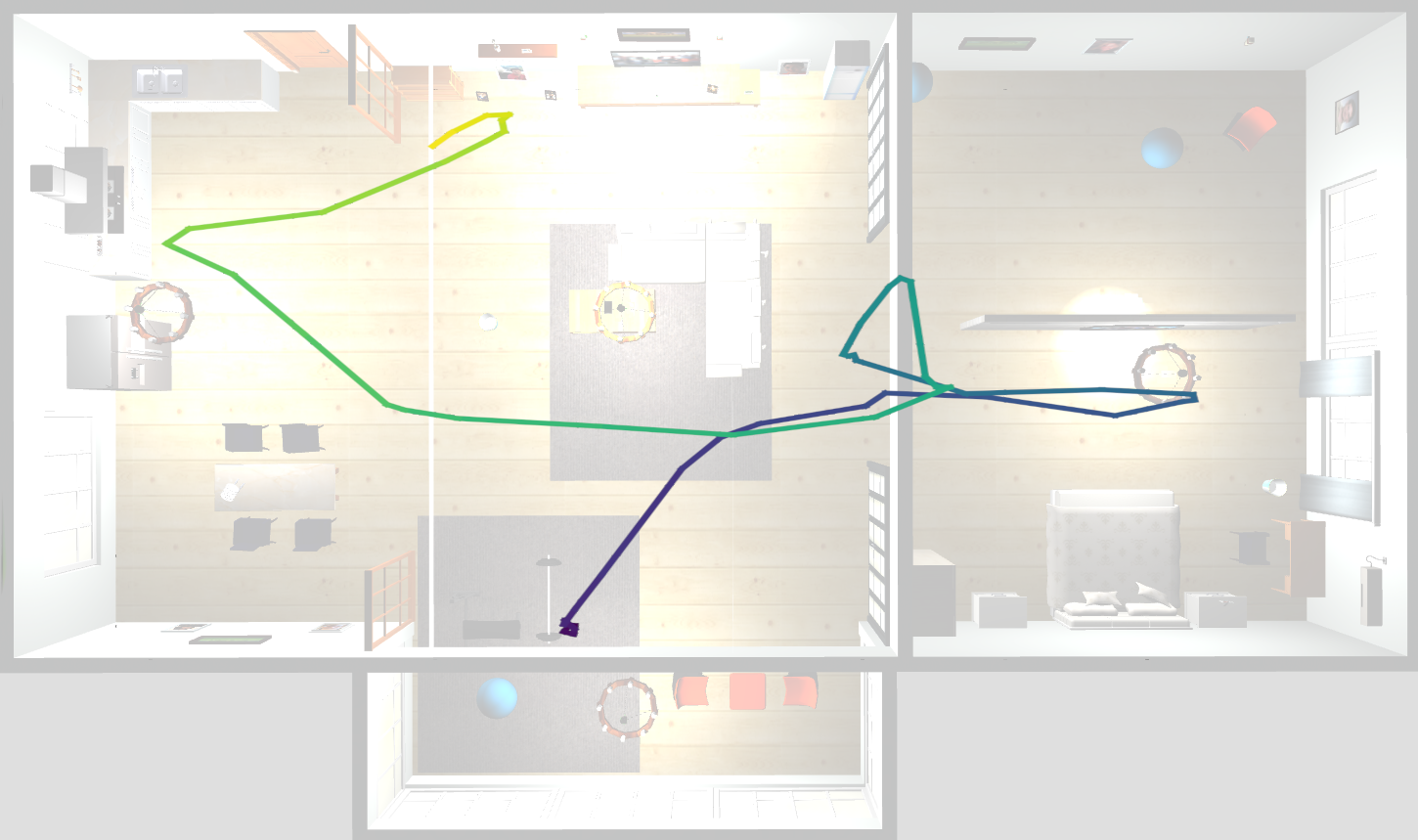}\\
(d) FAEL
\end{minipage}%
\hfill
\begin{minipage}{0.23\linewidth}
\centering
\includegraphics[width=\linewidth]{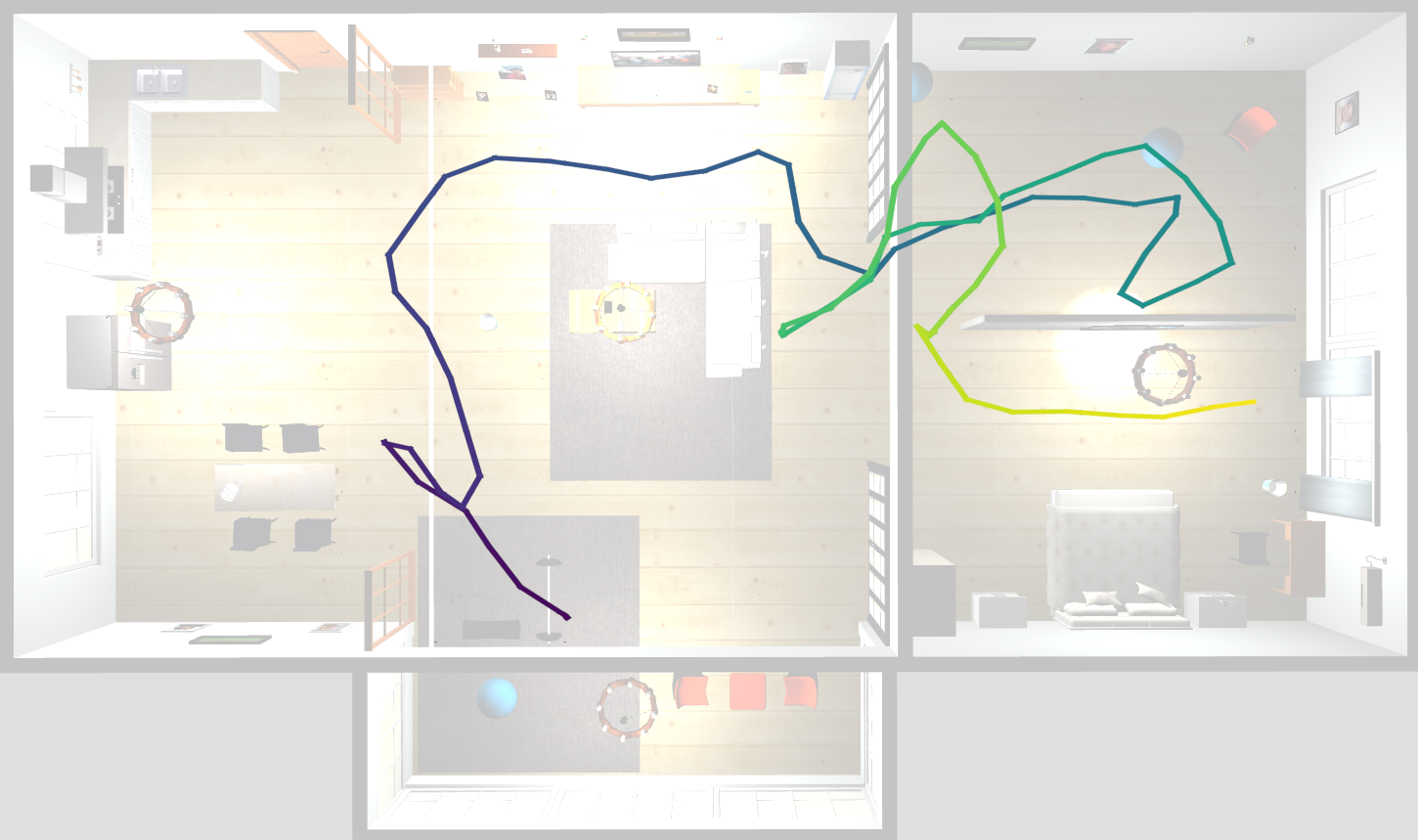}\\
(e) AIMAPP
\end{minipage}%
\hfill
\begin{minipage}{0.23\linewidth}
\centering
\includegraphics[width=\linewidth]{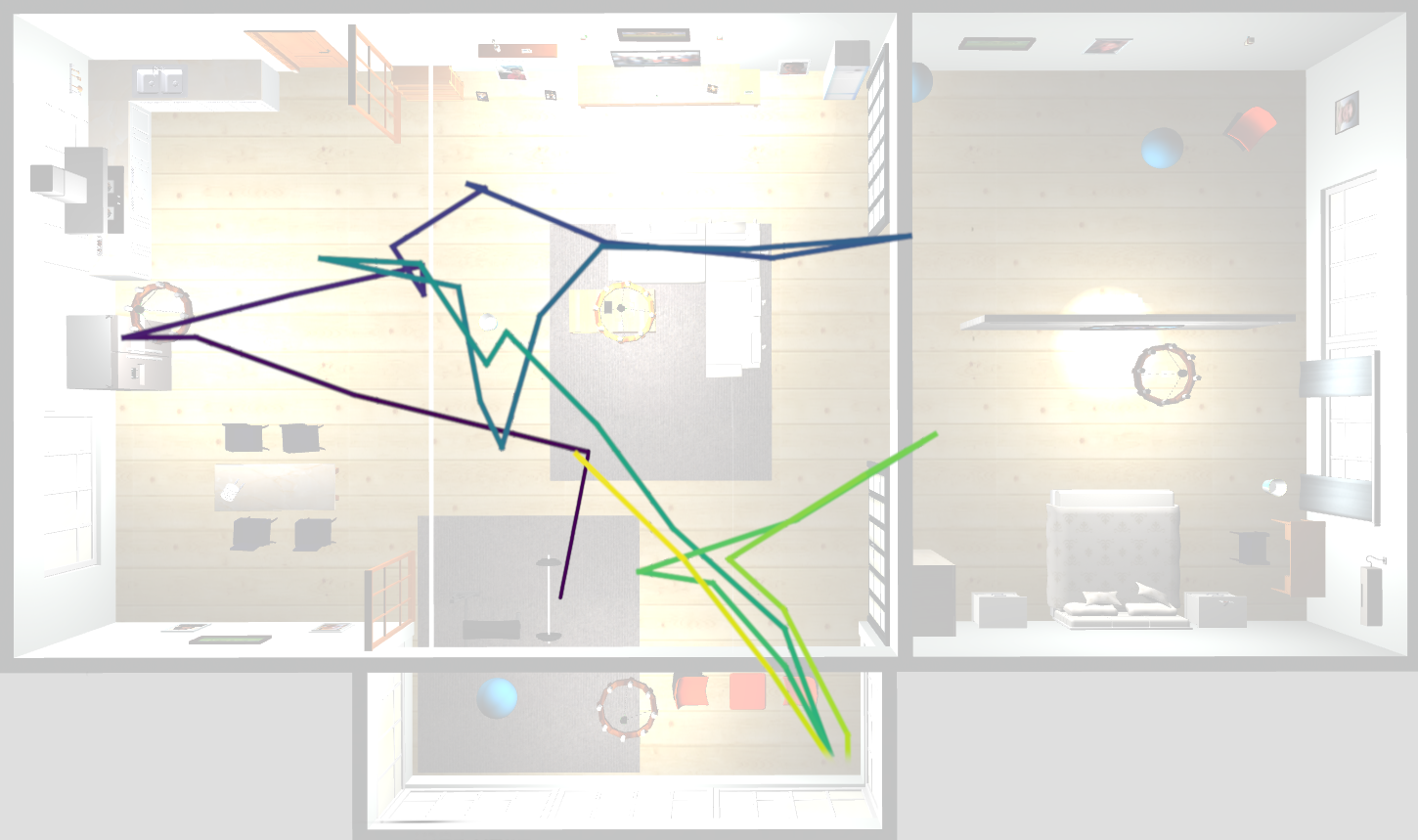}\\
(f) GBPlanner
\end{minipage}%
\hfill
\begin{minipage}{0.23\linewidth}
\centering
\includegraphics[width=\linewidth]{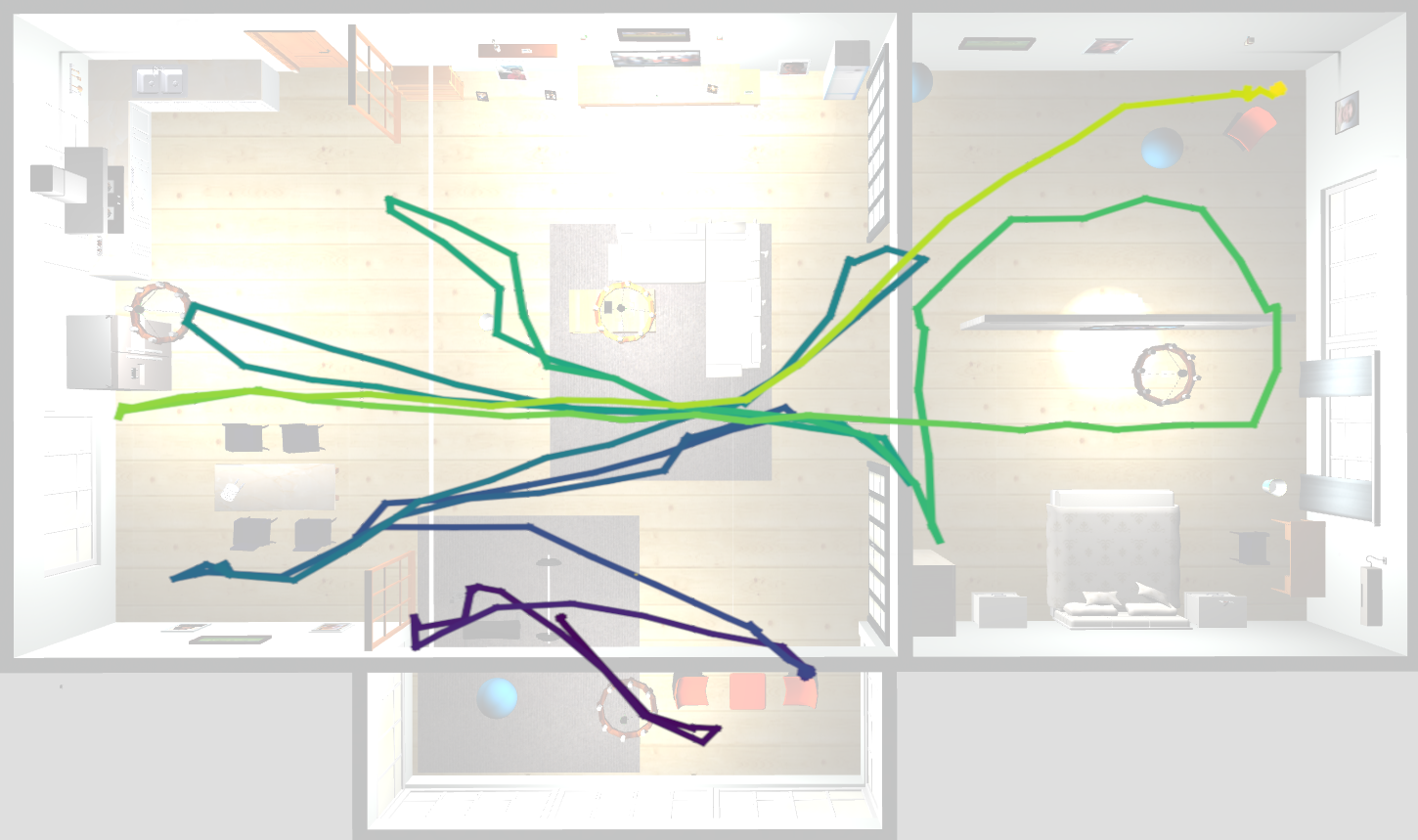}\\
(g) Frontiers
\end{minipage}%
\\[2ex]
\caption{Coverage efficiency in (a) small (80m$^2$), (b) mini warehouses (32m$^2$), and (c) real-world parking lot (325m$^2$). (d--g) Exploration trajectories for all models in the house environment; black dot indicates the shared starting point.}
\label{fig:experimental_results}
\end{figure*}

In smaller environments ($\leq 80 m^2$), model performance was comparable across methods (Figure~\ref{fig:experimental_results}), matching the average of five manual explorations from identical start points (Table \ref{tab:CE_app}). This convergence stems from the limited LiDAR range, which rapidly covers most of the area and reduces the benefit of long-horizon planning.

Exploration trajectories for AIMAPP, FAEL, GBPlanner, and the Frontier-based approach are shown in Figure \ref{fig:experimental_results} (d-g). All agents started near the dumbbell, with trajectories progressing from black to yellow. FAEL, AIMAPP, and Frontiers reached nearly full coverage; GBPlanner achieved $\pm$ 95\%, as it got trapped near the sports room chair.

Each strategy exhibited distinct behaviours. FAEL avoided the playroom but partially observed it from the doorway. Frontiers aggressively targeted unexplored regions, ensuring full coverage but with repeated backtracking and inefficient paths. AIMAPP alternated between forward progression and small corrective loops, accumulating more observations, but in this run, missed the playroom–bedroom connection on the far right. These qualitative differences are minor in small maps but grow significant in larger environments, where planning efficiency and robustness diverge more clearly.

\begin{table}[htb!]
\caption{Exploration efficiency metrics across environments. 
CE: Coverage Efficiency (m$^2$/m), nAUC: Normalised Area Under Coverage Curve. Values are mean~$\pm$~std over 5 runs.}
\begin{tabular}{llcc}
\hline
\textbf{Env.}                                                              & \textbf{Model} & CE            & nAUC          \\ \hline
\multirow{5}{*}{\begin{tabular}[c]{@{}l@{}}Small\\ Warehouse\end{tabular}} & Manual         & 3.13$\pm$0.43 & 0.74$\pm$0.03 \\
                                                                           & AIMAPP           & 2.25$\pm$0.87 & 0.87$\pm$0.04 \\
                                                                           & Frontiers      & 1.08$\pm$0.12 & 0.73$\pm$0.03 \\
                                                                           & FAEL           & 2.60$\pm$0.58 & 0.98$\pm$0.03 \\
                                                                           & GBPlanner      & 1.59$\pm$0.25 & 0.63$\pm$0.06 \\ \hline
\multirow{5}{*}{\begin{tabular}[c]{@{}l@{}}Mini \\ warehouse\end{tabular}} & Manual         & 2.66$\pm$0.16 & 0.92$\pm$0.01 \\
                                                                           & AIMAPP           & 2.64$\pm$0.98 & 0.91$\pm$0.04 \\
                                                                           & Frontiers      & 2.45$\pm$0.07 & 0.88$\pm$0.01 \\
                                                                           & FAEL           & 2.61$\pm$1.16 & 0.85$\pm$0.03 \\
                                                                           & GBPlanner      & 2.55$\pm$0.4  & 0.82$\pm$0.08 \\ \hline
\begin{tabular}[c]{@{}l@{}}Real\\ Garage\end{tabular}                      & Manual         & 13.18         & 0.74          \\
                                                                           & AIMAPP           & 7.41$\pm$0.87 & 0.71$\pm$0.03
\end{tabular}
\label{tab:CE_app}
\vspace{-5mm}
\end{table}

Direct coverage comparisons between AIMAPP and FAEL or GBPlanner could not be performed in real-world settings due to incompatible sensor requirements. Moreover, Frontiers relied on Nav2’s reliance on 2D occupancy maps, which are sensitive to drift and slippage. Consequently, results for Frontiers in the garage and home in the real world could not be fairly compared to our model. In these cases, drift and odometry loss led to misaligned or overlapping maps. To maintain consistency, AIMAPP’s real-world coverage values were averaged over five runs and compared with a manual teleoperation using Nav2 SLAM. Figure \ref{fig:experimental_results}(c) illustrates coverage in the parking lot, where parked cars altered the free space between runs. Despite these variations, AIMAPP achieved consistent and robust coverage across multiple trials.

\subsection{Reasons for human interventions}
\label{app:human_int}

\begin{table}[h!tb]
\vspace{-5mm}
\centering
\caption{Number of times the robot got stuck and required intervention per environment.}
\begin{tabular}{l|cccc}
\begin{tabular}[c]{@{}l@{}}External \\ intervention\end{tabular} & AIMAPP  & FAEL     & Gbplanner  & Frontiers \\ \hline
Home            & 2 & 5 & 1 & 5        \\ \hline
Big warehouse   & 0 & 5 & 2 & 0       \\ \hline
Small warehouse & 1 & 3 & 4 & 2        \\ \hline
Mini warehouse  & 0 & 2 & 2 & 3       \\ \hline
Real home       & 4 & x & x & x       \\ \hline
Real parking    & 2 & x & x & x       \\ \hline
Real warehouse  & 7 & x & x & 14     
\end{tabular}
\label{tab:human_int}
\vspace{-5mm}
\end{table}

The reasons for human interventions varied substantially across models:

\textbf{AIMAPP:} Most interventions followed LiDAR blind spots (e.g., flat-based chairs, forklifts, or reflective walls). In these cases, the robot collided or spun in place, and the motion failure had to be signalled manually. Large drift in unexplored areas occasionally required manual re-localisation for benchmarking consistency. \\
\textbf{Frontiers:} Often persisted in unreachable goals, repeatedly trying to navigate toward them until manually repositioned. \\
\textbf{FAEL:} Failures arose when nodes were created near obstacles in Voxblox or when open-space detection misclassified traversable areas. Repositioning typically resolved issues, though the model sometimes refused to continue for unknown reasons.\\
\textbf{GBPlanner:} Showed the fewest failures but sometimes planned paths through partially detected obstacles, requiring manual adjustment. In AIMAPP, obstacle-related re-planning is delegated to the motion planner (Nav2), which reports motion failure; the decision layer merely selects goals. Interventions to prevent collisions with moving vehicles in the parking lot were excluded from Table \ref{tab:human_int}

\end{appendices}
\FloatBarrier
\bibliography{main}

@misc{ETPNav,
      title={ETPNav: Evolving Topological Planning for Vision-Language Navigation in Continuous Environments}, 
      author={Dong An and Hanqing Wang and Wenguan Wang and Zun Wang and Yan Huang and Keji He and Liang Wang},
      year={2024},
      eprint={2304.03047},
      archivePrefix={arXiv},
      primaryClass={cs.CV},
      url={https://arxiv.org/abs/2304.03047}, 
}

@Article{review_path_plan,
AUTHOR = {Qin, Hongwei and Shao, Shiliang and Wang, Ting and Yu, Xiaotian and Jiang, Yi and Cao, Zonghan},
TITLE = {Review of Autonomous Path Planning Algorithms for Mobile Robots},
JOURNAL = {Drones},
VOLUME = {7},
YEAR = {2023},
NUMBER = {3},
ARTICLE-NUMBER = {211},
URL = {https://www.mdpi.com/2504-446X/7/3/211},
ISSN = {2504-446X},
DOI = {10.3390/drones7030211}
}

@article{GSLAM,
AUTHOR={Safron, Adam  and Çatal, Ozan  and Verbelen, Tim },
TITLE={Generalized Simultaneous Localization and Mapping (G-SLAM) as unification framework for natural and artificial intelligences: towards reverse engineering the hippocampal/entorhinal system and principles of high-level cognition},
JOURNAL={Frontiers in Systems Neuroscience},
VOLUME={Volume 16 - 2022},
YEAR={2022},
URL={https://www.frontiersin.org/journals/systems-neuroscience/articles/10.3389/fnsys.2022.787659},
DOI={10.3389/fnsys.2022.787659},
ISSN={1662-5137},
language = {eng},
organization = {Moran, Rosalyn J.},
}

@INPROCEEDINGS{ratslam,
  author={Milford, M.J. and Wyeth, G.F. and Prasser, D.},
  booktitle={IEEE International Conference on Robotics and Automation, 2004. Proceedings. ICRA '04. 2004}, 
  title={RatSLAM: a hippocampal model for simultaneous localization and mapping}, 
  year={2004},
  volume={1},
  number={},
  pages={403-408 Vol.1},
  doi={10.1109/ROBOT.2004.1307183}}

@inproceedings{NNslam,
  title={Learning To Explore Using Active Neural SLAM},
  author={Chaplot, Devendra Singh and Gandhi, Dhiraj and Gupta, Saurabh and Gupta, Abhinav and Salakhutdinov, Ruslan},
  booktitle={International Conference on Learning Representations (ICLR)},
  year={2020}
}

@article{orbslam3,
  title={{ORB-SLAM3}: An Accurate Open-Source Library for Visual, Visual-Inertial 
           and Multi-Map {SLAM}},
  author={Campos, Carlos AND Elvira, Richard AND Gomez, Juan J. AND Montiel, 
          Jose M. M. AND Tardos, Juan D.},
  journal={IEEE Transactions on Robotics}, 
  volume={37},
  number={6},
  pages={1874-1890},
  year={2021}
 }

@article{depth_based_lidar_slam,
  author={H. {Wang} and C. {Wang} and L. {Xie}},
  journal={IEEE Robotics and Automation Letters}, 
  title={Lightweight 3-D Localization and Mapping for Solid-State LiDAR}, 
  year={2021},
  volume={6},
  number={2},
  pages={1801-1807},
  doi={10.1109/LRA.2021.3060392}}

@article{FAST-LIO2,
  author       = {Wei Xu and
                  Yixi Cai and
                  Dongjiao He and
                  Jiarong Lin and
                  Fu Zhang},
  title        = {{FAST-LIO2:} Fast Direct LiDAR-inertial Odometry},
  journal      = {CoRR},
  volume       = {abs/2107.06829},
  year         = {2021},
  url          = {https://arxiv.org/abs/2107.06829},
  eprinttype    = {arXiv},
  eprint       = {2107.06829},
  bibsource    = {dblp computer science bibliography, https://dblp.org}
}

@misc{PLP-SLAM,
      title={Structure PLP-SLAM: Efficient Sparse Mapping and Localization using Point, Line and Plane for Monocular, RGB-D and Stereo Cameras}, 
      author={Fangwen Shu and Jiaxuan Wang and Alain Pagani and Didier Stricker},
      year={2023},
      eprint={2207.06058},
      archivePrefix={arXiv},
      primaryClass={cs.CV},
      url={https://arxiv.org/abs/2207.06058}, 
}

@article{map_induction,
  author       = {Sugandha Sharma and
                  Aidan Curtis and
                  Marta Kryven and
                  Joshua B. Tenenbaum and
                  Ila R. Fiete},
  title        = {Map Induction: Compositional spatial submap learning for efficient
                  exploration in novel environments},
  journal      = {CoRR},
  volume       = {abs/2110.12301},
  year         = {2021},
  url          = {https://arxiv.org/abs/2110.12301},
  eprinttype    = {arXiv},
  eprint       = {2110.12301},
  bibsource    = {dblp computer science bibliography, https://dblp.org}
}

@article{gaussian_process_nav,
  title={Autonomous Mapless Navigation on Uneven Terrains},
  author={Hassan Jardali and Mahmoud Ali and Lantao Liu},
  journal={2024 IEEE International Conference on Robotics and Automation (ICRA)},
  year={2024},
  pages={13227-13233},
  url={https://api.semanticscholar.org/CorpusID:267770585}
}

@misc{Nomad,
      title={NoMaD: Goal Masked Diffusion Policies for Navigation and Exploration}, 
      author={Ajay Sridhar and Dhruv Shah and Catherine Glossop and Sergey Levine},
      year={2023},
      eprint={2310.07896},
      archivePrefix={arXiv},
      primaryClass={cs.RO},
      url={https://arxiv.org/abs/2310.07896}, 
}

@misc{RECON,
      title={Rapid Exploration for Open-World Navigation with Latent Goal Models}, 
      author={Dhruv Shah and Benjamin Eysenbach and Gregory Kahn and Nicholas Rhinehart and Sergey Levine},
      year={2023},
      eprint={2104.05859},
      archivePrefix={arXiv},
      primaryClass={cs.RO},
      url={https://arxiv.org/abs/2104.05859}, 
}

@misc{BYOL,
      title={BYOL-Explore: Exploration by Bootstrapped Prediction}, 
      author={Zhaohan Daniel Guo and Shantanu Thakoor and Miruna Pîslar and Bernardo Avila Pires and Florent Altché and Corentin Tallec and Alaa Saade and Daniele Calandriello and Jean-Bastien Grill and Yunhao Tang and Michal Valko and Rémi Munos and Mohammad Gheshlaghi Azar and Bilal Piot},
      year={2022},
      eprint={2206.08332},
      archivePrefix={arXiv},
      primaryClass={cs.LG},
      url={https://arxiv.org/abs/2206.08332}, 
}

@inproceedings{viking,
   title={ViKiNG: Vision-Based Kilometer-Scale Navigation with Geographic Hints},
   url={http://dx.doi.org/10.15607/RSS.2022.XVIII.019},
   DOI={10.15607/rss.2022.xviii.019},
   booktitle={Robotics: Science and Systems XVIII},
   publisher={Robotics: Science and Systems Foundation},
   author={Shah, Dhruv and Levine, Sergey},
   year={2022},
   month=jun, 
   collection={RSS2022},
}

@misc{pietro,
      title={Learning to Navigate from Scratch using World Models and Curiosity: the Good, the Bad, and the Ugly}, 
      author={Daria de Tinguy and Sven Remmery and Pietro Mazzaglia and Tim Verbelen and Bart Dhoedt},
      year={2023},
      eprint={2308.15852},
      archivePrefix={arXiv},
      primaryClass={cs.RO},
      url={https://arxiv.org/abs/2308.15852}, 
}

@ARTICLE{path_planning_survey,
  author={Hewawasam, H. S. and Ibrahim, M. Yousef and Appuhamillage, Gayan Kahandawa},
  journal={IEEE Open Journal of the Industrial Electronics Society}, 
  title={Past, Present and Future of Path-Planning Algorithms for Mobile Robot Navigation in Dynamic Environments}, 
  year={2022},
  volume={3},
  number={},
  pages={353-365},
  doi={10.1109/OJIES.2022.3179617}
}

@misc{AIF_survey_robotics,
      title={Active Inference in Robotics and Artificial Agents: Survey and Challenges}, 
      author={Pablo Lanillos and Cristian Meo and Corrado Pezzato and Ajith Anil Meera and Mohamed Baioumy and Wataru Ohata and Alexander Tschantz and Beren Millidge and Martijn Wisse and Christopher L. Buckley and Jun Tani},
      year={2021},
      eprint={2112.01871},
      archivePrefix={arXiv},
      primaryClass={cs.RO},
      url={https://arxiv.org/abs/2112.01871}, 
}

@Article{bio-inspired-robots,
author = {Junfei Li and Zhe Xu and Danjie Zhu and Kevin Dong and Tao Yan and Zhu Zeng and Simon X. Yang},
TITLE = {Bio-inspired intelligence with applications to robotics: a survey},
JOURNAL = {Intelligence and Robotics},
VOLUME = {1},
YEAR = {2021},
NUMBER = {1},
ARTICLE-NUMBER = {58-83},
URL = {https://www.oaepublish.com/articles/ir.2021.08},
ISSN = {2770-3541},
DOI = {10.20517/ir.2021.08}
}

@misc{survey_RL_robots,
author = {Dehghani Tezerjani, Mohammad and Khoshnazar, Mohammad and Tangestanizade, Mohammadhamed and Yang, Qing},
year = {2024},
month = {07},
pages = {},
title = {A Survey on Reinforcement Learning Applications in SLAM},
doi = {10.48550/arXiv.2408.14518}
}

@article{RL_path_plan,
  author       = {Piotr Mirowski and
                  Razvan Pascanu and
                  Fabio Viola and
                  Hubert Soyer and
                  Andrew J. Ballard and
                  Andrea Banino and
                  Misha Denil and
                  Ross Goroshin and
                  Laurent Sifre and
                  Koray Kavukcuoglu and
                  Dharshan Kumaran and
                  Raia Hadsell},
  title        = {Learning to Navigate in Complex Environments},
  journal      = {CoRR},
  volume       = {abs/1611.03673},
  year         = {2016},
  url          = {http://arxiv.org/abs/1611.03673},
  eprinttype    = {arXiv},
  eprint       = {1611.03673},
  bibsource    = {dblp computer science bibliography, https://dblp.org}
}

@misc{world_model_explo,
      title={Discovering and Achieving Goals via World Models}, 
      author={Russell Mendonca and Oleh Rybkin and Kostas Daniilidis and Danijar Hafner and Deepak Pathak},
      year={2021},
      eprint={2110.09514},
      archivePrefix={arXiv},
      primaryClass={cs.LG},
      url={https://arxiv.org/abs/2110.09514}, 
}

@ARTICLE{RHNVBP,
  author={Selin, Magnus and Tiger, Mattias and Duberg, Daniel and Heintz, Fredrik and Jensfelt, Patric},
  journal={IEEE Robotics and Automation Letters}, 
  title={Efficient Autonomous Exploration Planning of Large-Scale 3-D Environments}, 
  year={2019},
  volume={4},
  number={2},
  pages={1699-1706},
  doi={10.1109/LRA.2019.2897343}}

@misc{grid-cell_fragmentation_map_building,
      title={Grid Cell-Inspired Fragmentation and Recall for Efficient Map Building}, 
      author={Jaedong Hwang and Zhang-Wei Hong and Eric Chen and Akhilan Boopathy and Pulkit Agrawal and Ila Fiete},
      year={2024},
      eprint={2307.05793},
      archivePrefix={arXiv},
      primaryClass={cs.AI},
      url={https://arxiv.org/abs/2307.05793}, 
}

@article{vision_based_place_reco,
author = {Michael Milford},
title ={Vision-based place recognition: how low can you go?},
journal = {The International Journal of Robotics Research},
volume = {32},
number = {7},
pages = {766-789},
year = {2013},
doi = {10.1177/0278364913490323},
URL = {https://doi.org/10.1177/0278364913490323},
eprint = {https://doi.org/10.1177/0278364913490323
}}

@ARTICLE{MCTS,
  author={Browne, Cameron B. and Powley, Edward and Whitehouse, Daniel and Lucas, Simon M. and Cowling, Peter I. and Rohlfshagen, Philipp and Tavener, Stephen and Perez, Diego and Samothrakis, Spyridon and Colton, Simon},
  journal={IEEE Transactions on Computational Intelligence and AI in Games}, 
  title={A Survey of Monte Carlo Tree Search Methods}, 
  year={2012},
  volume={4},
  number={1},
  pages={1-43},
  doi={10.1109/TCIAIG.2012.2186810}}

@article{object_nav_zero_shot,
   title={Can an Embodied Agent Find Your “Cat-shaped Mug”? LLM-Based Zero-Shot Object Navigation},
   volume={9},
   ISSN={2377-3774},
   url={http://dx.doi.org/10.1109/LRA.2023.3346800},
   DOI={10.1109/lra.2023.3346800},
   number={5},
   journal={IEEE Robotics and Automation Letters},
   publisher={Institute of Electrical and Electronics Engineers (IEEE)},
   author={Dorbala, Vishnu Sashank and Mullen, James F. and Manocha, Dinesh},
   year={2024},
   month=may, pages={4083–4090} }

@misc{object_nav_semantic,
      title={Object Goal Navigation using Goal-Oriented Semantic Exploration}, 
      author={Devendra Singh Chaplot and Dhiraj Gandhi and Abhinav Gupta and Ruslan Salakhutdinov},
      year={2020},
      eprint={2007.00643},
      archivePrefix={arXiv},
      primaryClass={cs.CV},
      url={https://arxiv.org/abs/2007.00643}, 
}

@misc{frontiers,
      title={Frontier Based Exploration for Autonomous Robot}, 
      author={Anirudh Topiwala and Pranav Inani and Abhishek Kathpal},
      year={2018},
      month = {01},
      pages = {},
      eprint={1806.03581},
      archivePrefix={arXiv},
      primaryClass={cs.RO},
      url={https://arxiv.org/abs/1806.03581}, 
      doi = {10.13140/RG.2.2.34130.40641}
}

@article{gbplanner,
  title={Graph-based subterranean exploration path planning using aerial and legged robots},
  author={Dang, Tung and Tranzatto, Marco and Khattak, Shehryar and Mascarich, Frank and Alexis, Kostas and Hutter, Marco},
  journal={Journal of Field Robotics},
  volume = {37},
  number = {8},
  pages = {1363-1388},  
  year={2020},
  note={Wiley Online Library}
}

@misc{Darpa_winners,
      title={Team CERBERUS Wins the DARPA Subterranean Challenge: Technical Overview and Lessons Learned}, 
      author={Marco Tranzatto and Mihir Dharmadhikari and Lukas Bernreiter and Marco Camurri and Shehryar Khattak and Frank Mascarich and Patrick Pfreundschuh and David Wisth and Samuel Zimmermann and Mihir Kulkarni and Victor Reijgwart and Benoit Casseau and Timon Homberger and Paolo De Petris and Lionel Ott and Wayne Tubby and Gabriel Waibel and Huan Nguyen and Cesar Cadena and Russell Buchanan and Lorenz Wellhausen and Nikhil Khedekar and Olov Andersson and Lintong Zhang and Takahiro Miki and Tung Dang and Matias Mattamala and Markus Montenegro and Konrad Meyer and Xiangyu Wu and Adrien Briod and Mark Mueller and Maurice Fallon and Roland Siegwart and Marco Hutter and Kostas Alexis},
      year={2022},
      eprint={2207.04914},
      archivePrefix={arXiv},
      primaryClass={cs.RO},
      url={https://arxiv.org/abs/2207.04914}, 
}

@article{FAEL,
  title={FAEL: Fast Autonomous Exploration for Large-scale Environments With a Mobile Robot},
  author={Junlong Huang and Boyu Zhou and Zhengping Fan and Yilin Zhu and Yingrui Jie and Longwei Li and Hui Cheng},
  journal={IEEE Robotics and Automation Letters},
  year={2023},
  volume={8},
  pages={1667-1674},
  url={https://api.semanticscholar.org/CorpusID:255901478}
}

@article{ufomap,
  author={Daniel Duberg and Patric Jensfelt},
  journal={IEEE Robotics and Automation Letters}, 
  title={{UFOMap}: An Efficient Probabilistic {3D} Mapping Framework That Embraces the Unknown}, 
  year={2020},
  volume={5},
  number={4},
  pages={6411-6418},
  doi={10.1109/LRA.2020.3013861}
}

@article{voxblox,
  author       = {Helen Oleynikova and
                  Zachary Taylor and
                  Marius Fehr and
                  Juan I. Nieto and
                  Roland Siegwart},
  title        = {Voxblox: Building 3D Signed Distance Fields for Planning},
  journal      = {CoRR},
  volume       = {abs/1611.03631},
  year         = {2016},
  url          = {http://arxiv.org/abs/1611.03631},
  eprinttype    = {arXiv},
  eprint       = {1611.03631},
  bibsource    = {dblp computer science bibliography, https://dblp.org}
}

@article{nav2_slam,
  author    = {Pratham Gyanani and Mridul Agarwal and Ranjeet Osari and others},
  title     = {Autonomous Mobile Vehicle Using ROS2 and 2D-Lidar and SLAM Navigation},
  year      = {2024},
  month     = {May},
  journal   = {Research Square},
  volume    = {Preprint (Version 1)},
  doi       = {10.21203/rs.3.rs-4323431/v1},
  url       = {https://doi.org/10.21203/rs.3.rs-4323431/v1},
  note      = {Available at Research Square}
}

@article {Johnson2007,
	author = {Johnson, Adam and Redish, A. David},
	title = {Neural Ensembles in CA3 Transiently Encode Paths Forward of the Animal at a Decision Point},
	volume = {27},
	number = {45},
	pages = {12176--12189},
	year = {2007},
	doi = {10.1523/JNEUROSCI.3761-07.2007},
	publisher = {Society for Neuroscience},
	issn = {0270-6474},
	URL = {https://www.jneurosci.org/content/27/45/12176},
	eprint = {https://www.jneurosci.org/content/27/45/12176.full.pdf},
	journal = {Journal of Neuroscience}
}

@article{Stachenfeld2017,
author = {Stachenfeld, Kimberly and Botvinick, Matthew and Gershman, Samuel},
year = {2017},
journal   = {Nature Neuroscience},
volume    = {20},
number    = {11},
pages     = {1643--1653},
title = {The hippocampus as a predictive map},
doi = {10.1101/097170}
}

@article{Kay2020,
title = {Constant Sub-second Cycling between Representations of Possible Futures in the Hippocampus},
journal = {Cell},
volume = {180},
number = {3},
pages = {552-567.e25},
year = {2020},
issn = {0092-8674},
doi = {https://doi.org/10.1016/j.cell.2020.01.014},
url = {https://www.sciencedirect.com/science/article/pii/S0092867420300611},
author = {Kenneth Kay and Jason E. Chung and Marielena Sosa and Jonathan S. Schor and Mattias P. Karlsson and Margaret C. Larkin and Daniel F. Liu and Loren M. Frank}
}

@article{mice_in_labyrith,
article_type = {journal},
title = {Mice in a labyrinth show rapid learning, sudden insight, and efficient exploration},
author = {Rosenberg, Matthew and Zhang, Tony and Perona, Pietro and Meister, Markus},
editor = {Mathis, Mackenzie W and Dulac, Catherine and Berman, Gordon J},
volume = 10,
year = 2021,
month = {jul},
pub_date = {2021-07-01},
pages = {e66175},
citation = {eLife 2021;10:e66175},
doi = {10.7554/eLife.66175},
url = {https://doi.org/10.7554/eLife.66175},
journal = {eLife},
issn = {2050-084X},
publisher = {eLife Sciences Publications, Ltd},
}

@article{hippo_nav,
title = {Assembly Responses of Hippocampal CA1 Place Cells Predict Learned Behavior in Goal-Directed Spatial Tasks on the Radial Eight-Arm Maze},
journal = {Neuron},
volume = {101},
number = {1},
pages = {119-132.e4},
year = {2019},
issn = {0896-6273},
doi = {https://doi.org/10.1016/j.neuron.2018.11.015},
url = {https://www.sciencedirect.com/science/article/pii/S0896627318310067},
author = {Haibing Xu and Peter Baracskay and Joseph O’Neill and Jozsef Csicsvari}
}

@article{Human_rodent_spatial_rep,
author = {Zhao, Mintao},
year = {2018},
month = {08},
pages = {},
title = {Human Spatial Representation: What We Cannot Learn from the Studies of Rodent Navigation},
volume = {120},
journal = {Journal of Neurophysiology},
doi = {10.1152/jn.00781.2017}
}

@article{humans-cognitive-map,
author = {Epstein, Russell and Patai, E Z and Julian, Joshua and Spiers, Hugo},
year = {2017},
month = {10},
pages = {1504-1513},
title = {The cognitive map in humans: Spatial navigation and beyond},
volume = {20},
journal = {Nature Neuroscience},
doi = {10.1038/nn.4656}
}

@article{humans-mapping,
author = {Foo, Patrick and Warren, William and Duchon, Andrew and Tarr, Michael},
year = {2005},
month = {04},
pages = {195-215},
title = {Do Humans Integrate Routes Into a Cognitive Map? Map- Versus Landmark-Based Navigation of Novel Shortcuts.},
volume = {31},
journal = {Journal of experimental psychology. Learning, memory, and cognition},
doi = {10.1037/0278-7393.31.2.195}
}

@article{cognitive_graph,
title = {Wormholes in virtual space: From cognitive maps to cognitive graphs},
journal = {Cognition},
volume = {166},
pages = {152-163},
year = {2017},
issn = {0010-0277},
doi = {https://doi.org/10.1016/j.cognition.2017.05.020},
url = {https://www.sciencedirect.com/science/article/pii/S0010027717301373},
author = {William H. Warren and Daniel B. Rothman and Benjamin H. Schnapp and Jonathan D. Ericson}
}

@article{grid_cell_nav,
  title={Using Grid Cells for Navigation},
  author={Daniel Bush and Caswell Barry and Daniel Manson and Neil Burgess},
  journal={Neuron},
  year={2015},
  volume={87},
  pages={507 - 520},
  url={https://api.semanticscholar.org/CorpusID:7275119}
}

@book{AIF_book,
author = {Parr, Thomas and Pezzulo, Giovanni and Friston, Karl},
year = {2022},
month = {03},
pages = {},
title = {Active Inference: The Free Energy Principle in Mind, Brain, and Behavior},
isbn = {9780262369978},
doi = {10.7551/mitpress/12441.001.0001},
publisher = {The MIT Press},
}

@article{plan_nav_AIF_friston,
author = {Kaplan, Raphael and Friston, Karl J.},
title = {Planning and navigation as active inference},
year = {2018},
issue_date = {August    2018},
publisher = {Springer-Verlag},
address = {Berlin, Heidelberg},
volume = {112},
number = {4},
issn = {0340-1200},
url = {https://doi.org/10.1007/s00422-018-0753-2},
doi = {10.1007/s00422-018-0753-2},
journal = {Biol. Cybern.},
month = aug,
pages = {323–343},
numpages = {21}
}

@article{sophisticated_AIF,
author = {Friston, Karl and Da Costa, Lancelot and Hafner, Danijar and Hesp, Casper and Parr, Thomas},
year = {2021},
month = {03},
pages = {713-763},
title = {Sophisticated Inference},
volume = {33},
journal = {Neural Computation},
doi = {10.1162/neco_a_01351}
}

@ARTICLE{AIF_bandit_explo_LATEST,
  author={Wakayama, Shohei and Candela, Alberto and Hayne, Paul and Ahmed, Nisar},
  journal={IEEE Transactions on Robotics}, 
  title={Active Inference for Bandit-Based Autonomous Robotic Exploration With Dynamic Preferences}, 
  year={2025},
  volume={41},
  number={},
  pages={3841-3851},
  doi={10.1109/TRO.2025.3577041}}

@article{robot_nav_hierarchy_ozan,
title = {Robot navigation as hierarchical active inference},
journal = {Neural Networks},
volume = {142},
pages = {192-204},
year = {2021},
issn = {0893-6080},
doi = {https://doi.org/10.1016/j.neunet.2021.05.010},
url = {https://www.sciencedirect.com/science/article/pii/S0893608021002021},
author = {Ozan Çatal and Tim Verbelen and Toon {Van de Maele} and Bart Dhoedt and Adam Safron},
}

@article{world_model_and_inference,
title = {World model learning and inference},
journal = {Neural Networks},
volume = {144},
pages = {573-590},
year = {2021},
issn = {0893-6080},
doi = {https://doi.org/10.1016/j.neunet.2021.09.011},
url = {https://www.sciencedirect.com/science/article/pii/S0893608021003610},
author = {Karl Friston and Rosalyn J. Moran and Yukie Nagai and Tadahiro Taniguchi and Hiroaki Gomi and Josh Tenenbaum},
}

@misc{inductive_AIF,
      title={Active Inference and Intentional Behaviour}, 
      author={Karl J. Friston and Tommaso Salvatori and Takuya Isomura and Alexander Tschantz and Alex Kiefer and Tim Verbelen and Magnus Koudahl and Aswin Paul and Thomas Parr and Adeel Razi and Brett Kagan and Christopher L. Buckley and Maxwell J. D. Ramstead},
      year={2023},
      eprint={2312.07547},
      archivePrefix={arXiv},
      primaryClass={q-bio.NC},
      url={https://arxiv.org/abs/2312.07547}, 
}

@article{bayesian_model_reduc,
      title={Bayesian model reduction}, 
      author={Karl Friston and Thomas Parr and Peter Zeidman},
      year={2019},
      eprint={1805.07092},
      journal= {},
      archivePrefix={arXiv},
      primaryClass={stat.ME}
}

@article{surpervised_struct_learning,
      title={Supervised structure learning}, 
      author={Karl J. Friston and Lancelot Da Costa and Alexander Tschantz and Alex Kiefer and Tommaso Salvatori and Victorita Neacsu and Magnus Koudahl and Conor Heins and Noor Sajid and Dimitrije Markovic and Thomas Parr and Tim Verbelen and Christopher L Buckley},
      year={2023},
      journal= {},
      eprint={2311.10300},
      archivePrefix={arXiv},
      primaryClass={cs.LG}
}

@article{aif_nav_imitation,
author = {Nozari, Sheida and Krayani, Ali and Marin, Pablo and Marcenaro, Lucio and Martín Gómez, David and Regazzoni, Carlo},
year = {2024},
month = {10},
pages = {},
title = {Exploring action-oriented models via active inference for autonomous vehicles},
volume = {2024},
journal = {EURASIP Journal on Advances in Signal Processing},
doi = {10.1186/s13634-024-01173-9}
}

@article{weird_HAIF,
    doi = {10.1371/journal.pone.0277199},
    author = {Neacsu, Victorita AND Mirza, M. Berk AND Adams, Rick A. AND Friston, Karl J.},
    journal = {PLOS ONE},
    publisher = {Public Library of Science},
    title = {Structure learning enhances concept formation in synthetic Active Inference agents},
    year = {2022},
    month = {11},
    volume = {17},
    url = {https://doi.org/10.1371/journal.pone.0277199},
    pages = {1-34},
    number = {11},

}

@inproceedings{MCTS_AIF,
      title={Deep active inference agents using Monte-Carlo methods}, 
      author={Zafeirios Fountas and Noor Sajid and Pedro A. M. Mediano and Karl Friston},
      year={2020},
      booktitle = {Advances in Neural Information Processing Systems},
      editor = {H. Larochelle and M. Ranzato and R. Hadsell and M. F. Balcan and H. Lin},
      pages = {11662-11675},
      publisher = {Curran Associates, Inc.},
      url = {https://proceedings.neurips.cc/paper/2020/file/865dfbde8a344b44095495f3591f7407-Paper.pdf},
      volume = {33},
}

@misc{wouter_message_passing,
      title={A Message Passing Realization of Expected Free Energy Minimization}, 
      author={Wouter W. L. Nuijten and Mykola Lukashchuk and Thijs van de Laar and Bert de Vries},
      year={2025},
      eprint={2508.02197},
      archivePrefix={arXiv},
      primaryClass={cs.AI},
      url={https://arxiv.org/abs/2508.02197}, 
}

@article{cscg_structuring_knowledge,
title = {Structuring Knowledge with Cognitive Maps and Cognitive Graphs},
journal = {Trends in Cognitive Sciences},
volume = {25},
number = {1},
pages = {37-54},
year = {2021},
issn = {1364-6613},
doi = {https://doi.org/10.1016/j.tics.2020.10.004},
url = {https://www.sciencedirect.com/science/article/pii/S1364661320302503},
author = {Michael Peer and Iva K. Brunec and Nora S. Newcombe and Russell A. Epstein},
}

@article{ours_model,
   title={Learning dynamic cognitive map with autonomous navigation},
   volume={18},
   ISSN={1662-5188},
   url={http://dx.doi.org/10.3389/fncom.2024.1498160},
   DOI={10.3389/fncom.2024.1498160},
   journal={Frontiers in Computational Neuroscience},
   publisher={Frontiers Media SA},
   author={de Tinguy, Daria and Verbelen, Tim and Dhoedt, Bart},
   year={2024},
   month=dec }

@article{ours_hierarchy,
  articleno    = {{83}},
  author       = {{de Tinguy, Daria and Van de Maele, Toon and Verbelen, Tim and Dhoedt, Bart}},
  issn         = {{1099-4300}},
  journal      = {{ENTROPY}},
  language     = {{eng}},
  number       = {{1}},
  pages        = {{32}},
  title        = {{Spatial and temporal hierarchy for autonomous navigation using active inference in minigrid environment}},
  url          = {{http://doi.org/10.3390/e26010083}},
  volume       = {{26}},
  year         = {{2024}},
}

@inproceedings{chunk_space,
   title={The Information Geometry of Space and Time},
   volume={803},
   ISSN={0094-243X},
   url={http://dx.doi.org/10.1063/1.2149814},
   DOI={10.1063/1.2149814},
   booktitle={AIP Conference Proceedings},
   publisher={AIP},
   author={Caticha, Ariel},
   year={2005},
   pages={355–365} }

@misc{turtlebots,
  author       = {{Turtlebot}},
  title        = {Turtlebot versions},
  year         = {2024},
  url          = {https://www.turtlebot.com/about/},
  note         = {Accessed: 2024-12-16}
}

@misc{jackal,
  author       = {{clearpathrobotics}},
  title        = {Jackal},
  url          = {https://clearpathrobotics.com/jackal-small-unmanned-ground-vehicle/},
  note         = {Accessed: 2025-07-28}
}

@misc{rosbotXL,
  author       = {{husarion}},
  title        = {rosbotXL},
  url          = {https://husarion.com/manuals/rosbot-xl/},
  note         = {Accessed: 2025-07-28}
}

@misc{warehouse,
  author       = {{aws-robotics}},
  title        = {aws-robomaker-small-warehouse-world},
  year         = {2020},
  url          = {https://github.com/aws-robotics/aws-robomaker-small-warehouse-world},
  note         = {Accessed: 2024-08-01}
}

@misc{house,
  author       = {{aws-robotics}},
  title        = {aws-robomaker-small-house-world},
  year         = {2021},
  url          = {https://github.com/aws-robotics/aws-robomaker-small-house-world},
  note         = {Accessed: 2024-10-01}
}

@misc{nav2,
  author       = {{nav2}},
  title        = {nav2},
  year         = {2021},
  url          = {https://docs.nav2.org/},
  note         = {Accessed: 2024-12-01}
}

@article{Liu_2025,
   title={Physics-Informed Neural Mapping and Motion Planning in Unknown Environments},
   volume={41},
   ISSN={1941-0468},
   url={http://dx.doi.org/10.1109/TRO.2025.3548495},
   DOI={10.1109/tro.2025.3548495},
   journal={IEEE Transactions on Robotics},
   publisher={Institute of Electrical and Electronics Engineers (IEEE)},
   author={Liu, Yuchen and Ni, Ruiqi and Qureshi, Ahmed H.},
   year={2025},
   pages={2200–2212} }


\end{document}